%% file: report.tex
\documentclass[preprint,review,12pt]{elsarticle}

\usepackage{amssymb}

\usepackage{algorithm}
\usepackage{algorithmic}
\usepackage{multirow}
\usepackage{times}
\usepackage{epsfig}
\usepackage{graphicx}
\usepackage{amsmath}
\usepackage{amssymb}
\usepackage[font=small,labelfont=bf]{caption,subfig}
\usepackage{color}
\usepackage{nomencl}
\usepackage{relsize}
\usepackage{array}
\usepackage{url}
\usepackage[leftcaption,raggedright]{sidecap}
\usepackage{booktabs}
\usepackage{empheq}

\newtheorem{theorem}{Theorem}[section]

\newtheorem{lemma}[theorem]{Lemma}
\newtheorem{proposition}[theorem]{Proposition}
\newtheorem{corollary}[theorem]{Corollary}

\newenvironment{proof}[1][Proof]{\begin{trivlist}
\item[\hskip \labelsep {\bfseries #1}]}{\end{trivlist}}

\DeclareMathOperator*{\argmin}{argmin}

\begin{document}

\begin{frontmatter}



\title{Active Contour with A Tangential Component}


\author[A]{{Junyan~Wang}\corref{cor_me}}\author[A]{Kap~Luk~Chan}
\address[A]{School of Electrical and Electronic Engineering, Nanyang Technological University,\\
 Singapore 639798}

\cortext[cor_me]{Corresponding author. E-mail: wa0009an@ ntu.edu.sg, Tel: +6596409523}

\begin{abstract}
Conventional edge-based active contours often require the normal component of an edge indicator function on the optimal contours to approximate zero, while the tangential component can still be significant. In real images, the full gradients of the edge indicator function along the object boundaries are often small. Hence, the curve evolution of edge-based active contours can terminate early before converging to the object boundaries with a careless contour initialization. We propose a novel Geodesic Snakes (GeoSnakes) active contour that requires the full gradients of the edge indicator to vanish at the optimal solution. Besides, the conventional curve evolution approach for minimizing active contour energy cannot fully solve the Euler-Lagrange (EL) equation of our GeoSnakes active contour, causing a Pseudo Stationary Phenomenon (PSP). To address the PSP problem, we propose an auxiliary curve evolution equation, named the equilibrium flow (EF) equation. Based on the EF and the conventional curve evolution, we obtain a solution to the full EL equation of GeoSnakes active contour. Experimental results validate the proposed geometrical interpretation of the early termination problem, and they also show that the proposed method overcomes the problem.
\end{abstract}


\end{frontmatter}


\input{Intro2}
\input{Bak}
\input{Prob_GAC}
\input{Others}

\appendix
\section{Proofs}\label{APP}
\subsection{Proof of Proposition \ref{PROP:EF}}
\begin{proof}
Since $\langle \vec{F}(C^*), \vec{T}(C^*) \rangle=0$, we can
directly obtain
\begin{equation}
\vec{N}(C^*)={\vec{F}(C^*)\over \|\vec{F}(C^*)\|}
\end{equation}
or,
\begin{equation}
\vec{F}(C^*) = \vec{0}
\end{equation}
which is one of the definitions of level set, and this completes the proof.
\end{proof}
\subsection{Proof of Proposition \ref{PROP:EF2}}
\begin{proof}
Substituting (\ref{EQ:EF}) and ${dp\over dt}$ into ${dC\over dt}$, we obtain the following.

\begin{equation}
\begin{split}
{dC\over dt} &= {\partial C\over \partial t} + {\partial C\over\partial p}{dp\over dt}\\
             &= \langle \vec{F}, \vec{T} \rangle\vec{N} - \langle \vec{F}, \vec{N} \rangle\vec{T}\\
             &= \langle \vec{F}, \mathcal{R}^T\vec{N} \rangle\vec{N} + \langle \vec{F}, \mathcal{R}^T\vec{T} \rangle\vec{T}\\
             &= \langle \mathcal{R}\vec{F}, \vec{N} \rangle\vec{N} + \langle \mathcal{R}\vec{F}, \vec{T}\rangle\vec{T}\\
             &= -\mathcal{R}\nabla g(p,t)
\end{split}
\end{equation}
where $\mathcal{R}$ is a $90^o$ rotation matrix of size $2\times2$. Thus, taking derivative of $g$ w.r.t. $t$ we obtain the following.
\begin{equation}
{dg\over dt} = \nabla g(p,t){dC\over dt} = -\langle\nabla g(p,t),\mathcal{R}\nabla g(p,t)\rangle = 0
\end{equation}
which completes the proof.
\end{proof}




\bibliographystyle{elsarticle-num}
\bibliography{LevelSetActiveContours,MRFseg}







\end{document}

%% file: Intro2.tex
\section{Introduction}
Energy minimization provides a principled framework for various fundamental computer vision problems. Active contour was proposed for object segmentation based on energy minimization. The essential idea of the active contour is to model the object boundaries by the contour curves that minimize the functional energy which measures the error of boundary detection. The active contour has been adopted in many application domains of computer vision, such as surveillance video analysis \cite{GACTracking} \cite{Schoenemann10CombSolShapeSegTrack} and medical image analysis \cite{Yezzi97GAC_J} \cite{Leventon00Statisticalshape} etc.

There are both edge-based active contours \cite{Malladi95ShMo} \cite{caselles97GAC} \cite{Yezzi97GAC_J} \cite{Kimmel03EdgeInt} \cite{Corsaro06Covolume} and region-based active contours \cite{Zhusongchun96RegComp} \cite{ChanVese01ActiveCon} \cite{Baris06GPAC} \cite{Brox2009InterpMumfordShah}. There are also some recent attempts on improving the region based methods \cite{Li08LBF}, and improving the edge based active contours \cite{GVFGAC04,Corsaro06Covolume,Xie08MAC}. More recently, some efforts have been devoted to convex relaxations and global optimization of region based active contour models such as \cite{Strekalovskiy-et-al-cvpr12ConvexVectorMumfordShah}. However, it is arguable that neither the region-based nor the edge-based model is superior to the other in general. There are also recent efforts on integrating the region based and edge based methods \cite{Sagiv06Texture,Lankton2008localizing}. Generally speaking, the edge-based active contours are capable of achieving more accurate boundary extraction comparing with region-based active contours, but they generally require careful initializations.

This paper revisits a classic problem with edge-based active contours and provides some new insights to the problem. In the literature, e.g. \cite{Balloon91}, \cite{Xu98GVF}, \cite{GVFGAC04}, \cite{LiSnake05Split} and \cite{Xie08MAC}, the problem is often stated as: \emph{when the curve is initialized relatively arbitrarily, the active contour can stop early and some part of the converged curve can still be far from the boundaries of the objects of interest}. We investigate the cause of the early termination of curve evolution in general edge-based active contours. We observe that the full gradients of the edge indicator function along the object boundaries are often small. However, conventional edge-based active contours, such as the Geodesic Active Contour (GAC), often only require the normal component of the edge indicator function on the optimal contours to approximate zero, while the tangential component can be still significant. Based on this observation, we propose a novel active contour model: the \emph{Geodesic Snakes} (GeoSnakes) active contour model. The derived Euler-Lagrange (EL) equation of the GeoSnakes model requires the full gradients of the edge indicator to be close to zero on the optimal contours. However, the conventional curve evolution method does not fully solve the EL equation of the GeoSnakes model, although the curve evolution can still converge stationarily. This phenomenon is named the Pseudo Stationary Phenomenon (PSP). To address the PSP, we propose an auxiliary curve evolution equation, named the Equilibrium Flow (EF). The full EL equation of the GeoSnakes for boundary extraction is solved by alternating the Equilibrium Flow and the conventional gradient descent curve evolution.

From our point of view, our contributions are as follows.
{{\begin{enumerate}
\item We elucidate importance of the tangential component of the gradient of the edge indicator along the boundary curve for boundary locating. This observation contradicts the conventional view that the tangential component is merely a useless reparameterization force.
\item We obtain a new active contour model of which the EL equation can be satisfied by a smooth contour curve if both the normal and tangential components of the gradient of the edge indicator along the contour approach zero.
\item We obtain a curve evolution method to solve the EL equations containing both the normal and tangential components.
\end{enumerate}}

The rest of the paper is organized as follows. In Section \ref{SEC:BG}, we introduce the geodesic active contour and the works relevant to the problem of early termination of curve evolution. In Section \ref{SEC:Prob-Solve}, we present our problem statement followed by our formulation and solution. In Section \ref{SEC:EXP}, experiments are conducted to evaluate our proposed method and to also valid our theory. In Section \ref{SEC:CON}, we conclude the paper and present some further possibilities beyond this paper. This work is based on the preliminary observations presented at BMVC 2008 \cite{WangChanWang08PSP}.


%% file: Bak.tex
\section{Background}\label{SEC:BG}
\subsection{The Geodesic Active Contour}


Our discussions are mostly based on the general Geodesic Active Contour (GAC) model. The energy functional of GAC is as follows.
\begin{equation}\label{EQ:GAC1}
\begin{split}
C^* &= \arg\min\limits_{C}~\mathcal{L(C)}\\
\mathcal{L(C)}&=  {\mathlarger{\mathlarger\int}}_{0}^{1}\left[{C_p}^T\left(\begin{array}{cc}
                                            g(\nabla I) & 0 \\
                                            0 & g(\nabla I) \\
                                          \end{array}\right){C_p}\right]^{1\over2}dp\\
                                            &=\int_0^1g\|C_p\|dp=\int_{C}gds
\end{split}
\end{equation}
where $ds=\|C_p\|dp$ is the arclength parameterization and the edge indicator function $g$ can be defined below.
\begin{equation}\label{EQ:EdgeInd}
g(\nabla I)={1\over{1+\|G_{\sigma}\ast\nabla I\|^q}}
\end{equation}
in which $G_{\sigma}$ is a Gaussian filter of width $\sigma$, and we may assume $q = 2$. The justifications of this choice of edge indicator can be found in \cite{Malladi95ShMo} \cite{Caselles93GeoModel} \cite{caselles97GAC}.

We rewrite the Euler-Lagrange equation of GAC as follows.
\begin{equation}\label{EQ:GACEL}
{\delta \mathcal{L}\over\delta C}=\langle\nabla g,\vec{N}\rangle\vec{N} - g\kappa\vec{N}= 0
\end{equation}
where $\kappa$ is the curvature of curve $C$, and $\vec{N}$ is the normal vector of $C$. Empirically, the smoothing term $g\kappa$ is close to zero on a converged smooth curve. Hence, $\langle\nabla g,\vec{N}\rangle$ is also close to zero on the curve.

\subsection{Previous works on the early termination problem in curve evolution}
This subsection will review some pioneering works on this early termination problem. Their methods have been proven effective for addressing this problem to various degrees. Our approach differs from them in that we aim to not only remedy the problem but also investigate the cause of the problem.

The early termination of local gradient based active contours was first reported in \cite{Balloon91}. He proposed the Balloon term to push the curve either to shrink or to expand before reaching the object boundary. Generally, the Balloons can be applied when there is little attraction sensed by the evolving curve. Xu and Prince \cite{Xu98GVF} observed that the Snakes might not converge to concave boundary even when there is large attraction. We quote their statement below.

\begin{quote}
\emph{Although the external forces \footnote{the gradient vectors} correctly point toward the object boundary, within the boundary concavity the forces point horizontally in opposite directions. Therefore, the active contour is pulled apart toward each of the ¡°fingers¡± of the U-shape, but not made to progress downward into the concavity}
\end{quote}

\begin{figure}
  \centering
  \includegraphics[width=0.4\textwidth]{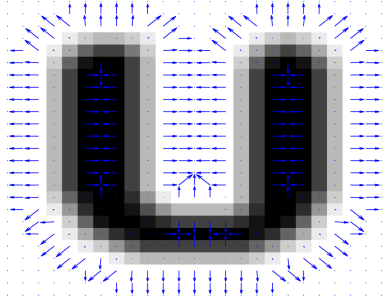}\\
  \caption{The U-shape and the opposite gradient vectors}\label{Fig:Ushape}
\end{figure}

We also reproduce a U-shape as well as the associated normalized velocity field in Figure \ref{Fig:Ushape} to visualize the statement above. Based on this observation, Xu and Prince \cite{Xu98GVF} also proposed the Gradient Vector Flow (GVF) to extend and smooth the gradient field in order to address the problem for extracting objects with moderately concave boundaries. This method is tested to be effective for moderate boundary concavity.

Paragios et al. \cite{GVFGAC04} applied the GVF to GAC with the level set method for extracting multiple objects. However, they discovered a problem which we quote below.
\begin{quote}
\emph{...the proposed flow does not perform propagation when the NGVF\footnote{normalized gradient vector flow} is close to orthogonal to the inward normal ... propagation will not take place, as well as change of the topology, even if they are supported by the level set technique.}
\end{quote}

We can compare this statement above with the observations by Xu and Prince \cite{Xu98GVF} to conclude that both the original Snakes and the level set based active contours suffer from the same problem. Paragios and his colleagues propose the adaptive balloon to force the active contour to evolve when the normal projection of the gradient is close to zero. This idea works for images of relatively simple topology, as shown in our experiments. For images of complex topology, the contour might be able to avoid converging to the positions where the normal projection of the gradient is close to zero, but it can oscillate near these positions since the Adaptive Balloon is formulated to be in the same direction of the external force.

Li et al. \cite{LiSnake05Split} proposed to pre-segment the image before curve evolution. The segmentation is based on the geometry of the GVF. By segmenting the image, the closed curve can also be cut into smaller ones. However, the segmentation of GVF is not in the active contour framework. The cause of the early termination problem of the curve evolution of general edge-based active contours was not addressed there.

More recently, Xie and Mirmehdi \cite{Xie08MAC} formulated a novel edge-based curve evolution equation motivated by the mathematical formulations of magnetostatic / Lorentz force, as an alternative to the conventional edge-based active contours. The curve evolution model was named the Magnetostatic Active Contour (MAC). In that paper, the authors conjectured that the early termination of gradient based curve evolution can be due to the undesired critical points, such as saddle points or maxima. The authors also demonstrated that MAC can surpass the undesired critical points and converge to boundaries with severe concavities or boundaries of multiple objects for many images while having less constraints on the initializations. However, MAC is a curve evolution framework, and the curve evolution does not necessarily minimize an energy functional. Hence, MAC is not formulated under energy minimization framework. Besides, MAC requires the edge detection to produce little spurious edge as a preprocessing step. The detected edge helps producing an indicator of region homogeneity, and the resultant curve evolution behaves like region based active contours as observed in the experiments presented in this paper.

This problem with GAC may be considered as a problem of local optimality of the curve evolution based method, but the globally optimal solution of GAC can be a dot, because GAC tries to find the contour having minimal weighted contour length.

In the experiments presented later, we will show that our method compares favorably to the related methods for object segmentation in images containing relatively complex structures.

%% file: Prob_GAC.tex

\section{Active contour with a tangential component}\label{SEC:Prob-Solve}

\subsection{An interpretation of the early termination problem}
\begin{figure}
\centering
  \subfloat[]{\includegraphics[width=0.2\textwidth,height=0.8in]{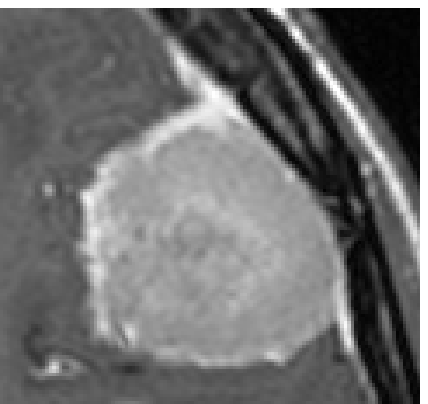}}~~
  \subfloat[]{\includegraphics[width=0.3\textwidth]{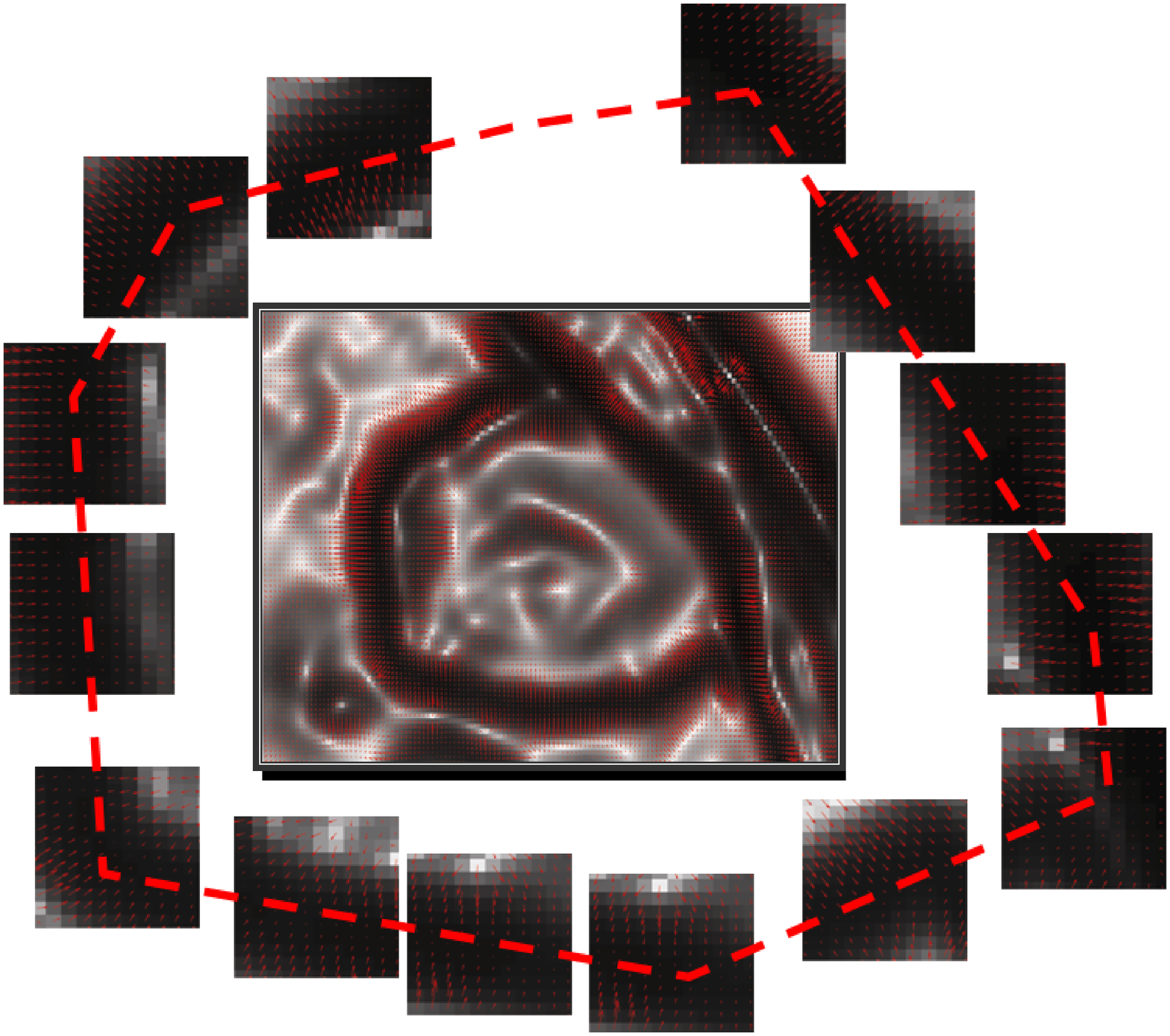}}
  \caption{The full gradients of the edge indicator near the object boundary are close to zero. (a) is an image containing an object of interest. In (b), the center is the edge indicator function for (a). The intensity of the edge indicator is visualized by the darkness in figure (b). The surrounding patches in (b) show the gradient vector field of the edge indicator along the object boundary (dashed curve).}\label{FIG:two_demoVF}
\end{figure}

The object boundaries perceived by human are located at the peaks of the magnitude of image gradients according to the theory of edge detection \cite{MarrHildreth80EdgeDet}. Based on the magnitude of image gradients, an edge indicator function can be formulated as in Eq. (\ref{EQ:EdgeInd}). Variants of the edge indicator functions have now been commonly used in the formulations of active contours, e.g. \cite{Kimmel03EdgeInt}. A prototypical edge-based active contour is the GAC. In GAC model, the boundary is considered as the contour corresponding to the minimal GAC energy. The GAC energy is small, if the edge indication on the contour is strong. The Euler-Lagrange (EL) equation for minimizing the GAC energy requires the gradients of the edge indicator in the normal direction of the optimal curve to approach zero. However, this requirement is insufficient. It has been observed that the curve evolution for solving the EL equation often converges at the non-boundary positions where the tangential component is still significant \cite{Xu98GVF} \cite{GVFGAC04}. In fact, the \emph{full} gradients of the edge indicator along the boundaries are often small, i.e. both the normal and tangential components of the gradients of the edge indicator along the boundary are close to zero, such as Figure \ref{FIG:two_demoVF}. In practice, the gradient field of the edge indicator functions of images containing noise or spurious edges are topologically complex. In such gradient fields, we can easily find the non-boundary curves on which the normal component of the gradients is zero but the tangential is non-zero. Some more examples can be found in the experiment section. Note that the assumption of small gradients along object boundary is not always valid. For example, the inhomogeneous region may lead to non-zero gradients on the boundary. The intensity inhomogeneity in medical images can often be corrected by preprocessing \cite{Vovk07CorInhomo}. 

\begin{figure*}[hbt]
\centering
\subfloat[]{\includegraphics[width=0.4\textwidth]{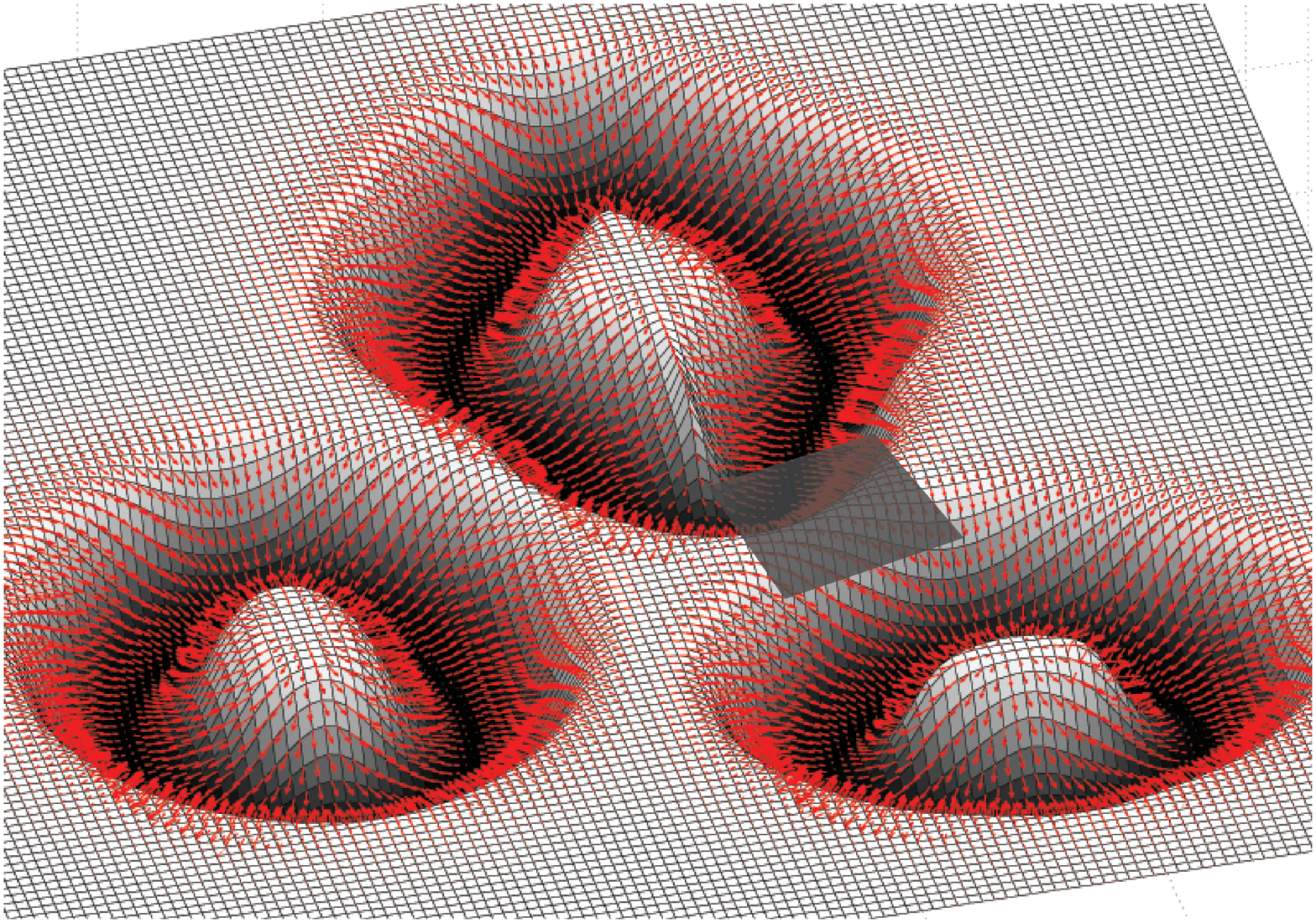}}\hspace{30pt} \subfloat[]{\includegraphics[width=0.4\textwidth]{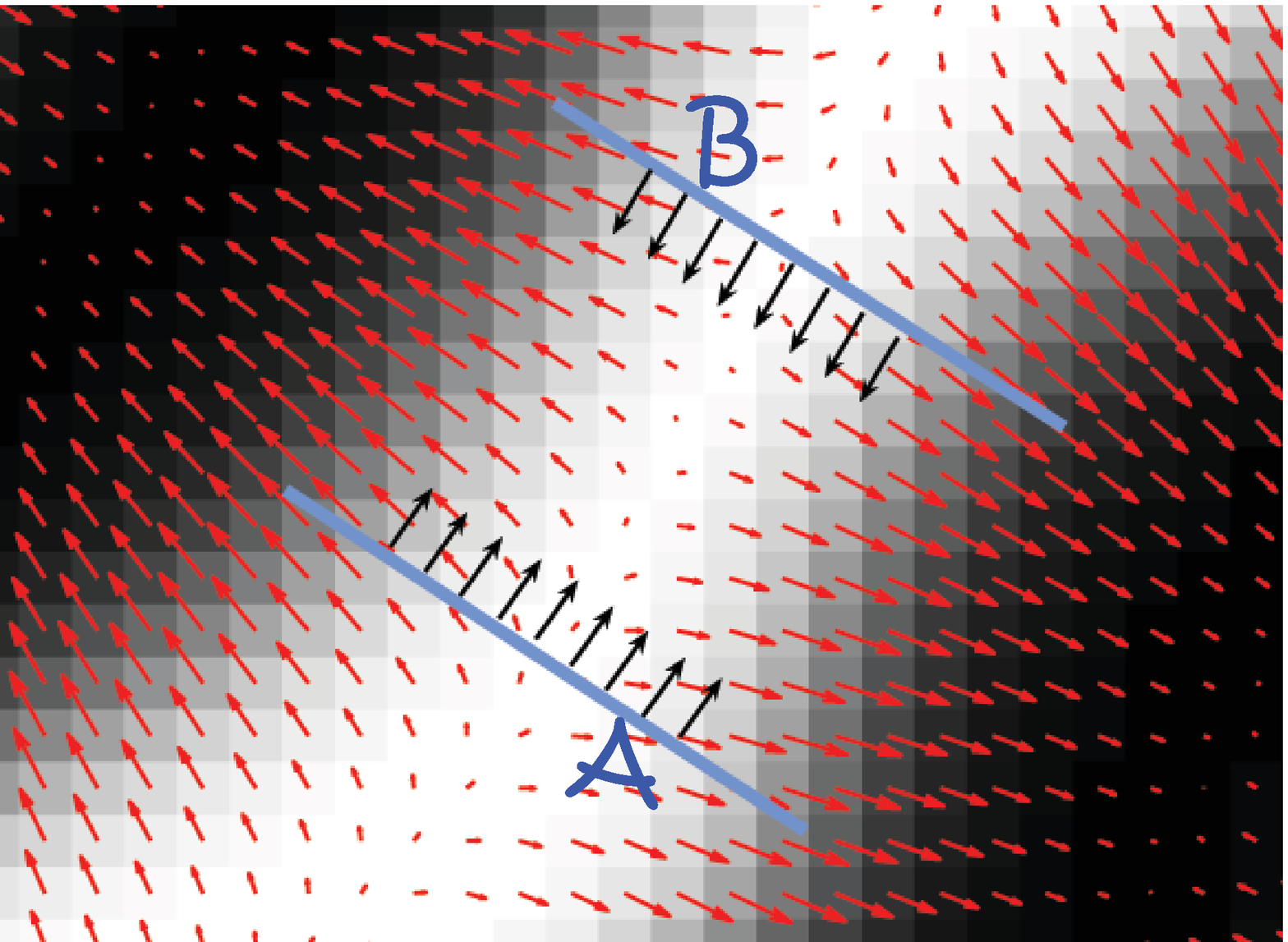}}
\caption{The curve stops (or oscillates) when the gradient is nearly orthogonal to the inward normal of the curve. (a) is a 3D visualization of a typical edge indication function overlayed by the negative gradient field; (b) demonstrates the curve evolution in the negative gradient field within the gray-shaded box region in (a). See text for details.}\label{Fig:PSPshow}
\end{figure*}

Without requiring the tangential component to be close to zero, we should not expect to obtain correct object boundaries. In practice, this leads to the early termination problem. Figure \ref{Fig:PSPshow} shows a possible situation of early termination of the curve evolution in GAC. Let the curve segments at $A$ and $B$ be the same curve segment from two successive iterations of the curve evolution in a vector field. The black arrows in Figure \ref{Fig:PSPshow} (b) denote the actual attraction velocity on the contour curves. The attraction velocity is the projection of the vector field, visualized by the red arrows, in the normal direction of the curve. As the curve evolves, the curve at $A$ at first moves towards position $B$ in the first iteration. Then, in the second iteration, the curve at $B$ will move towards $A$. By using a sufficiently small time step, the curve will finally stop at some place where the gradients are nearly orthogonal to the inward/outward normal, but not necessarily zero. The curve evolution terminates too early since the curve converges to the place in-between the boundaries where the normal components vanishes but not the tangential ones.

\subsection{The GeoSnakes and the Pseudo Stationary Phenomenon}
In the following, we propose a new active contour model that requires the full gradients of an edge indicator to approach zero at the optimal solution. The model is given as follows.
\begin{equation}\label{EQ:GeoSnakes}
C^* = \argmin\limits_{C}~\mathcal{L}(C) = \int_C(\ln\circ g\circ C) dp+\int_C\|C_p\|dp
\end{equation}
where $\circ$ is the function composition operator, $\ln\circ g\circ C=\ln(g(C))$ is the composition of the three functions, $g$ is the edge indicator function defined in Eq. (\ref{EQ:EdgeInd}). The corresponding EL equation is the following.
\begin{equation}\label{EQ:SnF2}
\begin{split}
&{\delta\mathcal{L}(C)\over\delta C}={1\over g}\nabla g-\kappa\vec{N} = 0
\Leftrightarrow \nabla g-g\kappa\vec{N} = 0\\
&\Leftrightarrow \langle\nabla g,\vec{T}\rangle\vec{T}+\langle\nabla g,\vec{N}\rangle\vec{N}-g\kappa\vec{N} = 0\\
\end{split}
\end{equation}
where $\vec{T}$ is the tangential of the contour curve and $\vec{N}$ is the normal. The first equivalence holds since $g$ is guaranteed positive. Comparing (\ref{EQ:SnF2}) with (\ref{EQ:GACEL}), we can observe that (\ref{EQ:SnF2}) contains an additional tangential component, which is also the tangential component of the gradient of the edge indicator.

There exists geometric and parametric active contours. A discussion regarding the relationship between the geometric active contours and the parametric active contours is in \cite{para&geoAC00}. The major difference between the two kinds of active contours is that the curve evolution equation of parametric active contours may contain a tangential component whereas the other one contains only the normal component. GAC is a typical geometric active contour. A typical parametric active contour containing tangential component is the original Snakes active contour \cite{kass88snakes} \cite{para&geoAC00}. Note that the normal component in the EL equation of GeoSnakes is accidentally the same as that of the GAC. The proposed model is named the \emph{Geodesic Snakes} (\emph{GeoSnakes}) since it is related to both the GAC and Snakes.

Now let us turn to the solution to the GeoSnakes model. The conventional solution to active contour is the gradient descent curve evolution. The corresponding curve evolution equation of GeoSnakes is as follows.
\begin{equation}
\partial_t C=g\kappa\vec{N}-\langle\nabla g,\vec{T}\rangle\vec{T}-\langle\nabla g,\vec{N}\rangle\vec{N}
\end{equation}

Unfortunately, it is commonly believed that the tangential component in the curve evolution equations will only automatically reparameterize the contour curve, and the tangential component is conventionally omitted in the implementation of the curve evolution. This behavior can be explained with the help of the Lemma of curve evolution stated in \cite{Epstein87CS}.
\begin{lemma}\label{LM:CE}
Given the closed curve $C(\tilde{p},t)$ parameterized by arbitrary $\tilde{p}\in \tilde{B}$ at an artificial time $t$ with the normal $\vec{N}$, the tangent $\vec{T}$ of the curve, and given the geometric flow of a curve evolution by
{\normalsize\begin{equation}\label{EQ:CE1}
{\partial_t C(\tilde{p},t)}=\alpha(\tilde{p},t) {\vec{T}}(C(\tilde{p},t)) +
\beta(\tilde{p},t) {\vec{N}}(C(\tilde{p},t))
\end{equation}}
If $\beta$ does not depend on the parametrization, meaning that $\beta$ is a geometric intrinsic characteristic of the curve, then the image of $C(\tilde{p},t)$ that satisfies Equation(\ref{EQ:CE1}) is identical to the image of the family of curves $C(p,t)$, parameterized by $p\in B$, that satisfies
\begin{subequations}\label{EQ:CE2}
\begin{empheq}[left=\empheqlbrace]{align}
&{\partial_t C(p,t)}= \beta(p,t)
{\vec{N}}(p,t)
\\
&{\partial p\over\partial t} = -{\alpha\over\|C_p\|}
\end{empheq}
\end{subequations}
\end{lemma}
The above lemma states that the geometry of the curve evolution only depends on the velocity in the normal direction of the curve, although the reparametrization by ${\partial p\over\partial t}$ may last forever due to the possibly non-vanishing $\alpha$. A brief proof of this lemma can be found in the book of \cite{SapiroBook}.

The omission of the tangential component in the original EL equation can also be seen in the level set method for implementing general curve evolution.
Recall the level set method for curve evolution \cite{OsherSethian88Fronts}.
\begin{equation}\label{EQ:ODELS3}
\begin{split}
\left.{\partial_t \Phi}\right|_{\Phi=0} &= \left.\langle\nabla\Phi,{\partial_t C}\rangle\right|_{\Phi=0}\\
&= \left.\langle\nabla\Phi,\beta\vec{N}+\alpha\vec{T}\rangle\right|_{\Phi=0},~\hbox{by Equation(\ref{EQ:CE1})}\\
&= \beta \|\nabla\Phi\|\Big|_{\Phi=0}\\
&\hbox{~by setting~} \vec{N} = {\nabla\Phi\over\|\nabla\Phi\|}\Rightarrow \langle\nabla\Phi,\vec{T}\rangle=0
\end{split}
\end{equation}
%

All the above says that the curve evolution is independent of the tangential velocity on the closed curve. It also implies that the converged solution of Eq. (\ref{EQ:CE2}) may not necessarily be the stationary solution of Eq. (\ref{EQ:CE1}). To be specific, there may exists a non-vanishing reparametrization velocity, i.e. the tangential velocity, on the converged curve. This can be ascertained by rephrasing Lemma \ref{LM:CE} to be the following.
\begin{corollary}[Pseudo Stationary Phenomenon]\label{CO:CE}
The converged curve of Eq. (\ref{EQ:CE2}), denoted by $C(\tilde{p},\infty)$ which satisfies
\begin{equation}
{\partial_t C(\tilde{p},\infty)}=
\beta(C(\tilde{p},\infty)) \vec{N}(C(\tilde{p},\infty))=\vec{0}
\end{equation}
can lead to
\begin{equation}
\begin{split}
{\partial_t C(p,\infty)} &= \alpha(C(p,\infty))
\vec{T}(C(p,\infty))\\
&\hspace{10pt}+ \beta(C(p,\infty))\vec{N}(C(p,\infty))\\
&\neq\vec{0}
\end{split}
\end{equation}
where $p=p(\tilde{p},t)$, $\alpha=-\|C_p\|{\partial p\over\partial t}$, ${\partial p\over\partial t}$ is the parametrization and $\alpha(C(p,\infty))$ can be large. $C(p,\infty)$ is the converged curve.
\end{corollary}
The above property is termed the Pseudo Stationary Phenomenon (PSP), since the stationarily converged curve may not be a solution of the original EL equation such as (\ref{EQ:SnF2}). Interestingly, the early termination of GAC is corresponded to the PSP of GeoSnakes.

\subsection{The Equilibrium Flow and the Alternating Curve Evolution}
On the one hand, the curve evolutions with only the normal component is insufficient for boundary location and the tangential component should also be considered. On the other hand, the curve evolutions can only solve for the normal component of the curve evolution equation, even when the curve evolution equation contains a tangential component. Thus, we find ourselves in a dilemma.

In the following, we propose a solution to the curve evolution equation with a tangential component, i.e., a solution to the PSP problem. The curve evolution equation concerned can be written in the most general form as follows.
\begin{equation}\label{EQ:GDF2}
\begin{split}
{\partial_t C} &= \beta(p,t) \vec{N}(p,t)+\langle \vec{F}(p,t),\vec{T}(p,t)\rangle \vec{T}(p,t)\\
\end{split}
\end{equation}
where $\vec{F}$ is a vector field. The objective is to ensure both the normal component, $\beta$, and the tangential component, $\langle \vec{F}(p,t),\vec{T}(p,t)\rangle$, to vanish when the curve evolution stops. Since the curve evolution is independent of the tangential velocity, the tangential component plays no role in the whole process of curve evolution. Therefore, an auxiliary flow termed the \emph{Equilibrium Flow} (EF) is proposed to ensure the tangential velocity to be zero when the curve evolution stops. The EF equation is defined as follows.
\begin{equation}\label{EQ:EF}
{\partial_t C(p,t)}=\langle \vec{F}(p,t), \vec{T}(p,t) \rangle\vec{N}(p,t)
\end{equation}
This equations is named the Equilibrium Flow due to the following property.

\begin{proposition}\label{PROP:EF}
The stationary solution $C^*$ of the Eq. (\ref{EQ:EF}) is either on stationary points in $\vec{F}$ or on level sets of the potential $g$ corresponding to the gradient field $\vec{F}$.
\end{proposition}
The proof is given in Appendix.

Another more interesting property of EF is the following.
\begin{proposition}\label{PROP:EF2}
The curve evolution of (\ref{EQ:EF}) and the reparametrization ${dp\over dt}=-{\langle \vec{F}(p,t), \vec{N}(p,t) \rangle\over\|C_p\|}$ where $\vec{F}=-\nabla g$, leads to $g(C(p,t))=g(C(p,t+\tau))$ for any $\tau$ before termination of the curve evolution.
\end{proposition}
The proof is given in Appendix. This proposition implies that the function $g$ for a fixed $p$ would not change during the curve evolution driven by EF. With a proper contour parametrization, the curve evolution driven by EF can solve for the tangential component of EL while the energy on the contour will not be changed.

By using this EF flow, we obtain a solution to the EL equation involving both the normal and tangential components, which is formally stated below.
\begin{proposition}\label{Prop_opt}
Given an arbitrary initial curve $C_0$, the convergent curve $C(p,\infty)$  driven by the following system of flows is the stationary solution of the original curve evolution
(\ref{EQ:GDF2}) as well as the corresponding EL equation.
\begin{subequations}\label{EQS:ALT}
\begin{align}
&{\partial_t C_1^k} = \beta(p,t)\vec{N},~~C_1^k(p,0)=C_2^{k-1}(p,\infty) \label{EQ:ALT_a} \\
&{\partial_t C_2^k} = \langle \vec{F}, \vec{T} \rangle \vec{N},~~C_2^k(p,0)=C_1^k(p,\infty)\label{EQ:ALT_b}
\end{align}
\begin{equation}
~\left\{\begin{split}
&k = 1,2,\ldots\\
&C_1^0(p,0) = C_2^0(p,\infty) = C_0\\
&C(\tau,\infty)=\lim\limits_{k\rightarrow\infty}C_1^k(\tau,\infty)=\lim\limits_{k\rightarrow\infty}C_2^k(\tau,\infty)
\end{split}\right.
\end{equation}
\end{subequations}
\end{proposition}
This claim is true since Eq. (\ref{EQ:GDF2}) can be satisfied only when both the $\beta$ and $\alpha$ are zero, and the curve evolutions (\ref{EQ:ALT_a}) and (\ref{EQ:ALT_b}) defines a process to ensure the $\beta$ and $\alpha$ to be zero respectively. This proposition tells that the alternation of the two curve evolutions can provide a solution to the original EL equation containing both normal and tangential component, which is previously unknown.

The derived new curve evolution framework can have the following interpretation. Starting from an initial contour curve, the curve evolution of (\ref{EQ:ALT_a}) may converge to a pseudo stationary position, where the PSP occurs. Then the curve evolution of (\ref{EQ:ALT_b}) can help the curve evolution to escape from the pseudo stationary position. At the convergence of (\ref{EQ:ALT_b}), the value of the edge indicator at every contour point of $C(p)$ can be guaranteed unchanged (by Proposition \ref{PROP:EF2}), and the curve evolution of (\ref{EQ:ALT_a}) has been reactivated because $\langle\nabla g,\vec{N}\rangle$ tends to be large according to Proposition \ref{PROP:EF}. The process continues until the system of equations in (\ref{EQS:ALT}) converges.

The curve evolution of (\ref{EQ:ALT_b}) cannot be directly implemented in a level set framework in which the tangent $\vec{T}$ cannot be conveniently used. Fortunately, in 2D case, there is a useful relation that $\vec{N}=\mathcal{R}\vec{T}$, where $\mathcal{R}$ is a matrix for $\pm 90$ degree rotation. Therefore, the EF can be rewritten as follows.
\begin{equation}\label{EQ:REF}
\begin{split}
{\partial_t C(p,t)} &= \langle \vec{F},
\vec{T} \rangle \vec{N}\\
&=(\vec{F}^T
\vec{T})\vec{N}=(\vec{F}^T\mathcal{R}\vec{N} )\vec{N}\\
&=((\mathcal{R}^T\vec{F})^T\vec{N})\vec{N}=\langle\mathcal{R}^T\vec{F},\vec{N}\rangle\vec{N}
\end{split}
\end{equation}

\begin{figure}[t]
\centering
\subfloat{\includegraphics[width = 0.25\textwidth]{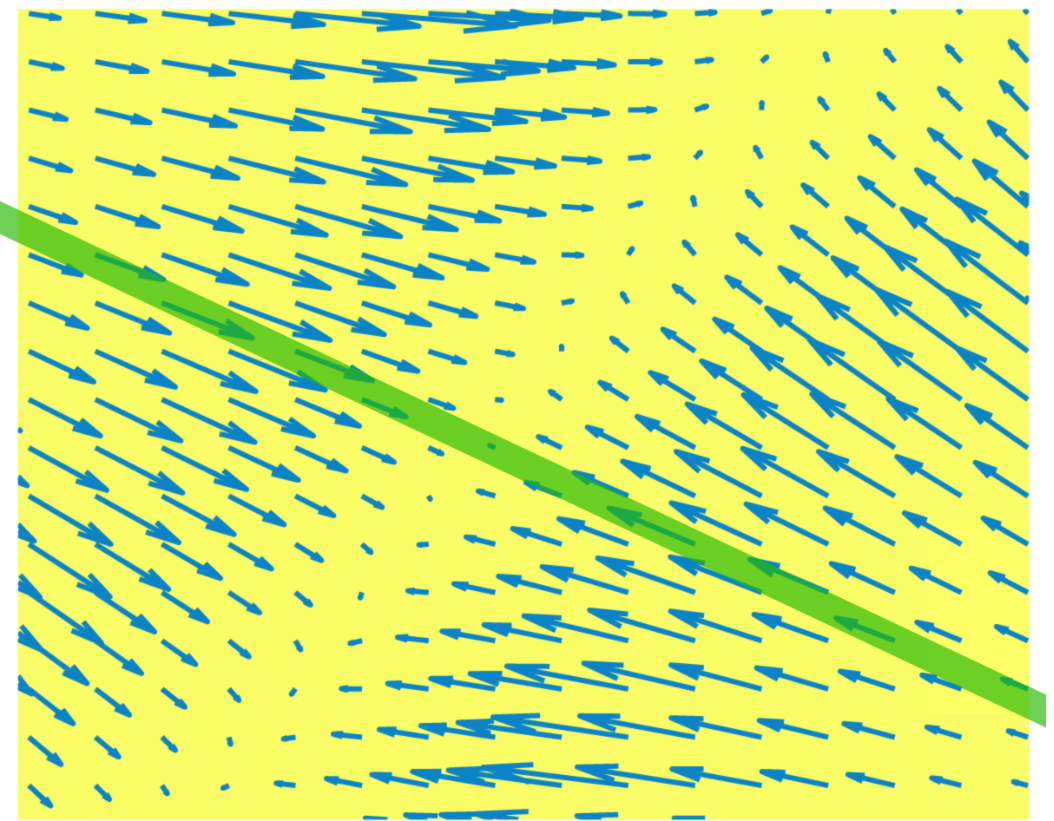}}
\subfloat{\includegraphics[width =0.25\textwidth]{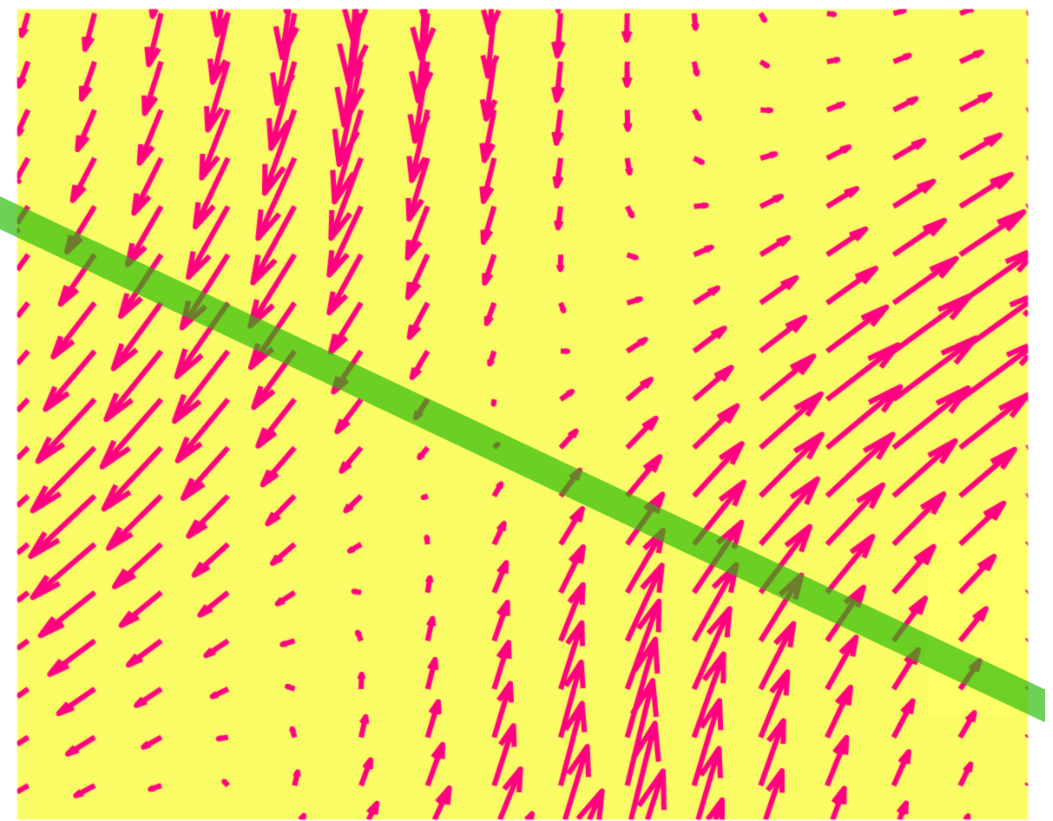}}\\
\subfloat{\includegraphics[width = 0.5\textwidth, height = 0.2\textwidth]{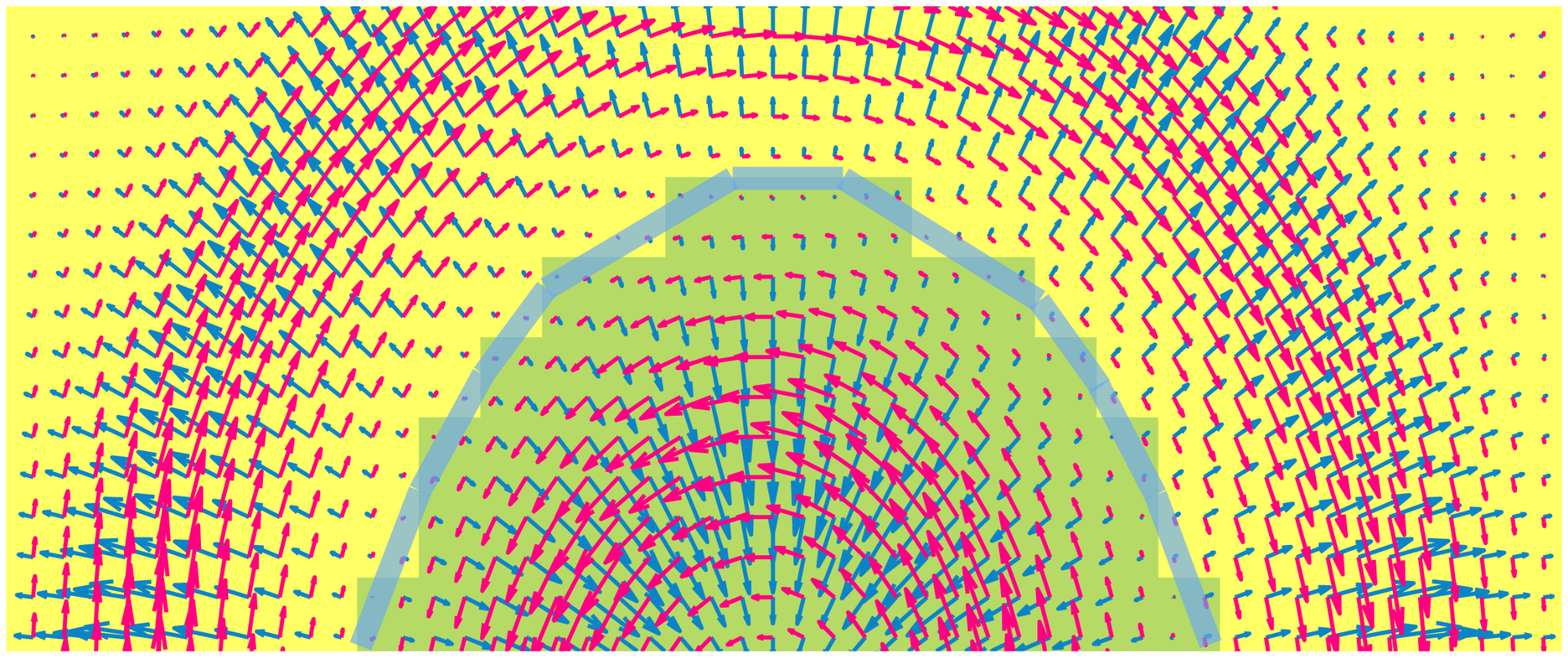}}
\caption{The practical effect of rotating gradient field. The top left is a typical gradient field, visualized by dark blue arrows, in which the green line segment cannot be moved by the local gradients due to the PSP. The top right is the vector field generated by the rotation, visualized by the arrows in red, in which the line segment can now be moved by this vector field. The bottom visualizes both the gradient field and the rotated vector field. It shows that the stationary positions, i.e., the light blue curve, remain stationary after the rotation}\label{Fig:RotGrad}
\end{figure}

%
The EF (\ref{EQ:ALT_b}) in the system Eqs. (\ref{EQ:ALT_a}) and (\ref{EQ:ALT_b}) can be understood as a repositioning process of the curve. The repositioning process can help the curve to escape from the pseudo stationary positions. The visualization of the idea of EF is presented in Figure \ref{Fig:RotGrad}. The rotation of gradient field can reactivate the curve evolution when it is trapped by PSP, while the true stationary positions remain stationary.

\subsection{The algorithm}
The problem we discussed previously is general, but we still focus on analyzing and solving problems with the GeoSnakes where $\beta=g\kappa-\langle\nabla g,\vec{N}\rangle$. We adopted the level set method \cite{OsherBookDynamic} for implementation. The pseudo code is shown in Algorithm \ref{Alg:GAC+EF}. Note that the maximum iteration is reached if the contour has little motion. The maximum cycle is chosen to be 3 in the experiment.
\renewcommand{\algorithmicrequire}{\textbf{Input:}}
\renewcommand{\algorithmicensure}{\textbf{Output:}}
\begin{algorithm}[!h]
\caption{Alternating Curve Evolutions}
\label{Alg:GAC+EF}
\begin{algorithmic}[1]
  \REQUIRE Input Image $I$, MaximumCycle, MaximumIteration
  \ENSURE $C_k$
  \STATE $I_{sm}\Leftarrow I$, by anisotropic smoothing
  \STATE $g \Leftarrow \nabla I_{sm}$, by Equation \ref{EQ:EdgeInd}
  \STATE $\vec{F}\Leftarrow -g$, by Gradient Vector Flow (GVF)
  \STATE $c=0$
  \STATE Initialize $C_0$
  \WHILE{$c\leq$MaximumCycle}
    \STATE $k = 0$
    \WHILE{$k\leq$MaximumIteration}
        \IF{$\hbox{mod(cycle},2)=0$}
            \STATE $C_{k+1}\Leftarrow C_k$ by the direct gradient decent flow of Eq. (\ref{EQ:ALT_a})
        \ELSE
            \STATE $C_{k+1}\Leftarrow C_k$ by the EF, i.e., Eq. (\ref{EQ:REF}).
        \ENDIF
        \STATE k = k + 1;
    \ENDWHILE
    \STATE c = c + 1;
  \ENDWHILE
\end{algorithmic}
\end{algorithm} 

%% file: Others.tex
\section{Experimental results}\label{SEC:EXP}
In this section, we present the experimental results that validate our interpretation and formulation for solving the early termination problem. The proposed method is also compared with other related methods to show the practical usefulness of this work.

\subsection{Experiment settings}
We describe the details of the parameter settings for implementation of the methods for preprocessing and comparison. The detailed implementation can be found in the original papers that are cited below.

\paragraph{Edge map\cite{Malladi95ShMo}\cite{caselles97GAC}} The $\sigma$ of the Gaussian kernel that is used to generate edge map $g$ in Eq. (\ref{EQ:EdgeInd}) is empirically chosen to be 3 for all the images;
\paragraph{GVF\cite{Xu98GVF}} The diffusion of the edge map $g$ is done by GVF as a preprocessing for all the gradient based methods that are evaluated. We compute the GVF by using the source code provided in the GVF web site\footnote{\url{http://iacl.ece.jhu.edu/projects/gvf/}}. We select the parameter $\mu=0.1$;
\paragraph{Balloon \cite{Balloon91}\cite{caselles97GAC}} The Balloon used for comparison is chosen to be positive to shrink the curve;
\paragraph{Adaptive Balloon\cite{GVFGAC01}\cite{GVFGAC04}} For comparison, we adopt the version in \cite{GVFGAC01}, which is simpler for analysis comparing to \cite{GVFGAC04}, while almost equivalent. In our implementation $\lambda=1$, $\beta=0.1$;
\paragraph{Chan-Vese model\cite{ChanVese01ActiveCon}} We set $\lambda_1,\lambda_2=1$, $\mu=0.1$, and $\upsilon=0$ which is a Balloon term. The $\delta_{\epsilon}$ and $H_{\epsilon}$ are chosen as in \cite{ChanVese01ActiveCon};
\paragraph{MAC \cite{Xie08MAC}} We use Canny edge detection as a preprocessing to specify the current along the boundaries of possible objects, and we set the $\alpha=0.5$ in the paper for all the images;
\paragraph{Approximated Dirac delta} The approximated delta function $\delta_{\epsilon}$ for level set implementation, excluding Chan-Vese model, are chosen to be as,
\begin{equation*}
\delta_{1}(x)=\left\{\begin{array}{ll}                                0, & |x| > 1 \\                                {1\over2} (1+\cos\left({\pi x}\right)), & |x|\leq1                              \end{array}\right.;
\end{equation*}
\paragraph{Re-initialization} We adopted the $1^{st}$ order accurate Essentially Non-Oscillatory (ENO1) based on Baris Sumengen's MATLAB toolbox of the Level Set method \footnote{\url{http://barissumengen.com/level_set_methods/}};

The Balloon is chosen to be $1$. The algorithms are implemented in MATLAB. The coding of the active contours is largely inspired by Chunming Li's implementation\footnote{\url{http://www.engr.uconn.edu/~cmli/}}.

\subsection{Segmentation results}
\begin{figure*}
  \subfloat[]{\includegraphics[width=0.16\textwidth]{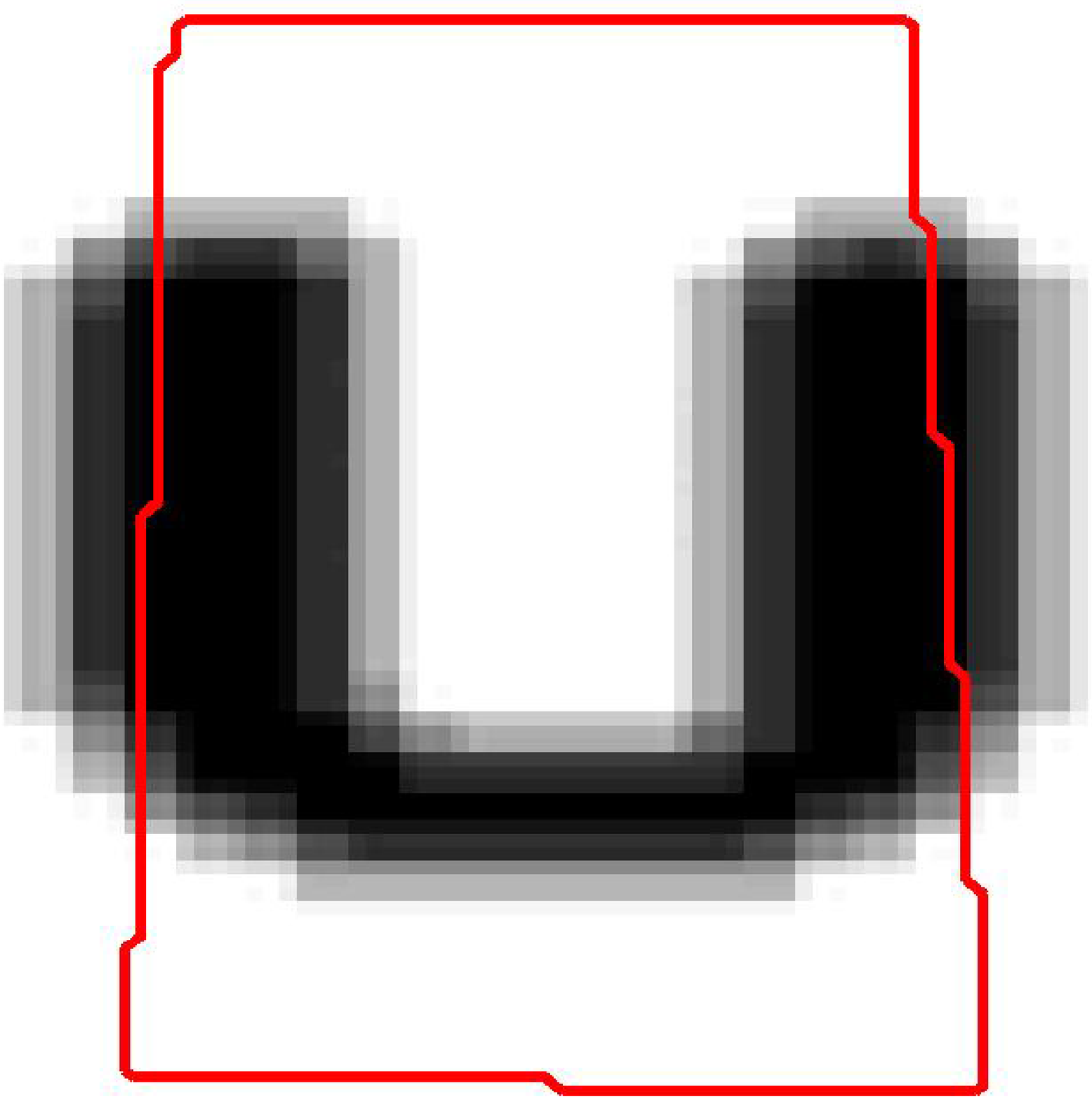}}
  \subfloat[]{\includegraphics[width=0.16\textwidth]{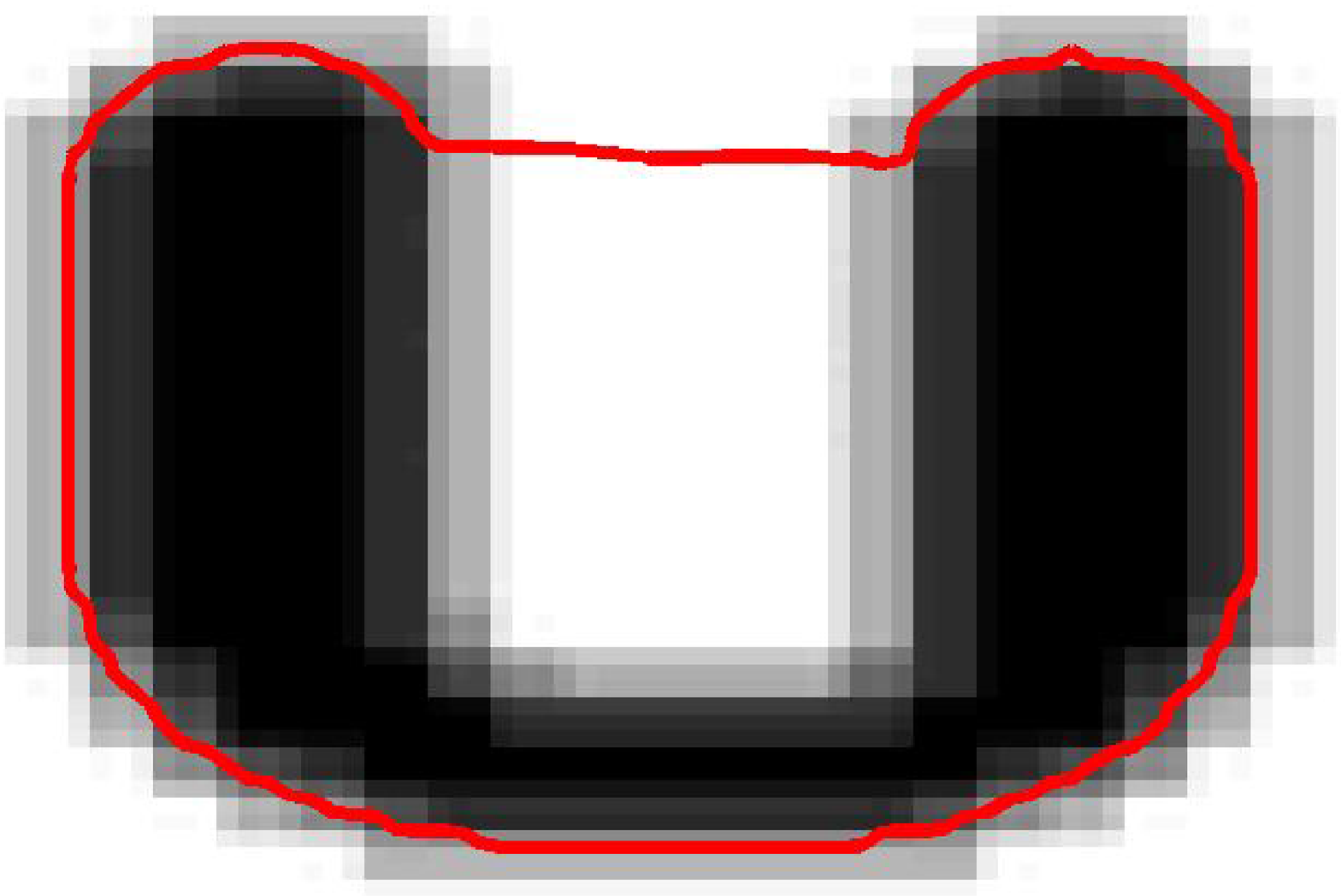}}
  \subfloat[]{\includegraphics[width=0.16\textwidth]{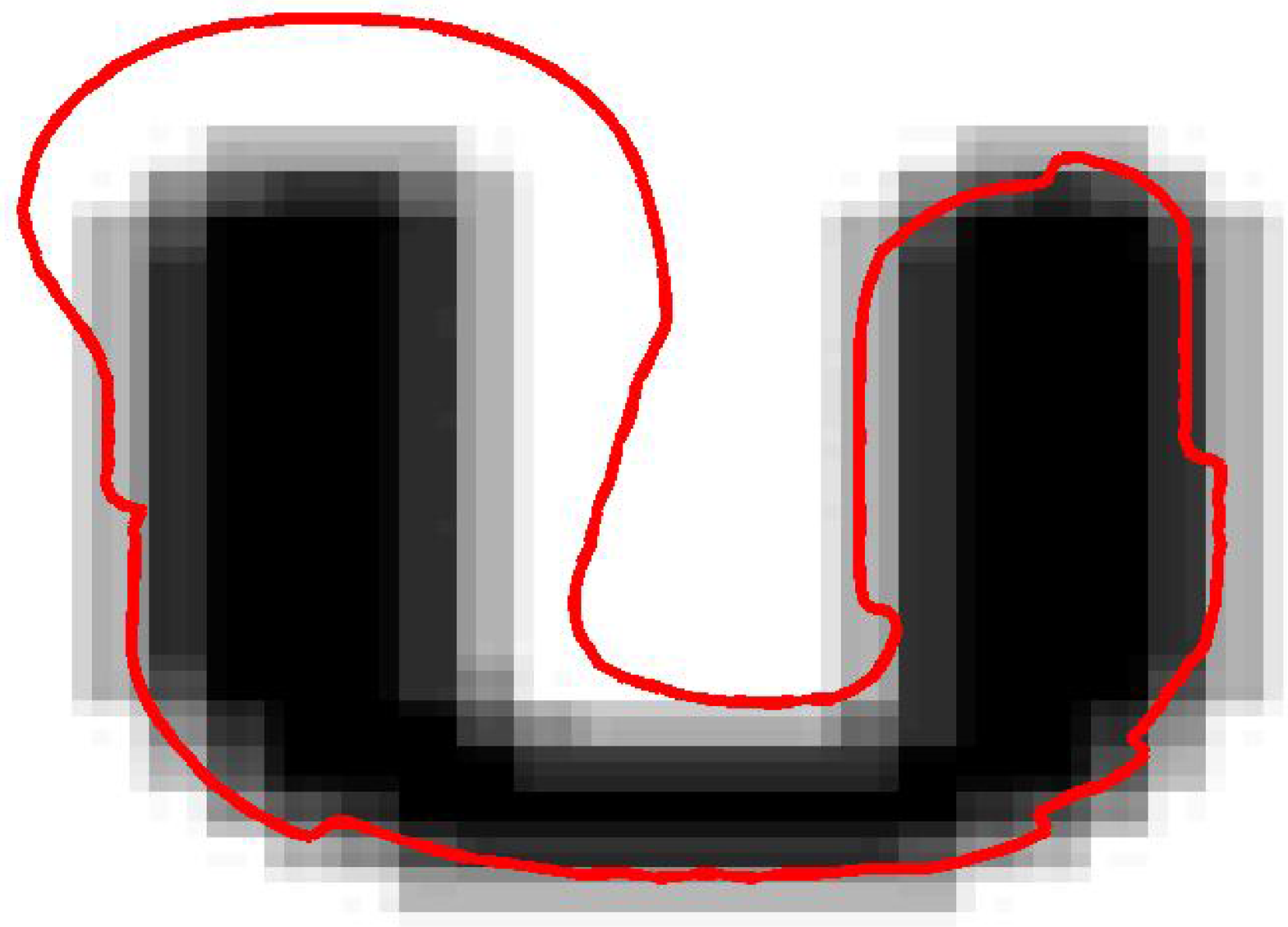}}
  \subfloat[]{\includegraphics[width=0.16\textwidth]{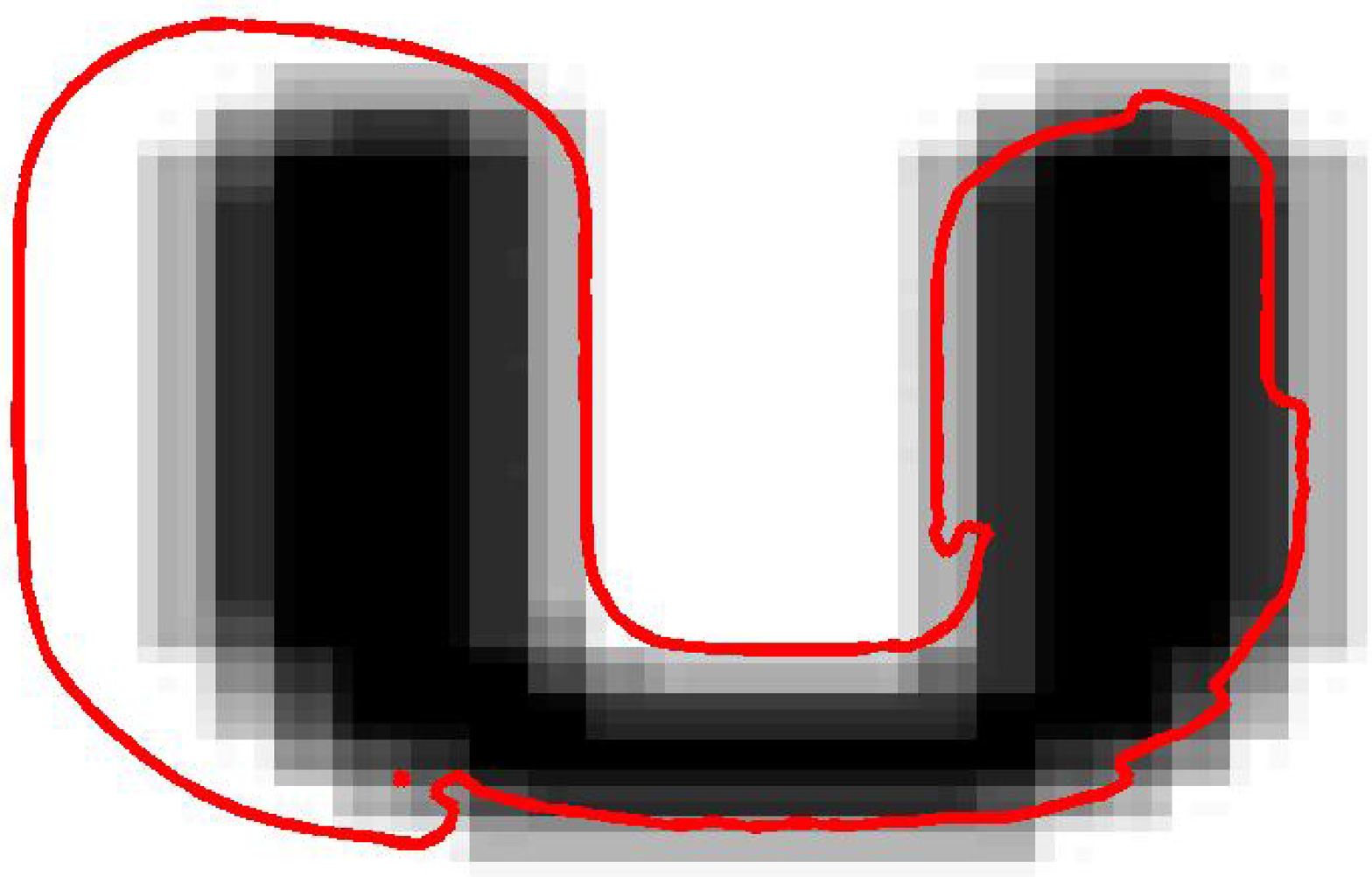}}
  \subfloat[]{\includegraphics[width=0.16\textwidth]{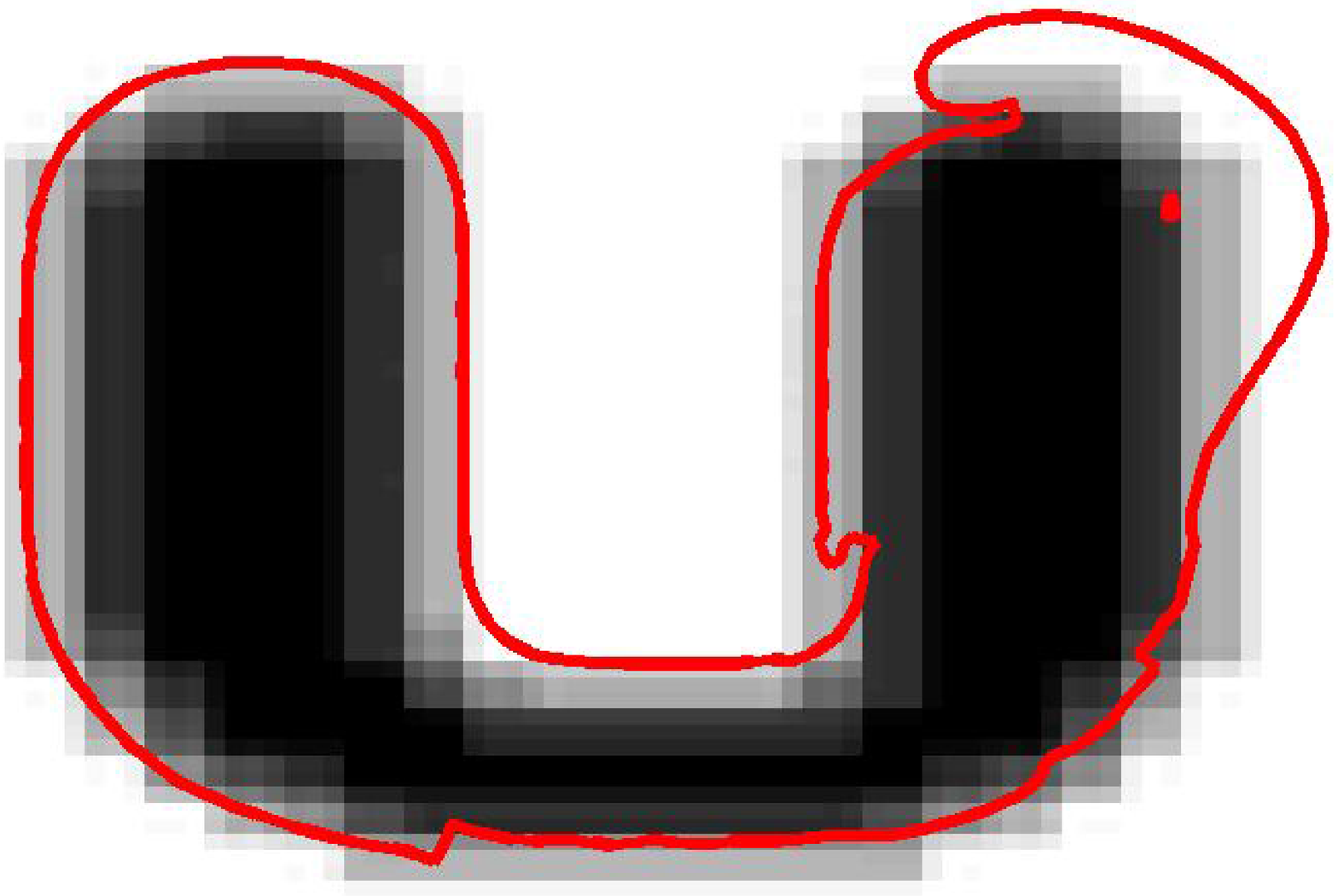}}
  \subfloat[]{\includegraphics[width=0.16\textwidth]{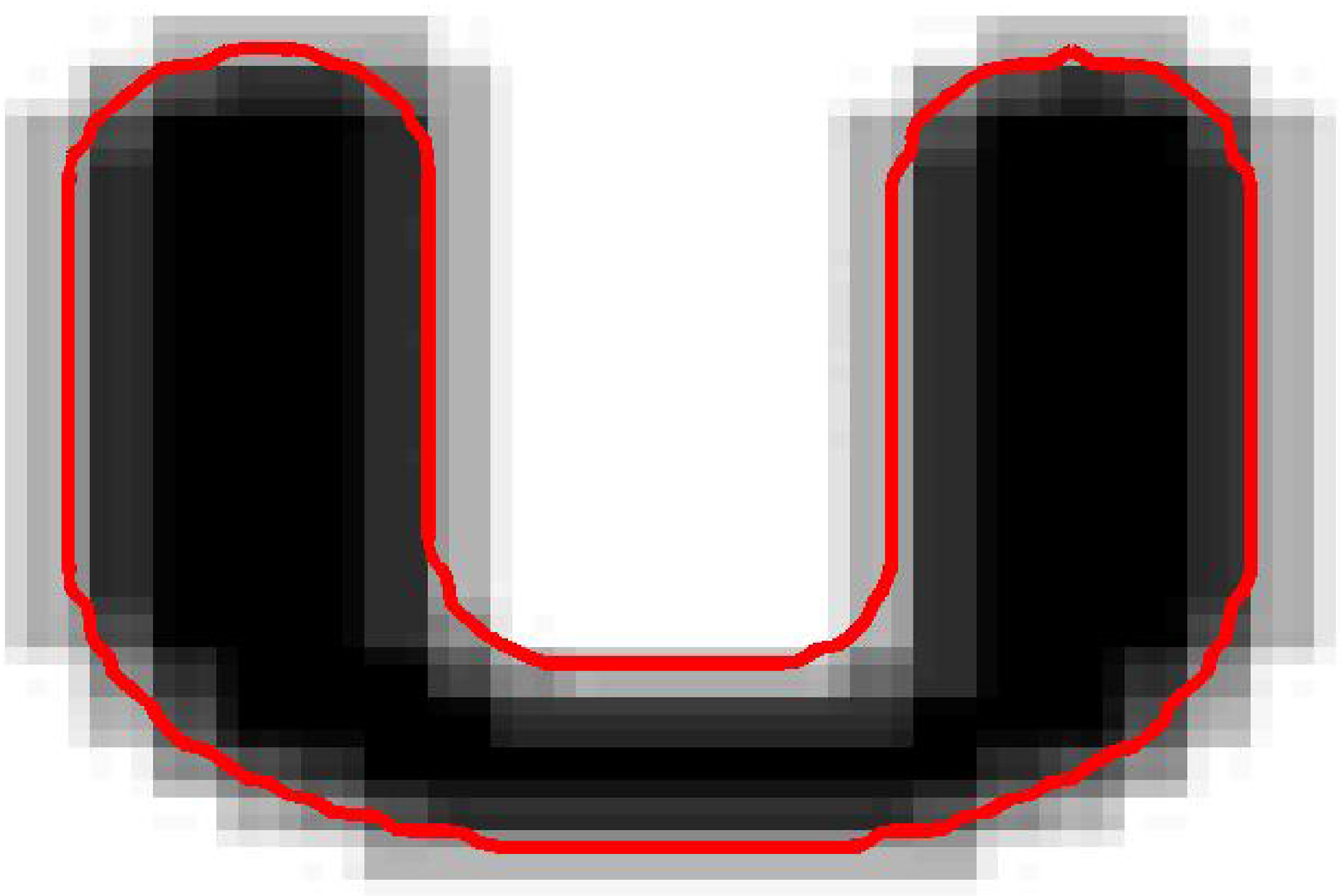}}
  \caption{Segmentation of the U-shape.(a) is the initialization. (b) is the PSP due to the curve evolution by gradient descent. (c)-(f) is the curve evolution by EF. The curve in (f) is also stationary to the curve evolution by gradient descent.}\label{FIG:Seg_U}
\end{figure*}

\setlength{\fboxrule}{0.01in}
\setlength{\fboxsep}{0.01in-\fboxrule}
\definecolor{gray}{rgb}{0.25,0.25,0.25}
Before presenting the main results of the segmentation, the proposed method is demonstrated for segmentation of the U-shape in Figure \ref{Fig:Ushape}. The result is shown in Figure \ref{FIG:Seg_U}.

\begin{figure}[!h]
\centering
\setlength{\tabcolsep}{0pt}
\begin{tabular}{cc}
\subfloat{\fbox{\includegraphics[width=0.235\textwidth]{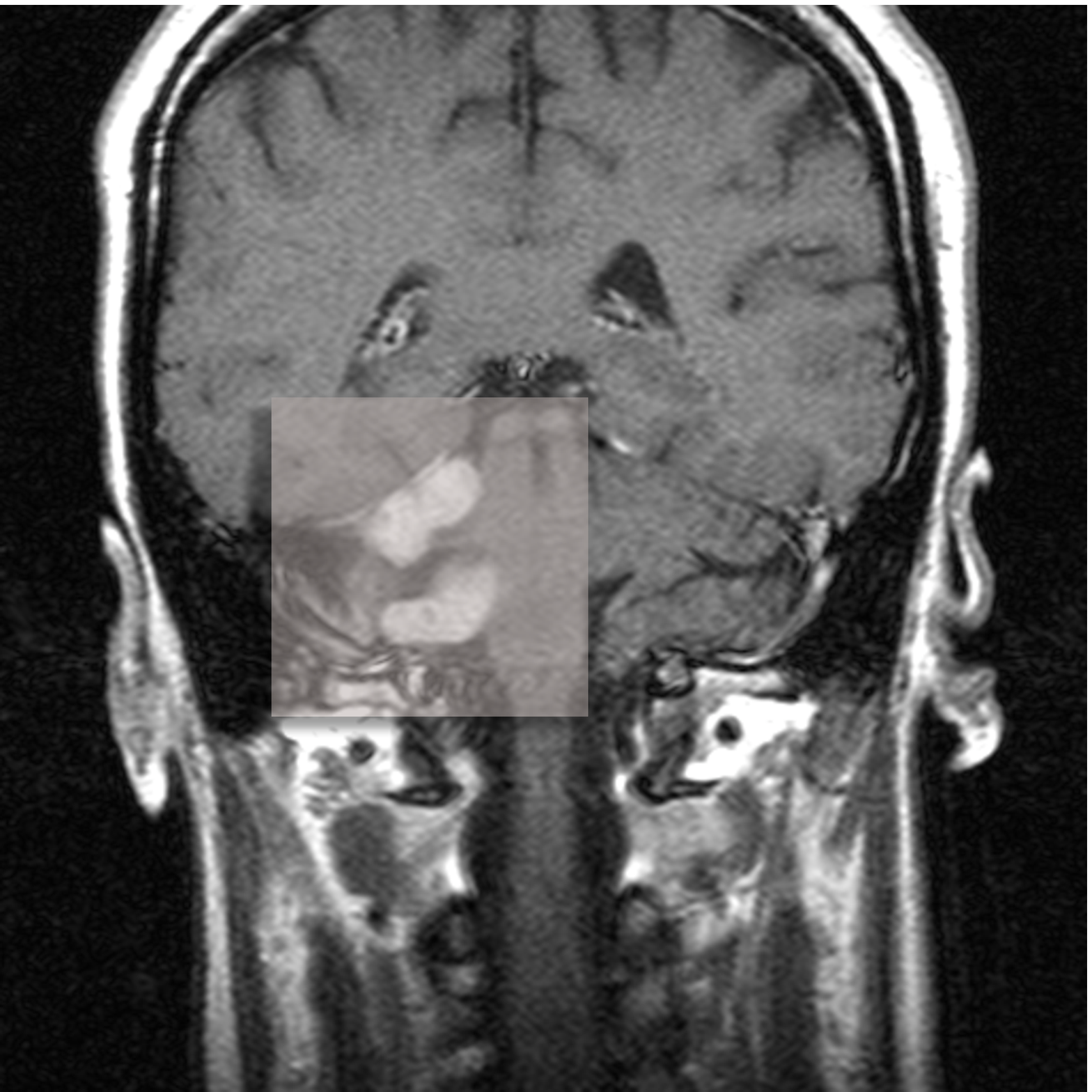}}}&
\subfloat{\fbox{\includegraphics[width=0.235\textwidth]{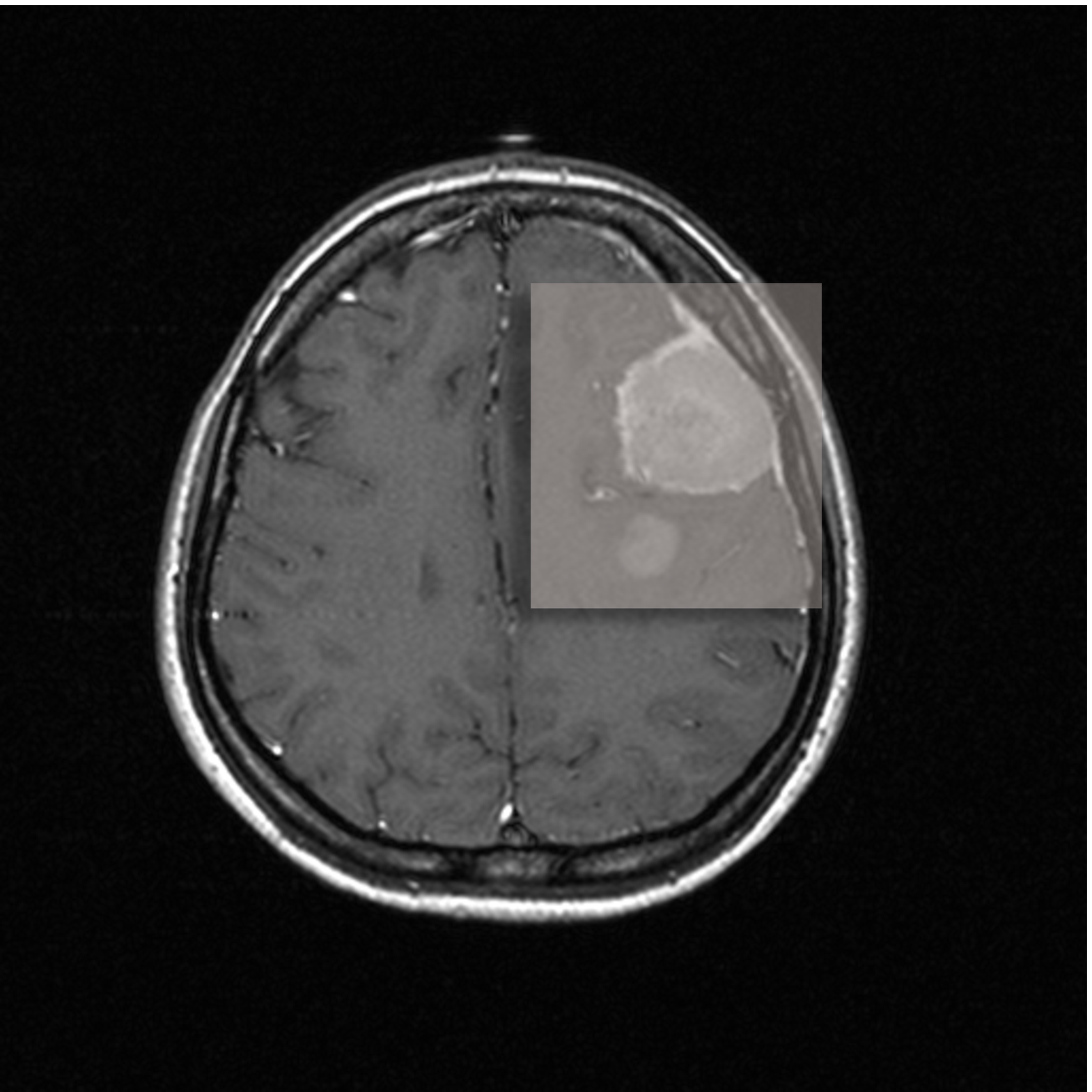}}}
\end{tabular}
\caption{The original real images and the tumor regions (in boxes)}\label{Fig:Realorg}
\end{figure}

\begin{figure*}[thb]
\centering
\setlength{\tabcolsep}{0pt}
\begin{tabular}{ccccc}
\vspace{-10px}
\subfloat[]{{\color{gray}\fbox{\includegraphics[width=0.15\textwidth,height = 1.1in]{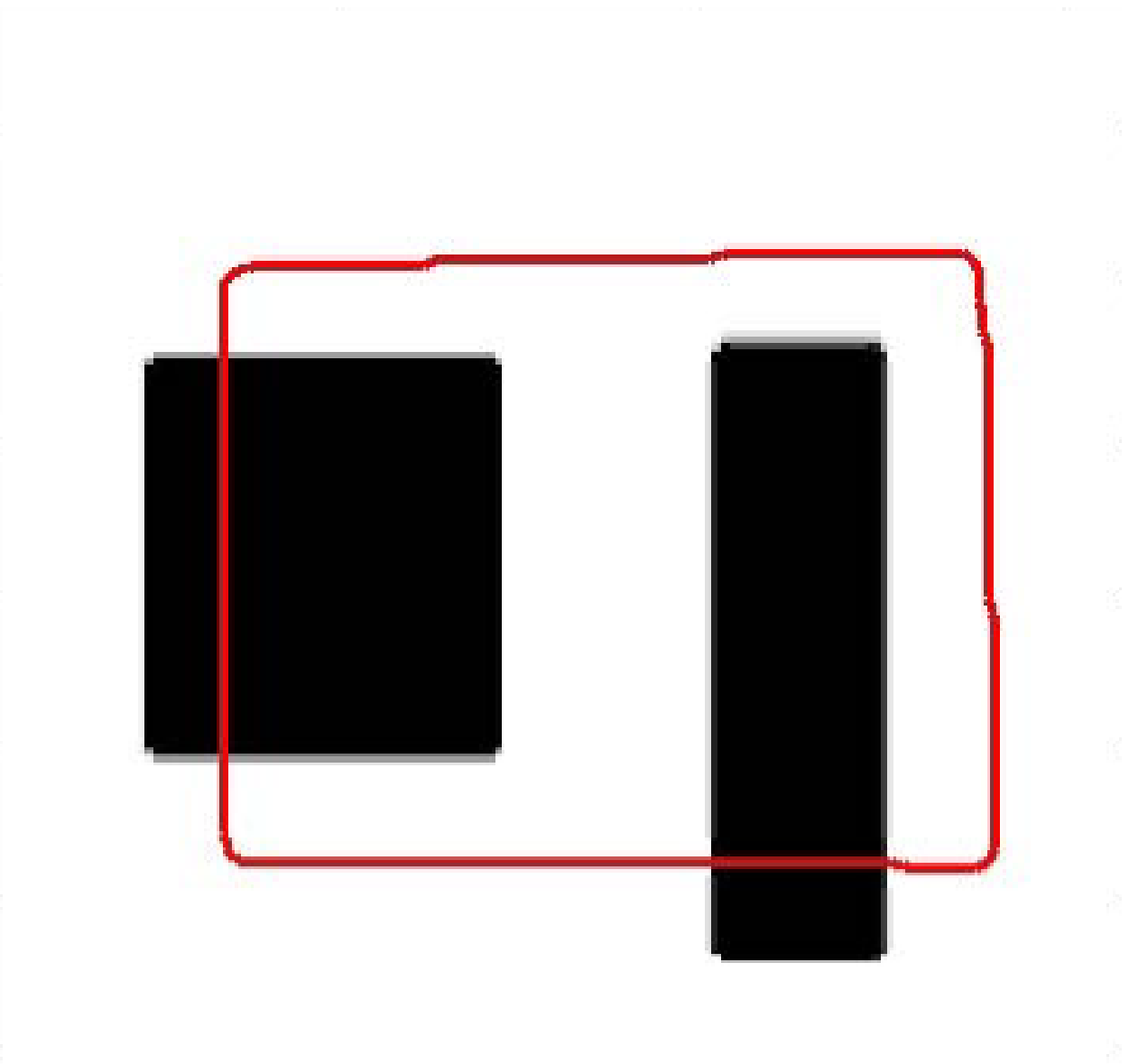}}}}&
\subfloat[]{{\color{gray}\fbox{\includegraphics[width=0.15\textwidth,height = 1.1in]{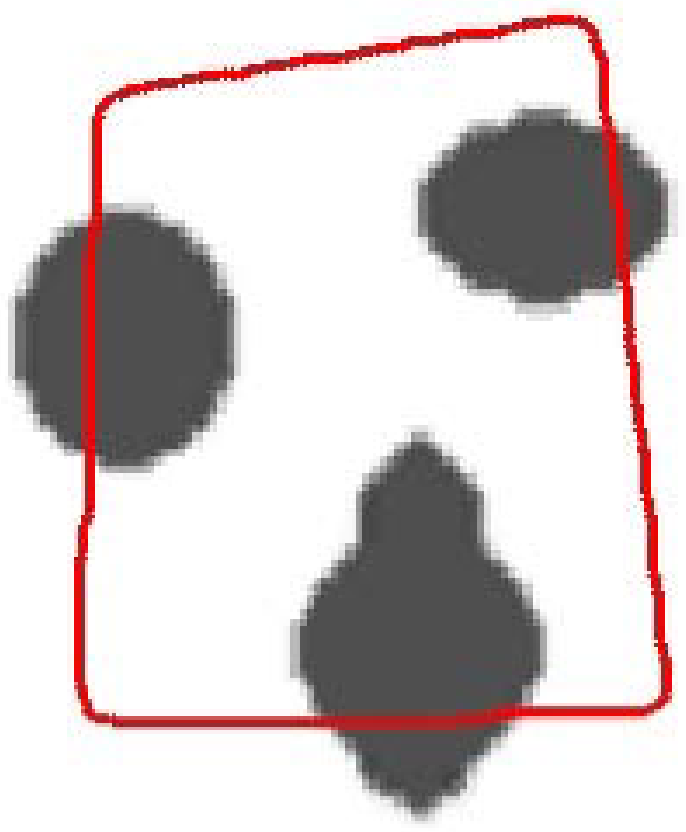}}}}&
\subfloat[]{{\color{gray}\fbox{\includegraphics[width=0.15\textwidth,height = 1.1in]{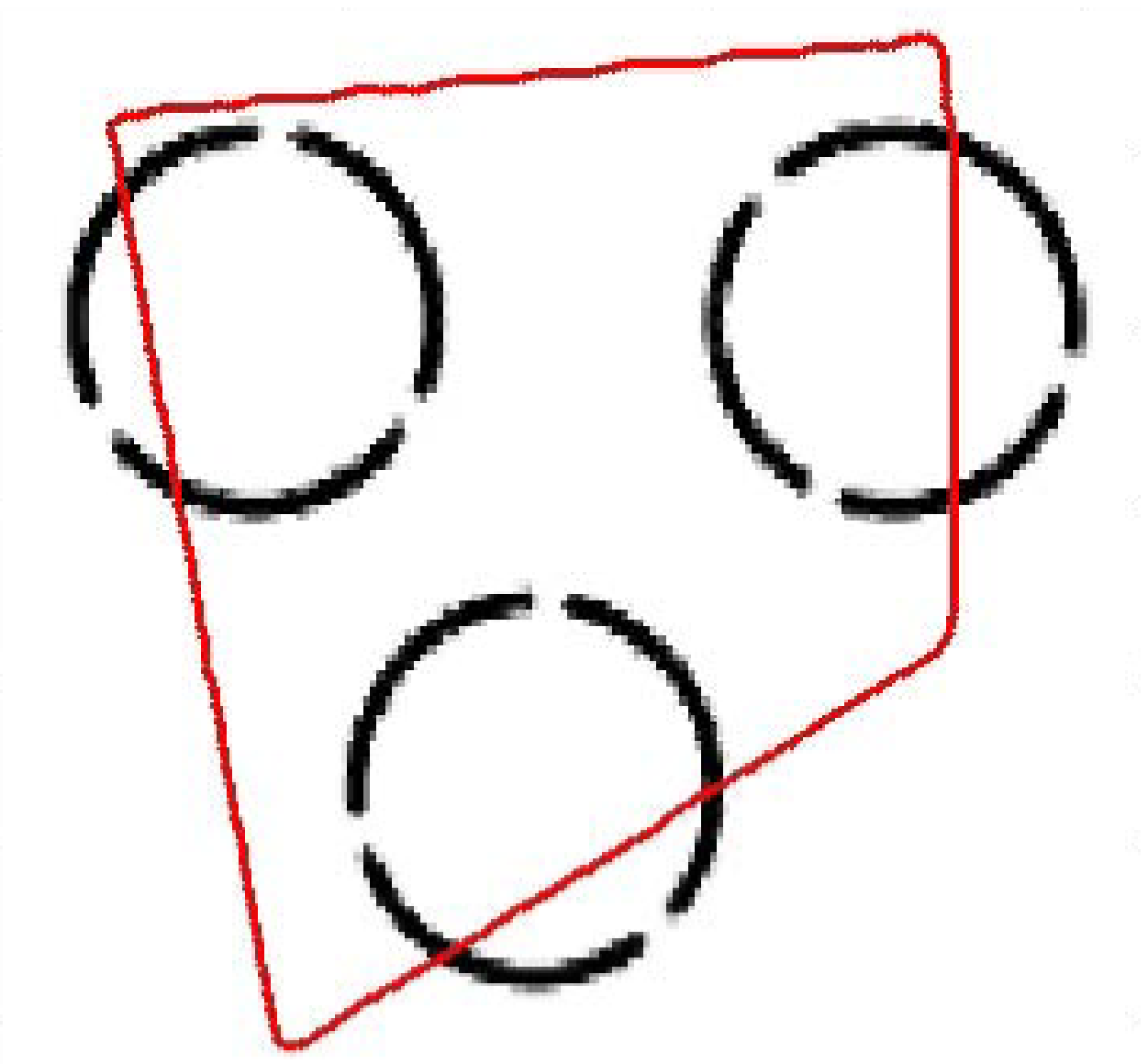}}}}&
\subfloat[]{{\color{gray}\fbox{\includegraphics[width=0.15\textwidth,height = 1.1in]{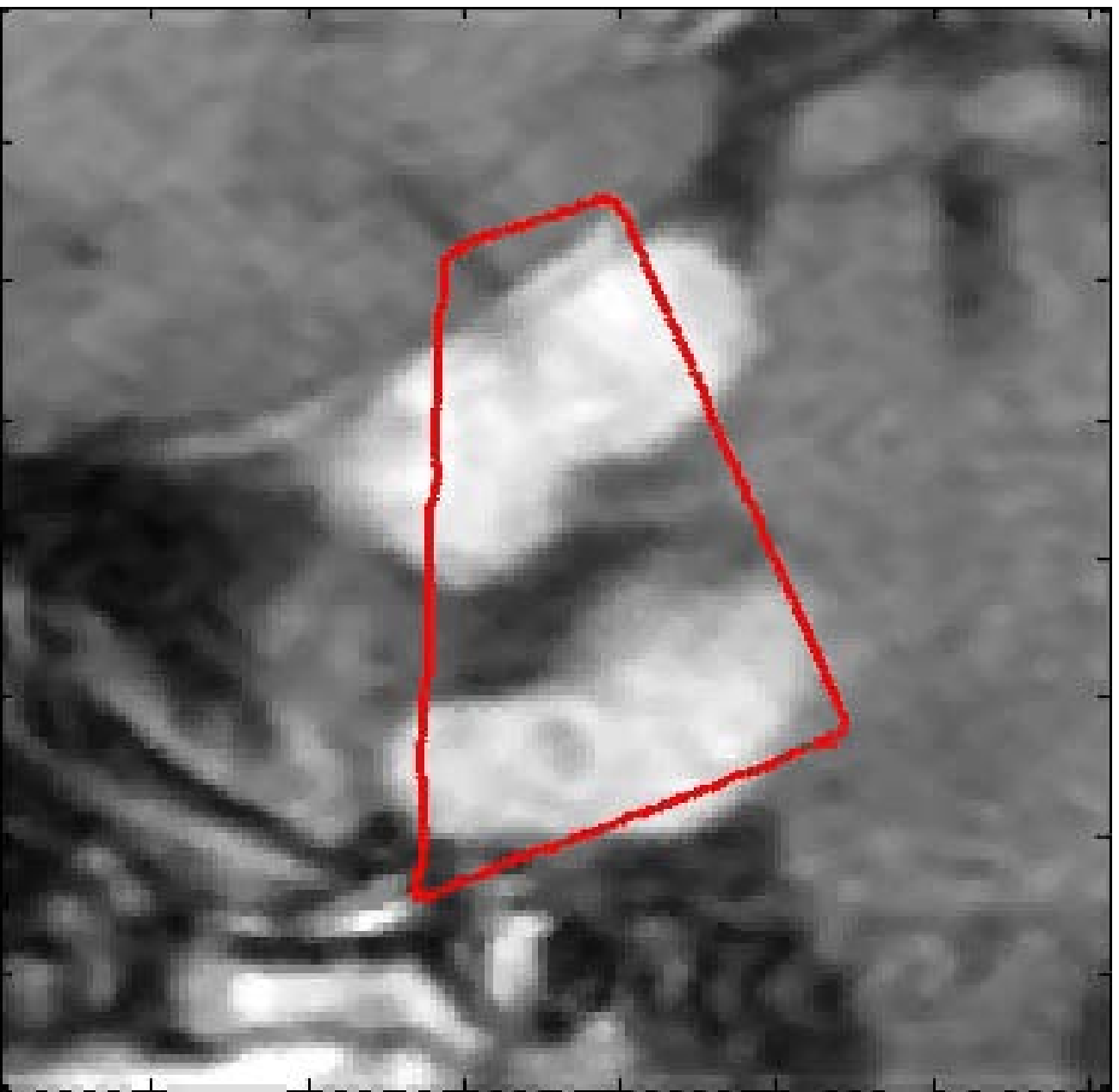}}}}&
\subfloat[]{{\color{gray}\fbox{\includegraphics[width=0.15\textwidth,height = 1.1in]{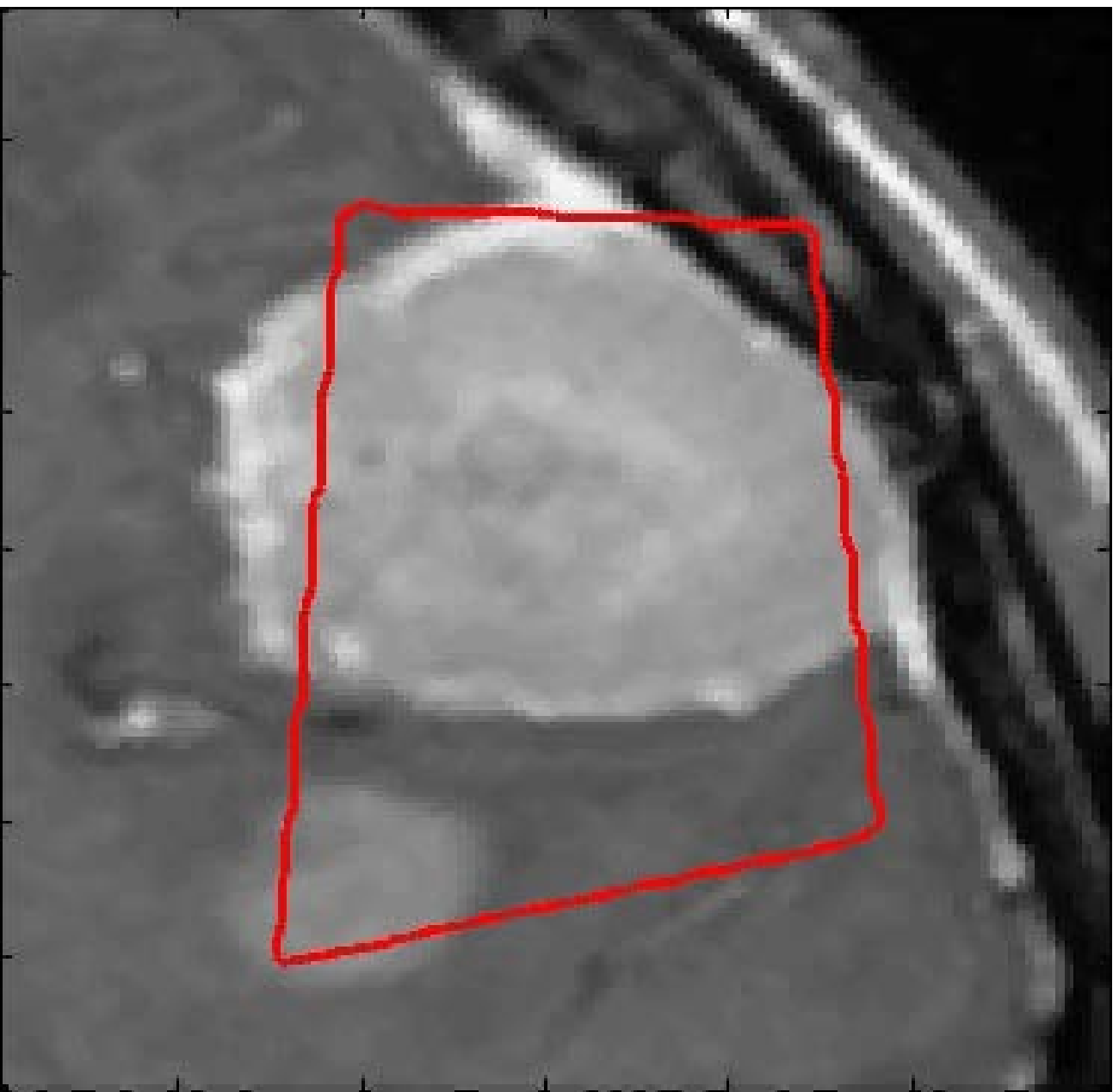}}}}
\end{tabular}
\caption{Inputs and initializations}\label{Fig:Init_inputs}
\end{figure*}

The other images used for evaluating the proposed method are shown in Figure \ref{Fig:Init_inputs} overlaid with the red initial curves. The same initializations are used in all the comparisons. The initial curves are partially inside the objects and partially outside, which means that solely shrinking or solely expanding the initial curves cannot extract the objects. The image sizes can be found in Table \ref{TB:KEYIter}. The chosen images are of different characteristics. The image of two rectangles in Figure \ref{Fig:Init_inputs}(a) is relatively simple for segmentation since the gray level of both foreground and background are homogeneous and distinct, the edges are strong too. Besides, there are only two objects, hence the topology is simple. In Figure \ref{Fig:Init_inputs}(b), the image is also simple except that there are three objects, which is a little more difficult than the former one. In Figure \ref{Fig:Init_inputs}(c), there are nine curve segments forming three circles. The edge-based active contours may see three circles of disconnected boundaries in the image, and the region-based method may see the objects in the image as composed of nine narrow and curved regions. Figures \ref{Fig:Init_inputs}(d) and \ref{Fig:Init_inputs}(e) are MRI medical images of human brain with multiple tumor regions. The original brain scans are in Figure \ref{Fig:Realorg}. From Figures \ref{Fig:Init_inputs}(d) and (e), we see that there are many spurious curved-edges in the images.  Besides, there are some regions with intensities similar to the targeted tumor regions. These make the region-based active contour to be unsuitable for such images. For edge-based active contours, there also exists the Pseudo Stationary Phenomenon.

\begin{figure*}[htb]
\centering
\setlength{\tabcolsep}{0pt}
\begin{tabular}{ccc}
\subfloat[]{\fbox{\includegraphics[width = 0.252\textwidth,height=1.3in]{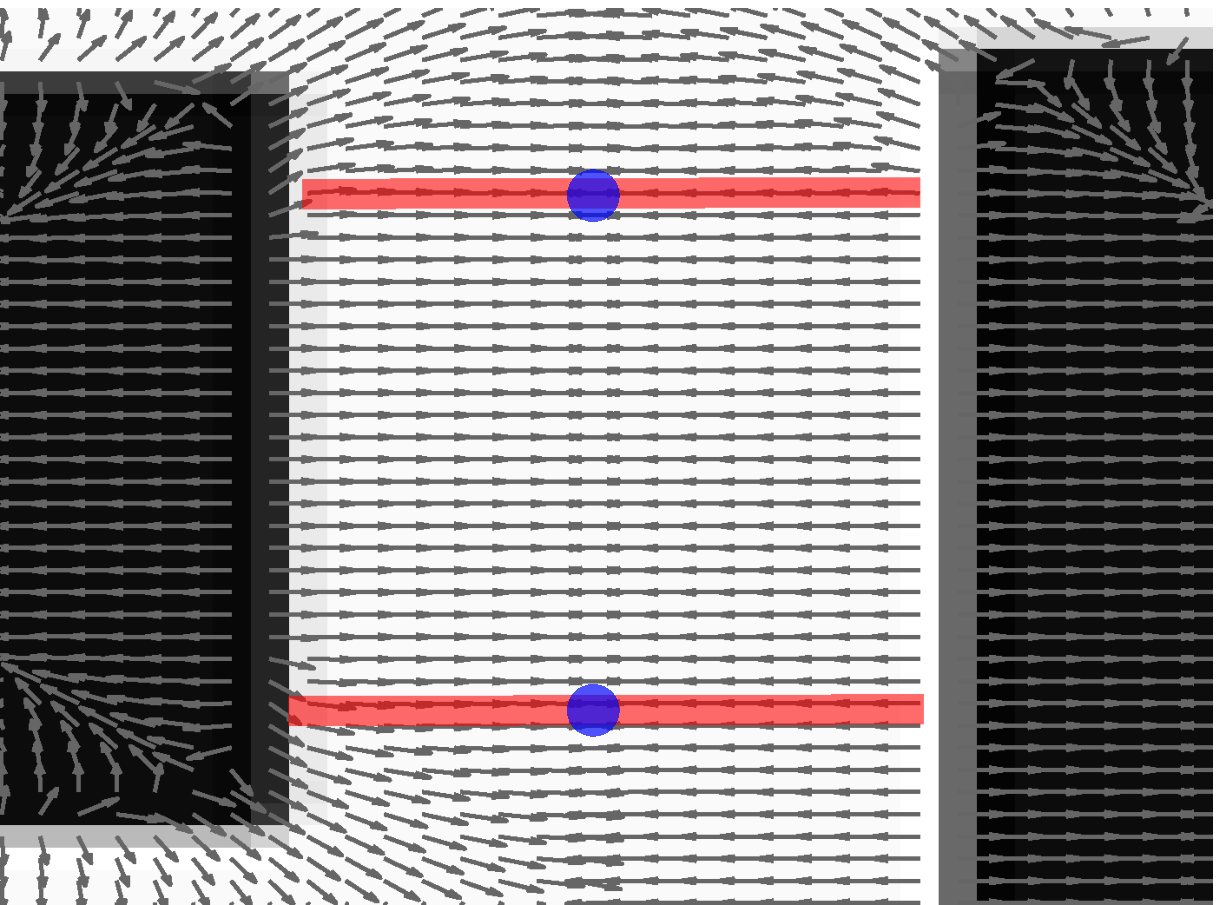}}}&
\subfloat[]{\fbox{\includegraphics[width = 0.252\textwidth,height=1.3in]{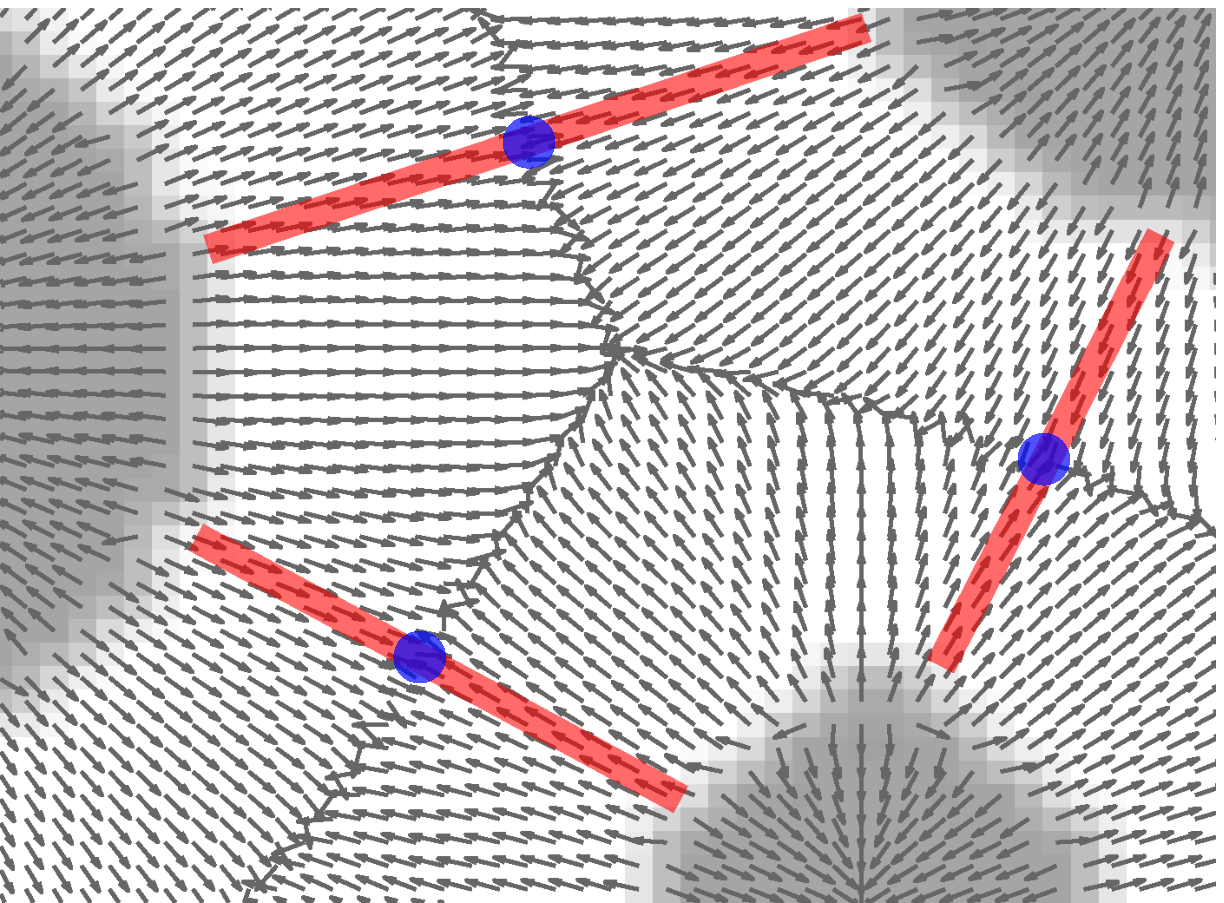}}}&
\subfloat[]{\fbox{\includegraphics[width = 0.252\textwidth,height=1.3in]{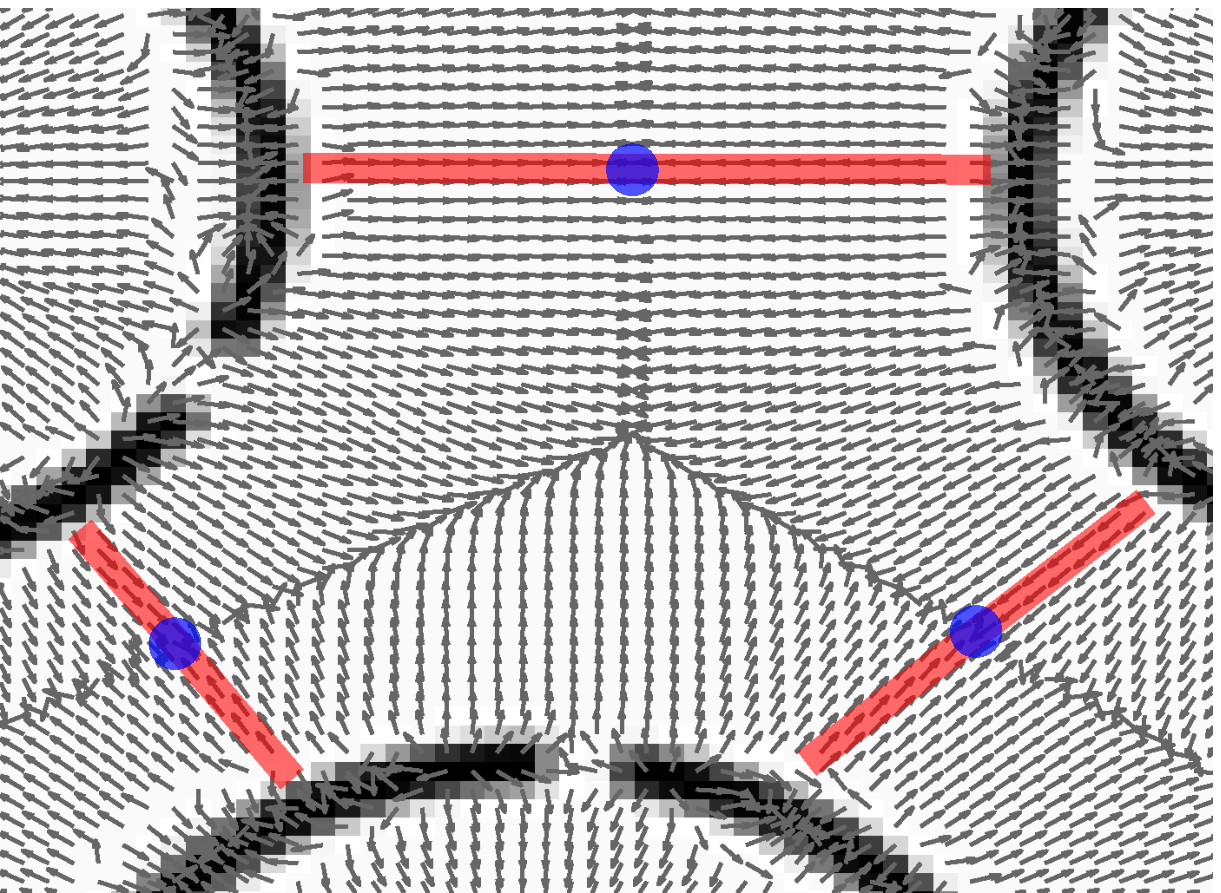}}}
\end{tabular}
\caption{Gradient fields, PSP and Saddle points}\label{Fig:saddles}
\end{figure*}

In Figure \ref{Fig:saddles}, we visualize the {\emph{normalized}} gradient field of the edge indication functions of the images in Figures \ref{Fig:Init_inputs} (a), (b) and (c) respectively to demonstrate the PSP. We also visualize the saddle points by the blue dots on the red lines of which the gradients are in the tangential direction. Later we shall see the early termination problem of GAC, i.e. the PSP problem of the GeoSnakes, near those lines.

\begin{figure*}[!h]
\centering
\setlength{\tabcolsep}{0pt}
\begin{tabular}{ccccc}
{{\color{gray}\fbox{\includegraphics[width=0.15\textwidth,height = 1.1in]{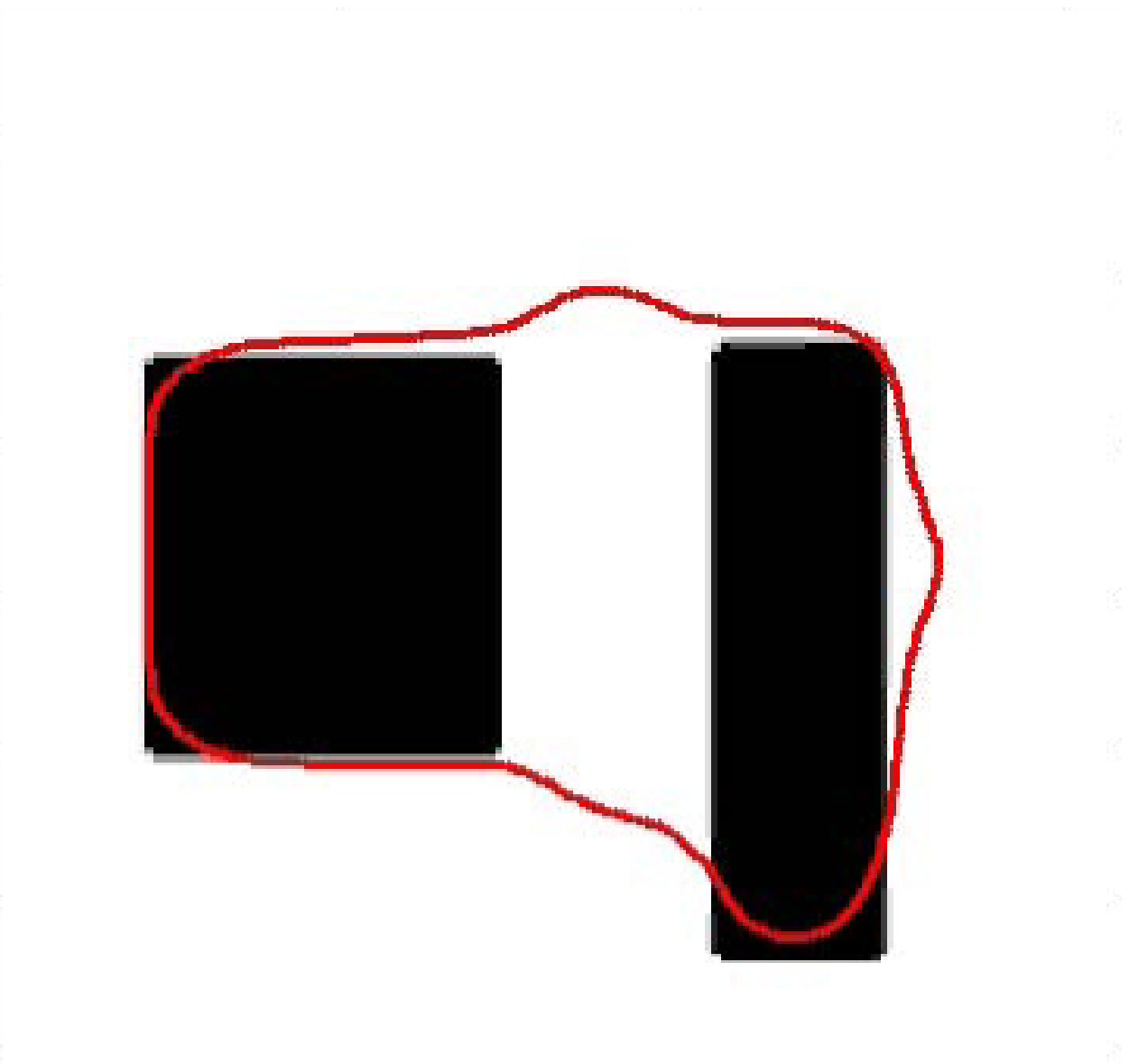}}}}
&{{\color{gray}\fbox{\includegraphics[width=0.15\textwidth,height = 1.1in]{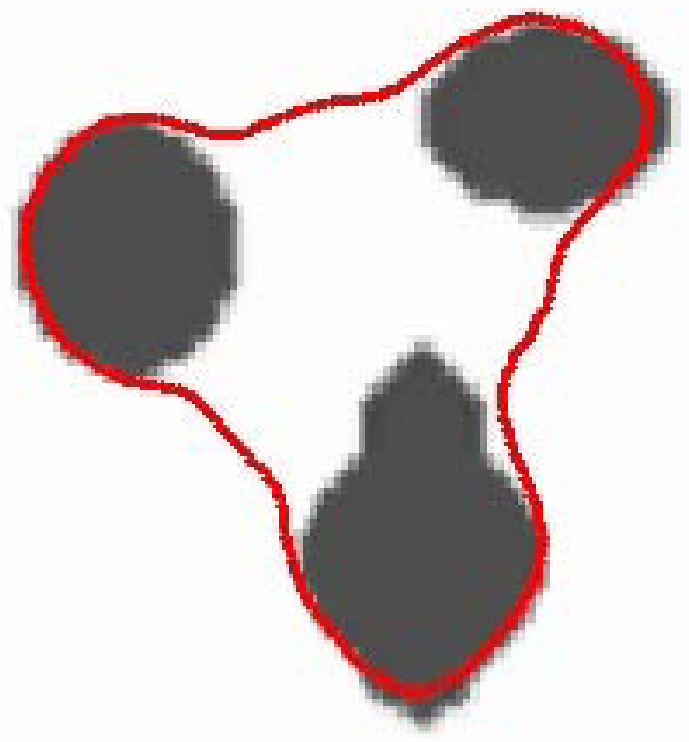}}}}
&{{\color{gray}\fbox{\includegraphics[width=0.15\textwidth,height = 1.1in]{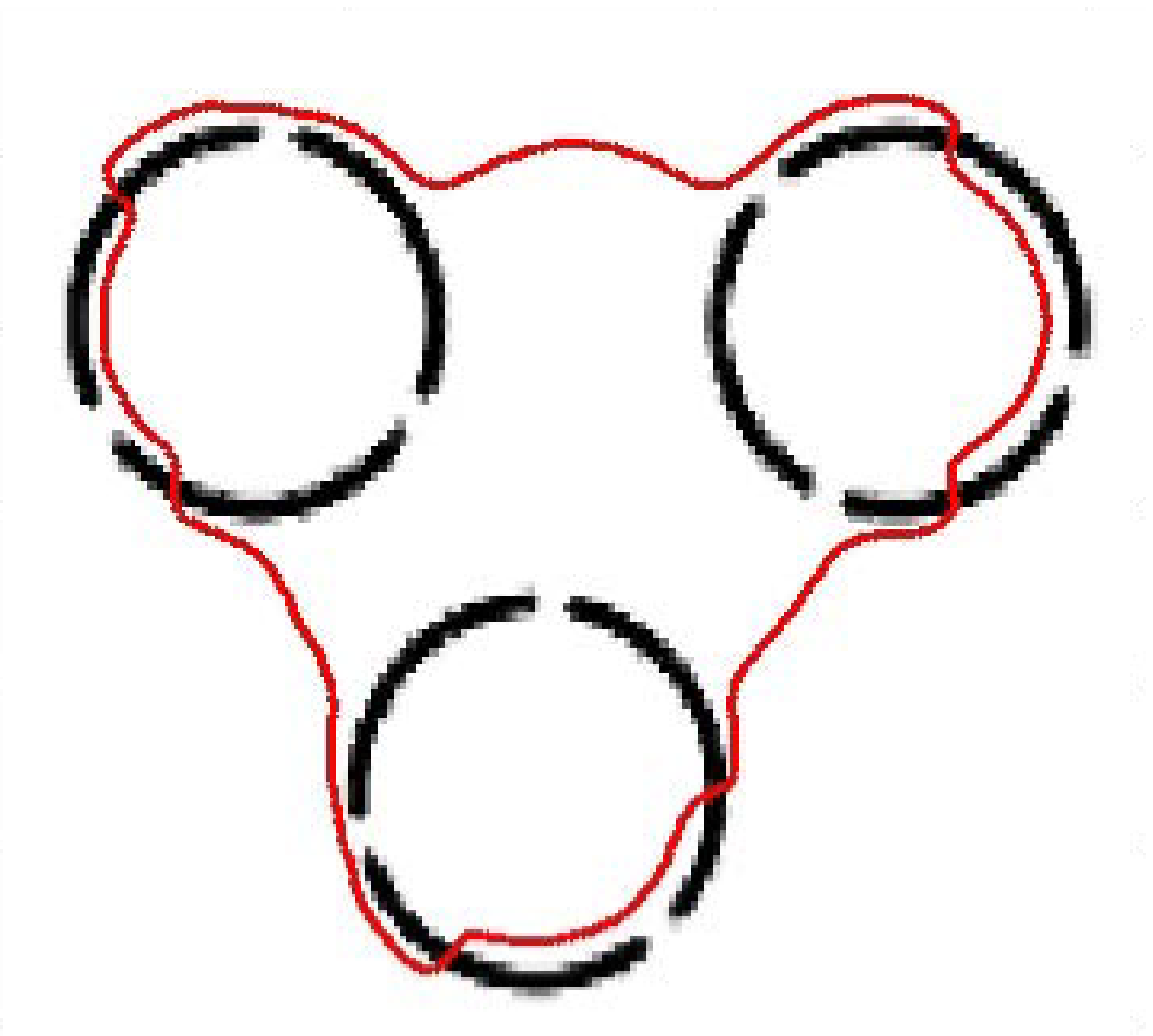}}}}
&{{\color{gray}\fbox{\includegraphics[width=0.15\textwidth,height = 1.1in]{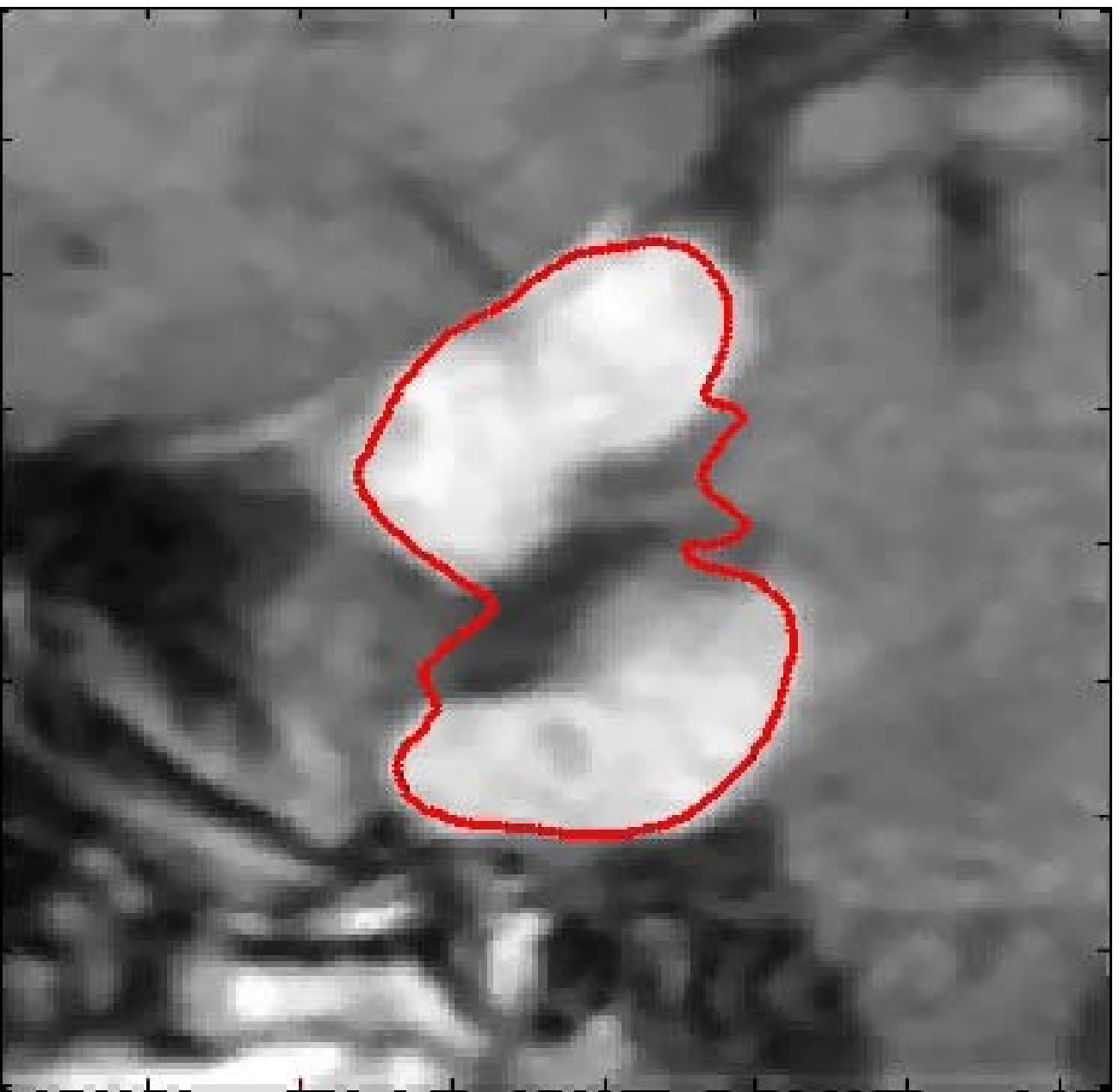}}}}
&{{\color{gray}\fbox{\includegraphics[width=0.15\textwidth,height = 1.1in]{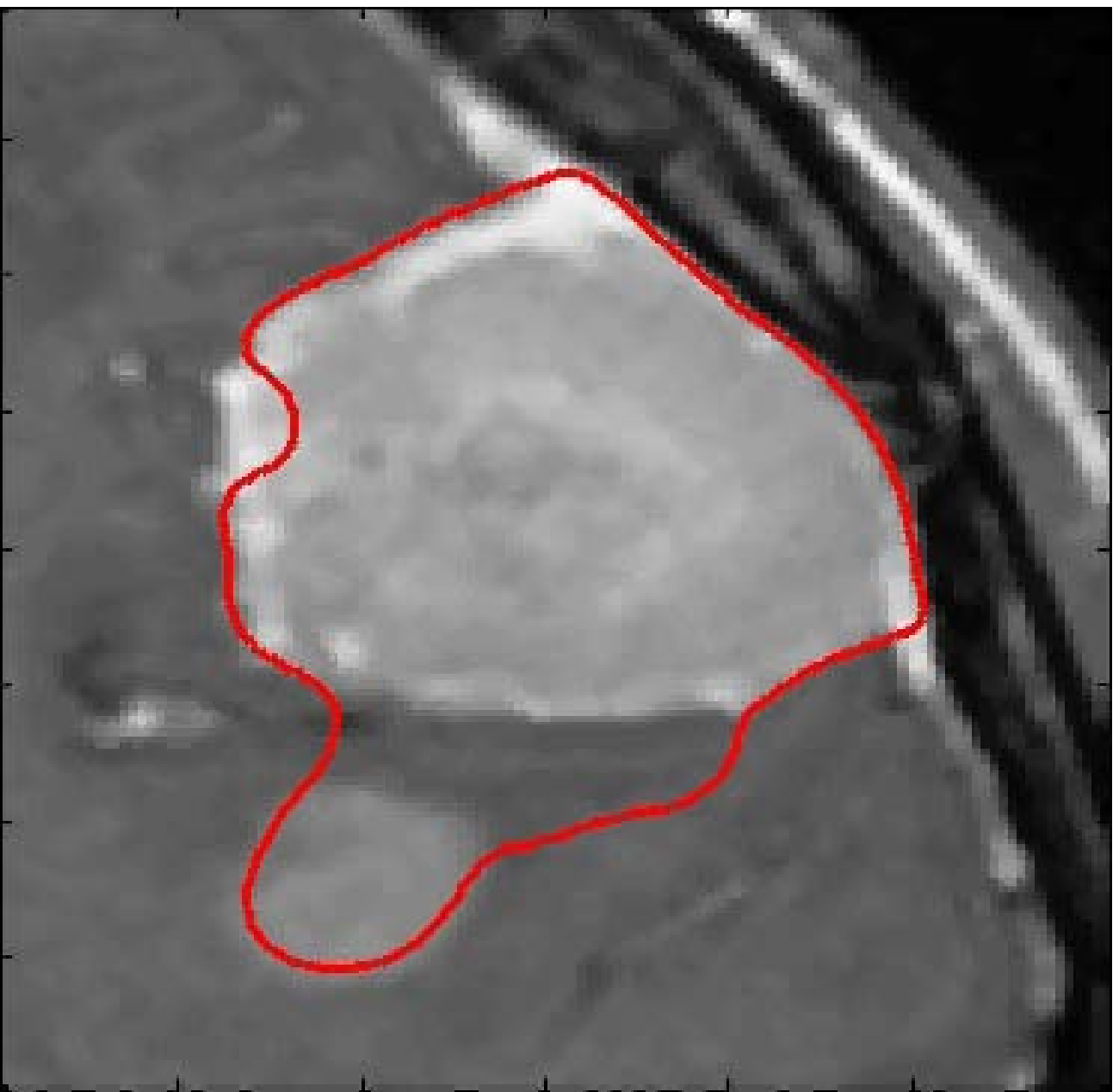}}}}\\
{{\color{gray}\fbox{\includegraphics[width=0.15\textwidth,height = 1.1in]{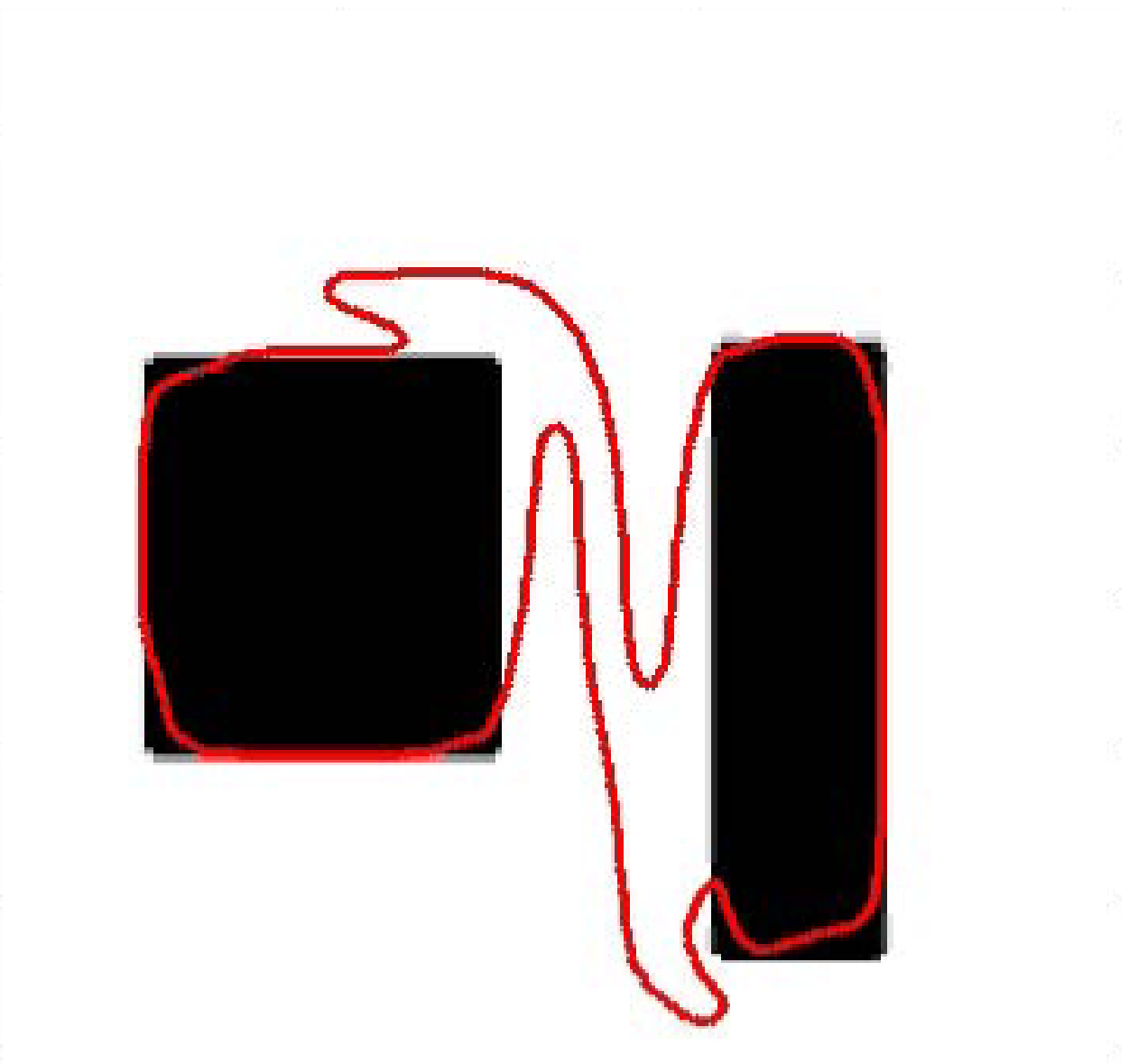}}}}
&{{\color{gray}\fbox{\includegraphics[width=0.15\textwidth,height = 1.1in]{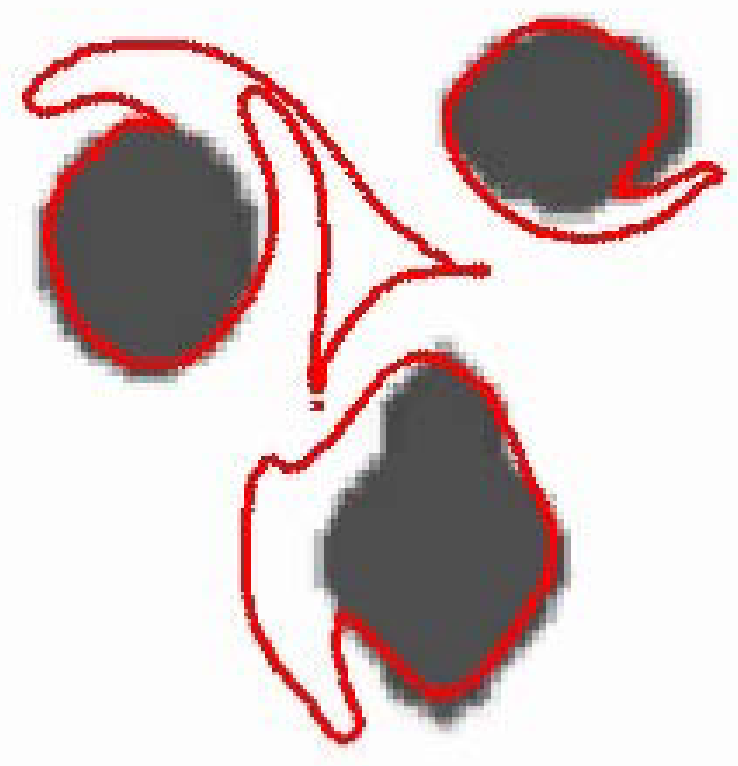}}}}
&{{\color{gray}\fbox{\includegraphics[width=0.15\textwidth,height = 1.1in]{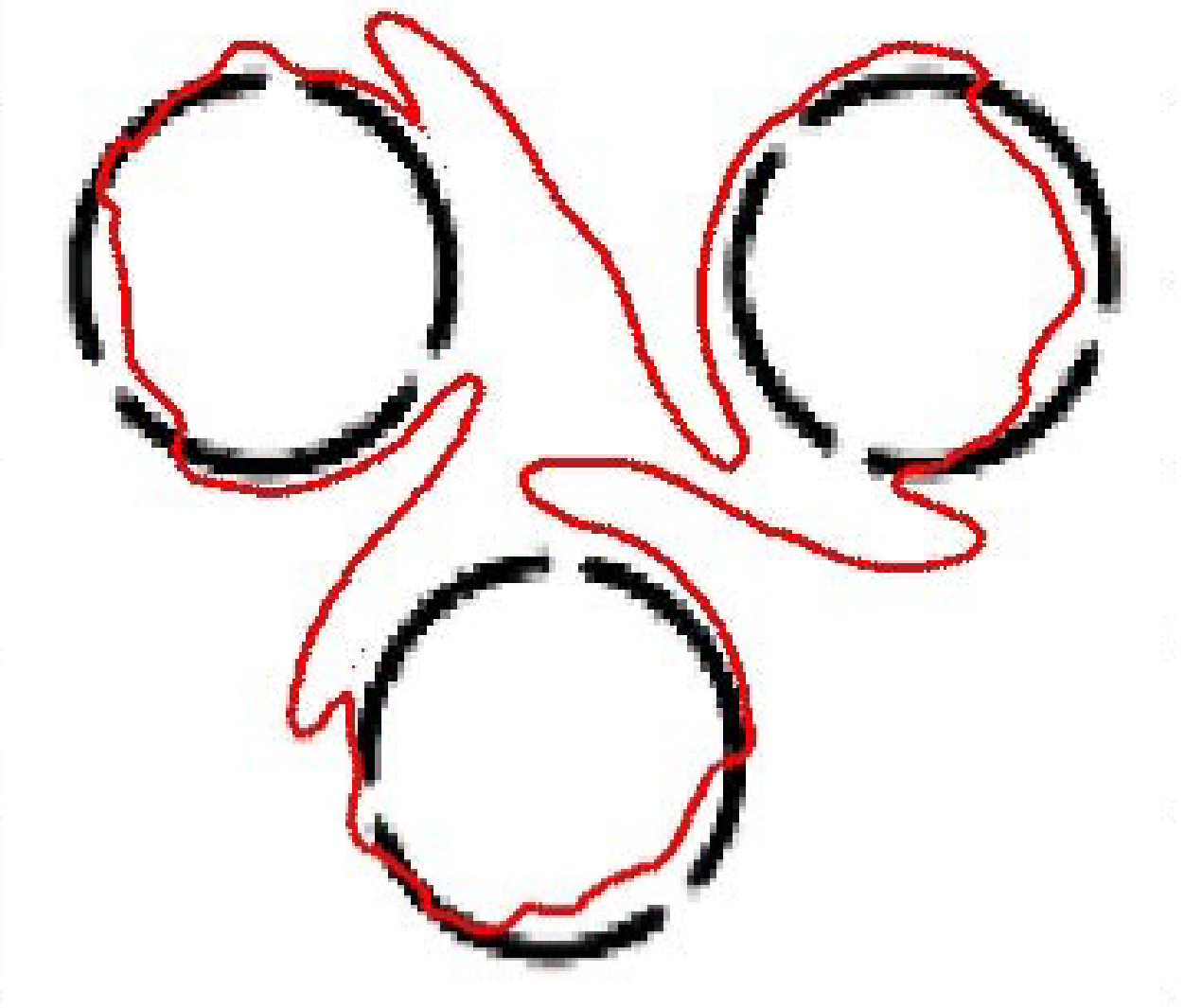}}}}
&{{\color{gray}\fbox{\includegraphics[width=0.15\textwidth,height = 1.1in]{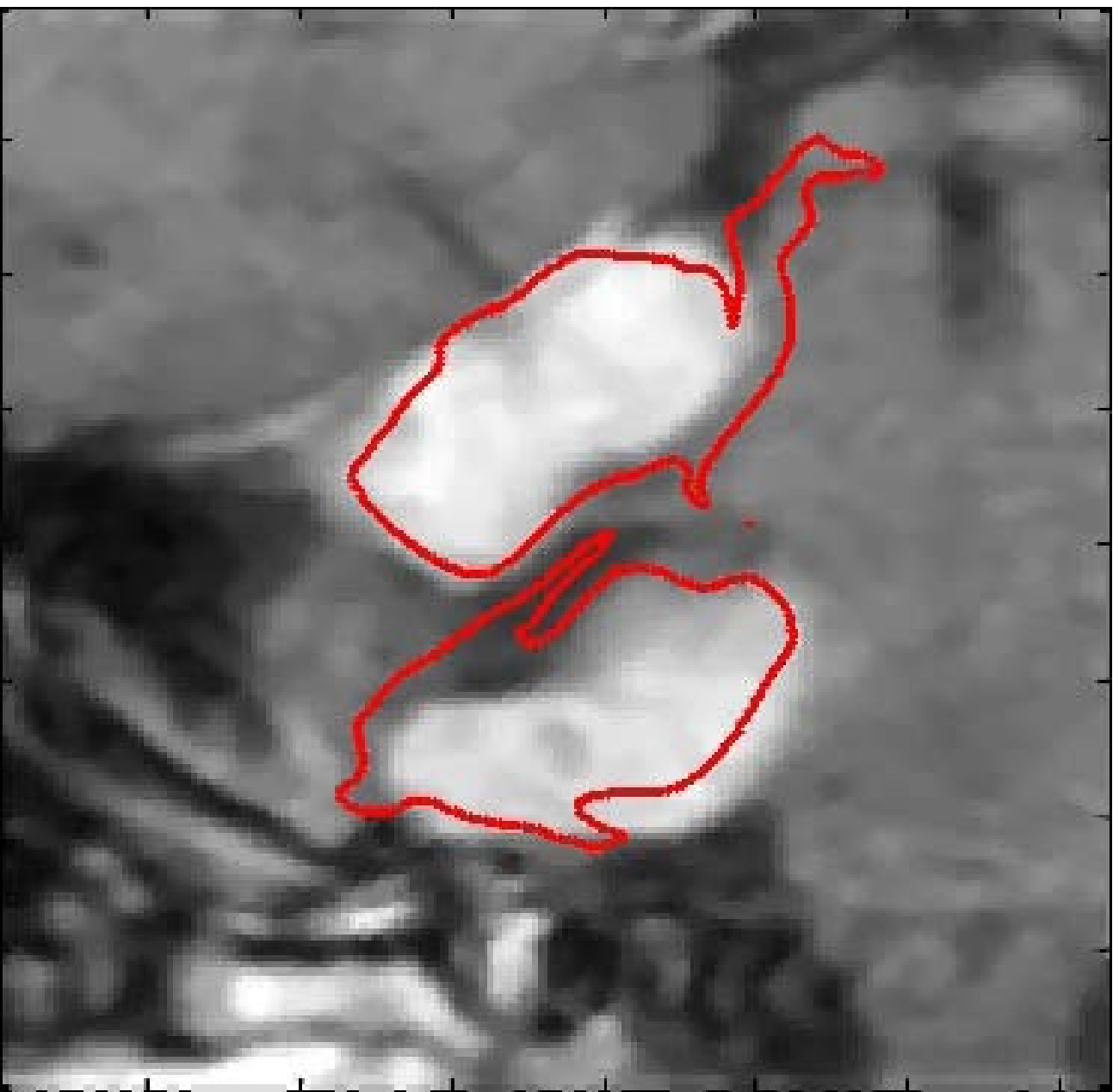}}}}
&{{\color{gray}\fbox{\includegraphics[width=0.15\textwidth,height = 1.1in]{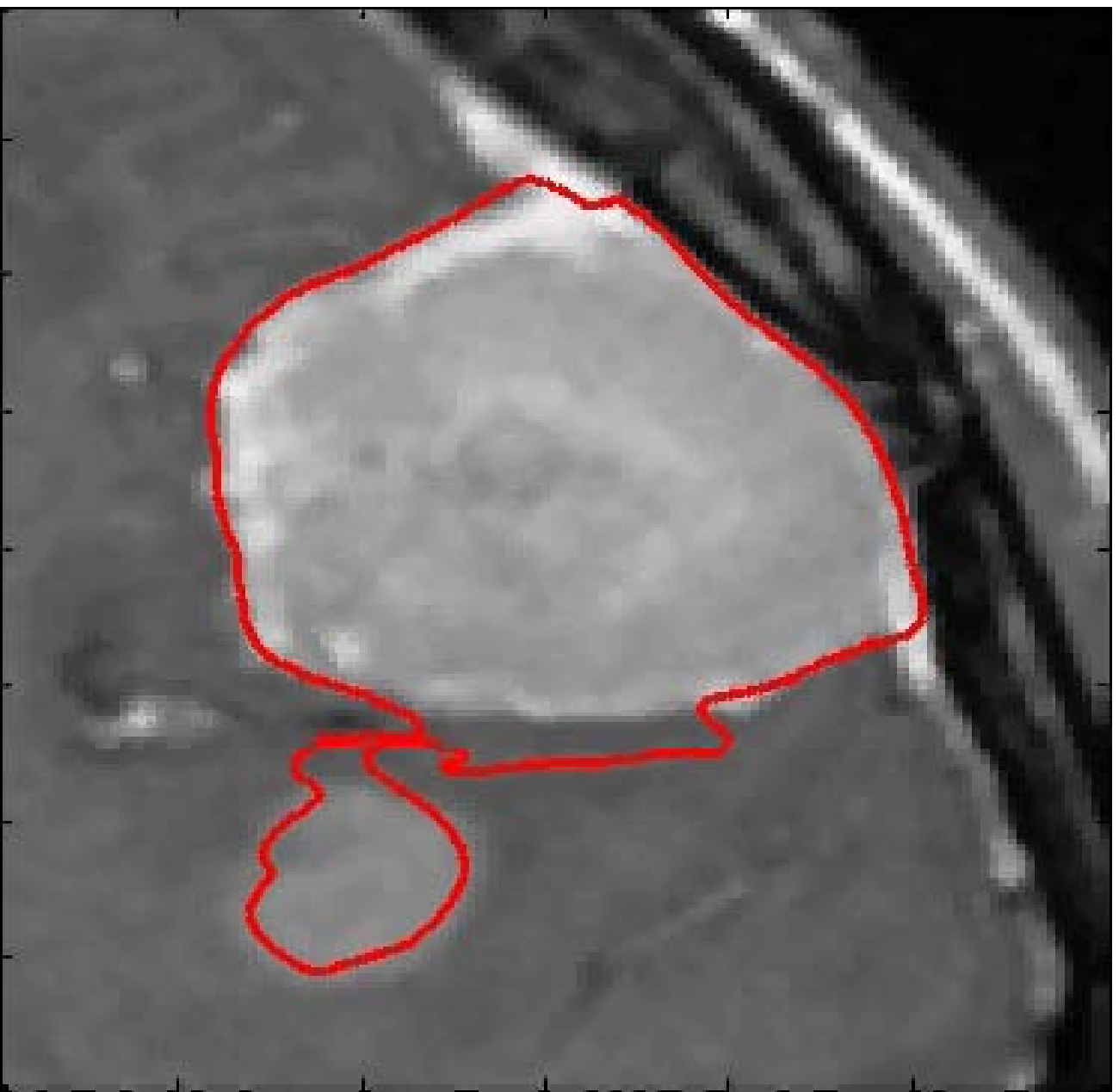}}}}\\
{{\color{gray}\fbox{\includegraphics[width=0.15\textwidth,height = 1.1in]{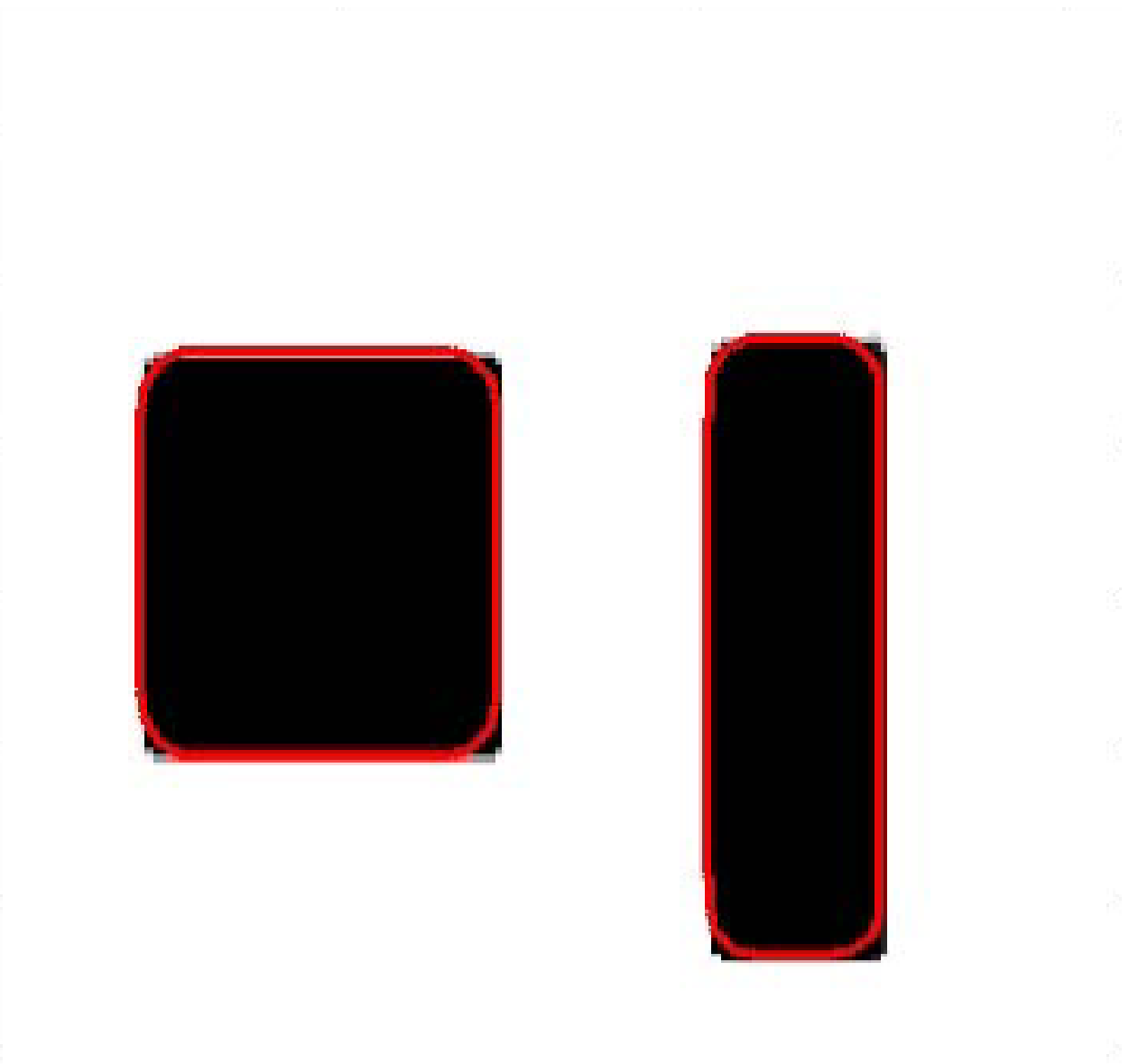}}}}
&{{\color{gray}\fbox{\includegraphics[width=0.15\textwidth,height = 1.1in]{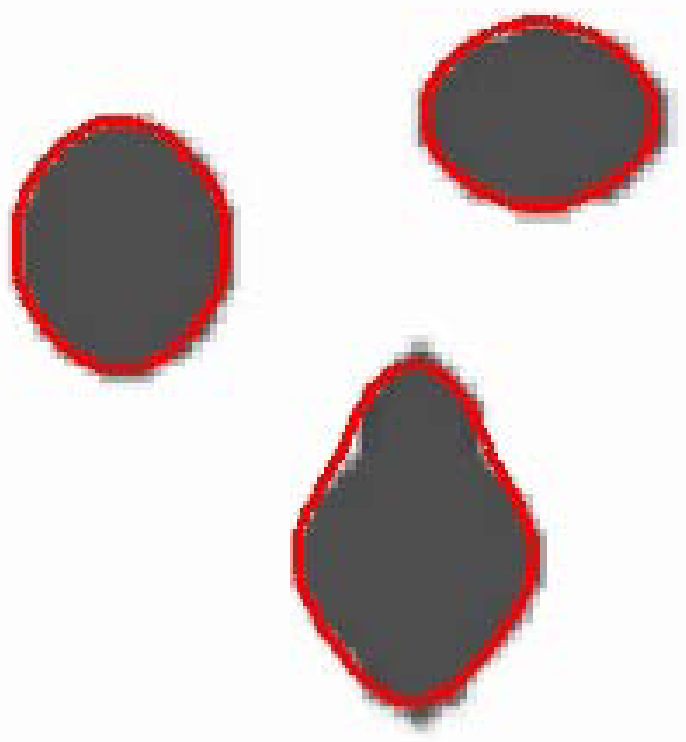}}}}
&{{\color{gray}\fbox{\includegraphics[width=0.15\textwidth,height = 1.1in]{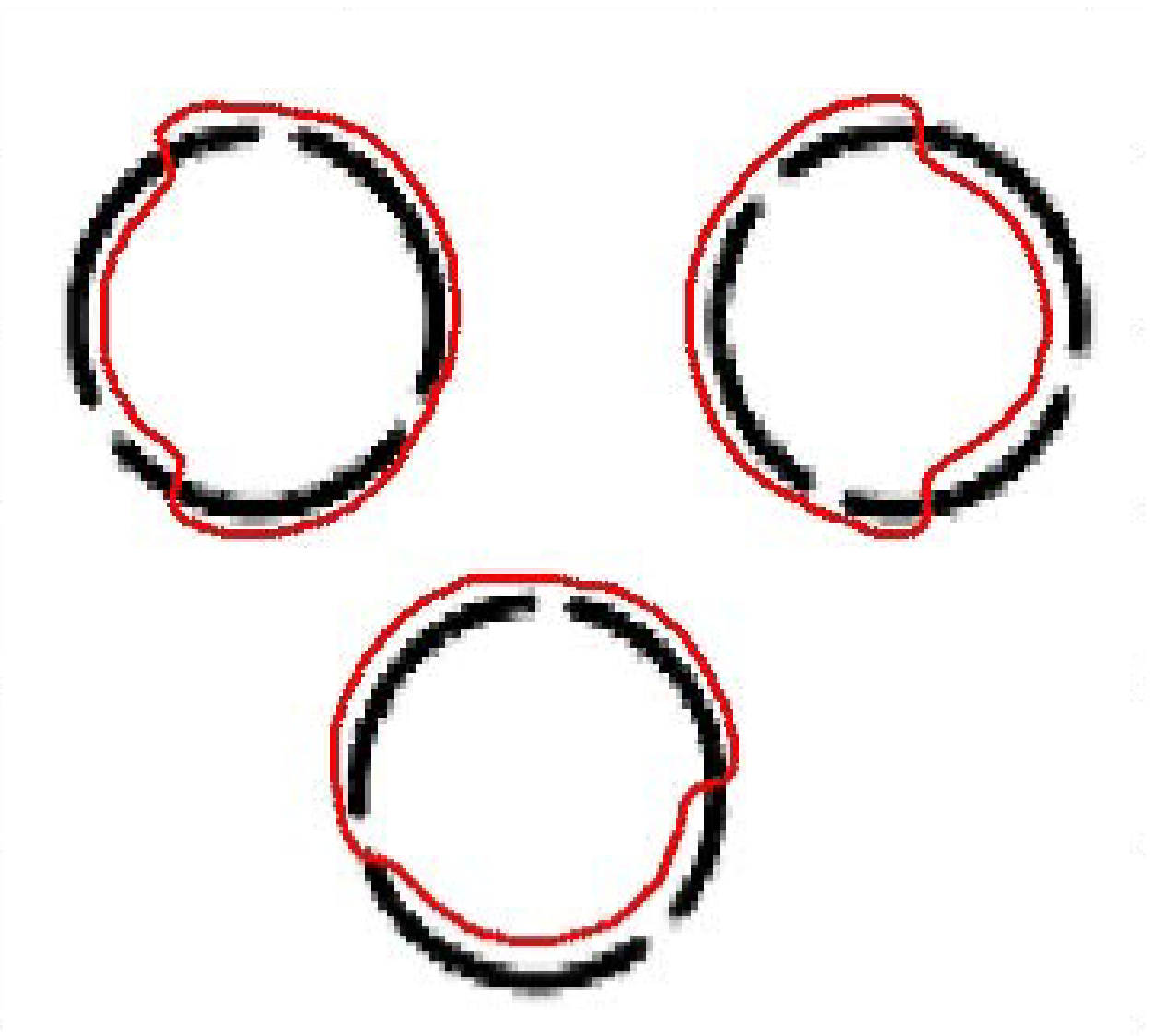}}}}
&{{\color{gray}\fbox{\includegraphics[width=0.15\textwidth,height = 1.1in]{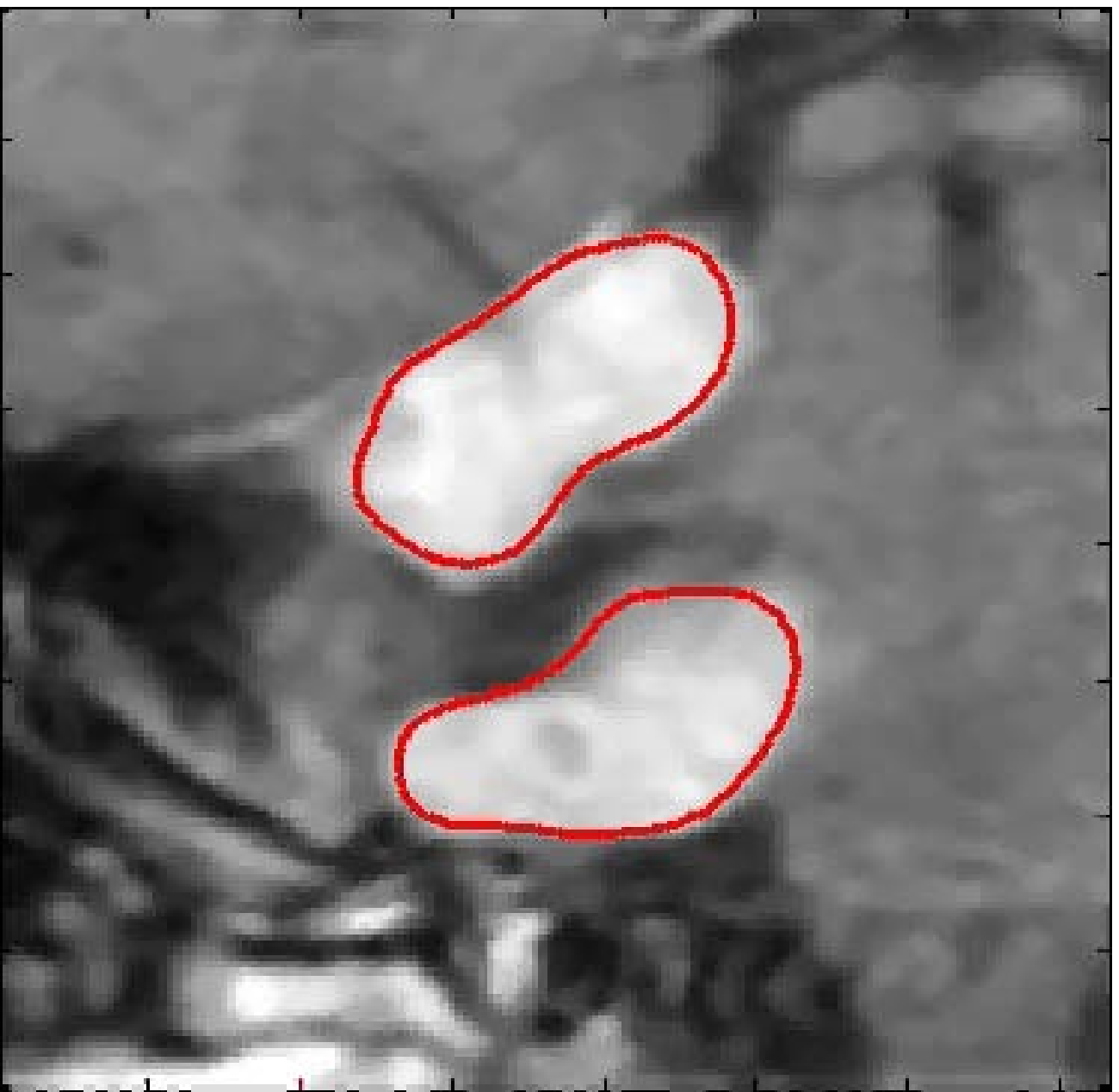}}}}
&{{\color{gray}\fbox{\includegraphics[width=0.15\textwidth,height = 1.1in]{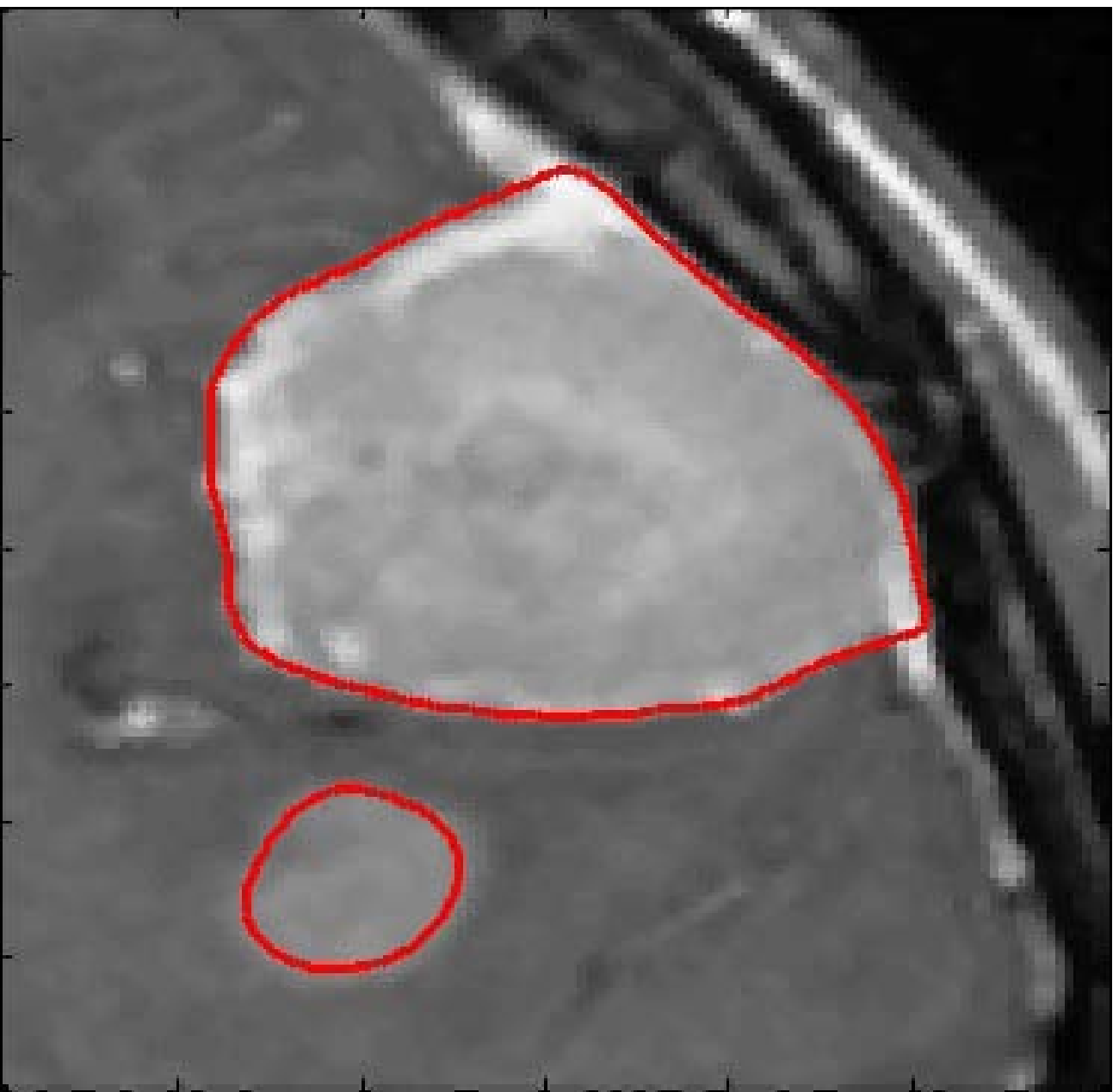}}}}\\
\end{tabular}
\caption{Object Extraction by proposed method. The top row is the curve evolution by GAC; The middle is the curve evolution by EF; The bottom row is the final convergence.}\label{Fig:Result_GACEF}
\end{figure*}

\begin{figure*}[!h]
\centering
\setlength{\tabcolsep}{2pt}
\begin{tabular}{cc|c|c|c|cc}
&GAC & GAC+Bl & GAC+AdaBl & MAC & Chan-Vese& \\
\midrule
&{\includegraphics[width=0.15\textwidth,height = 1.1in]{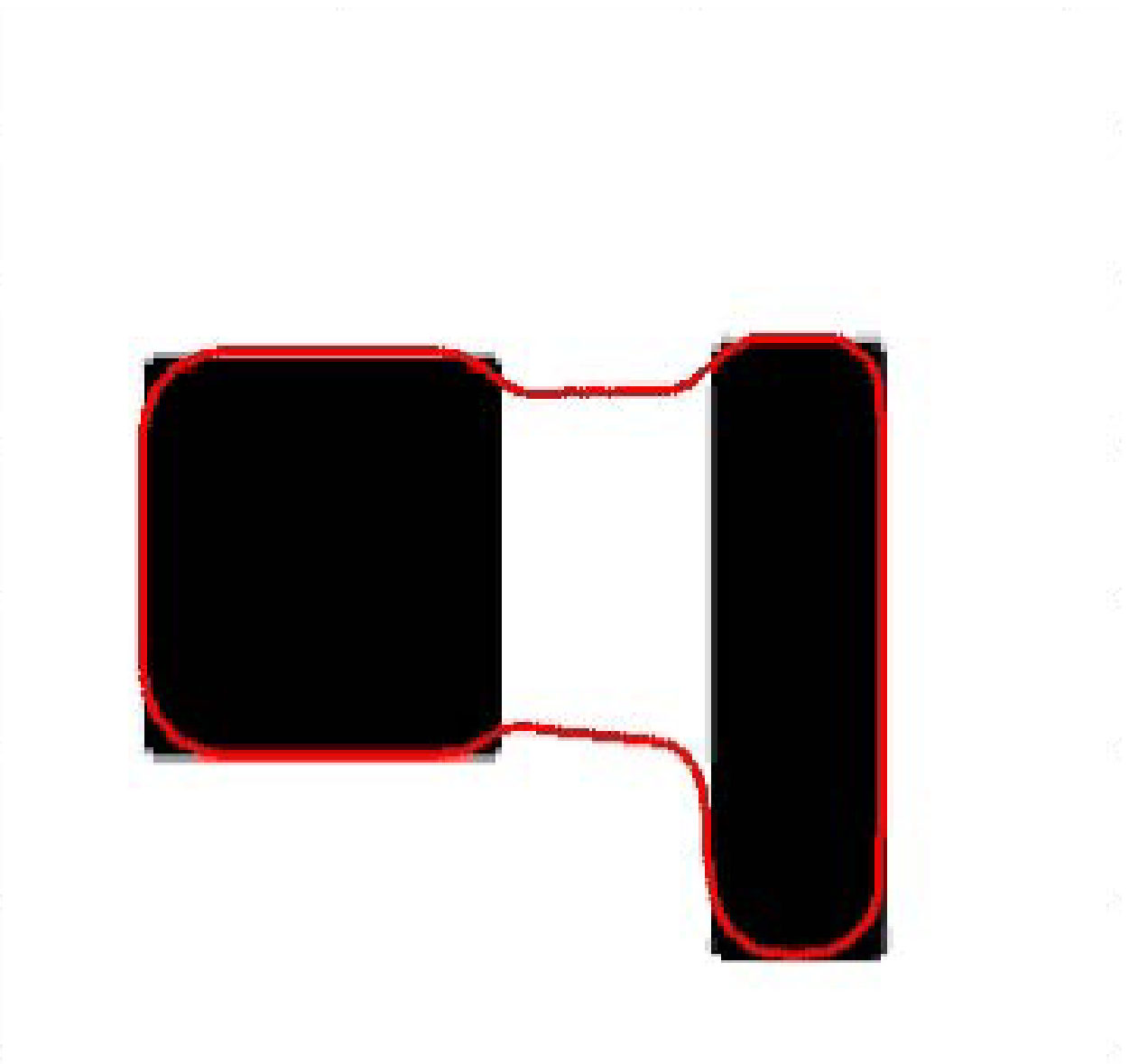}}&
{\includegraphics[width=0.15\textwidth,height = 1.1in]{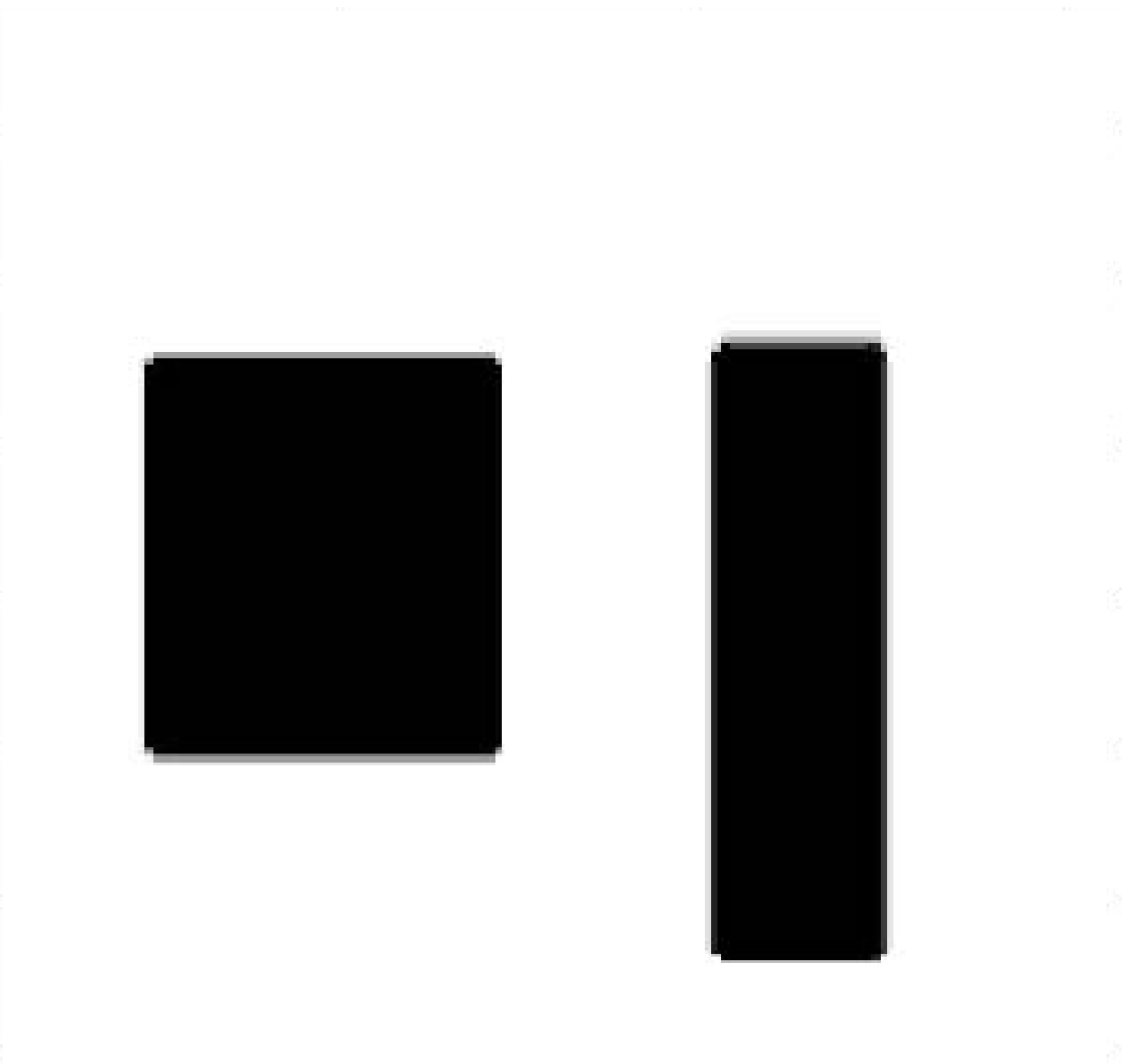}}&
{\includegraphics[width=0.15\textwidth,height = 1.1in]{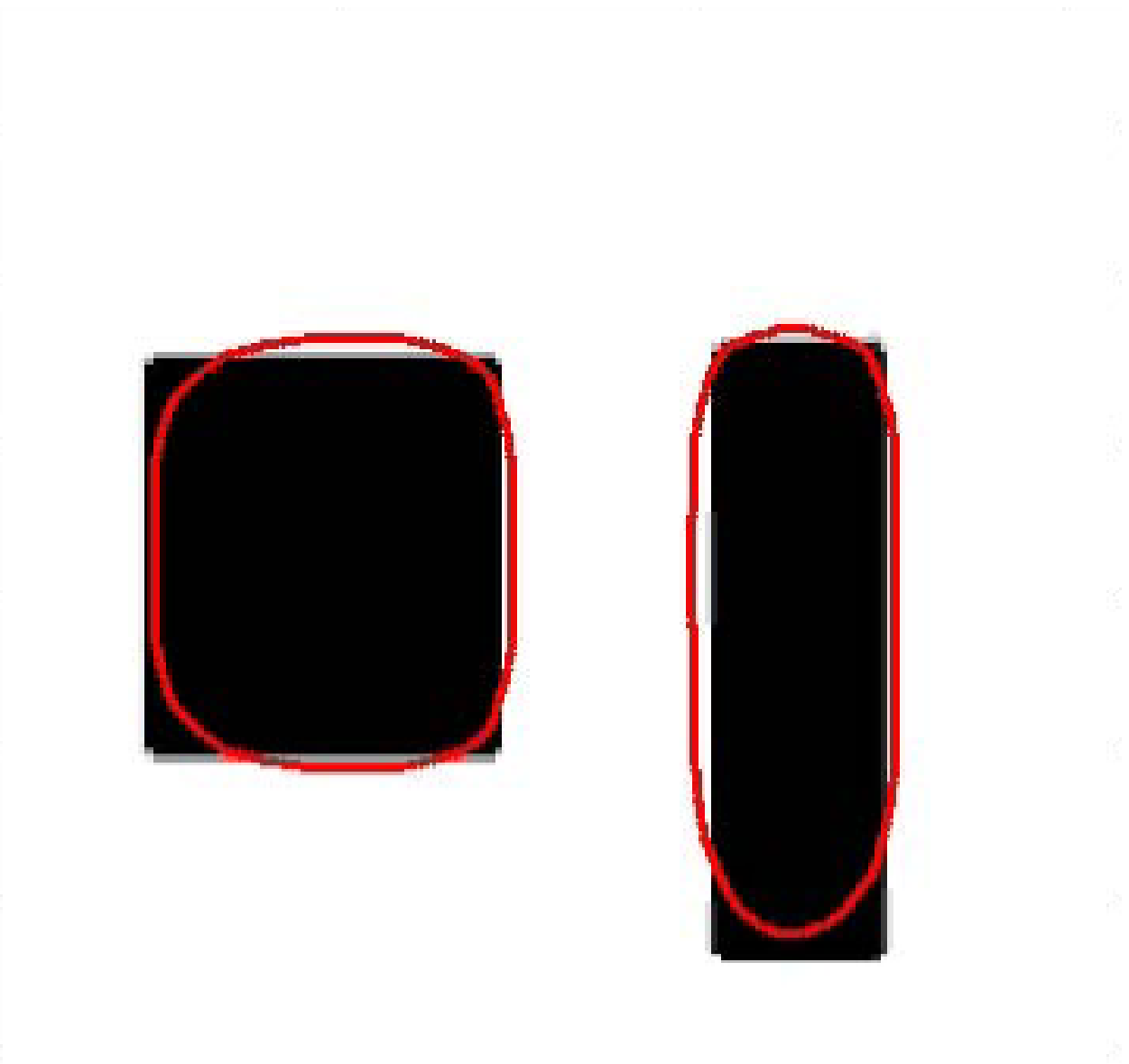}}&
{\includegraphics[width=0.15\textwidth,height = 1.1in]{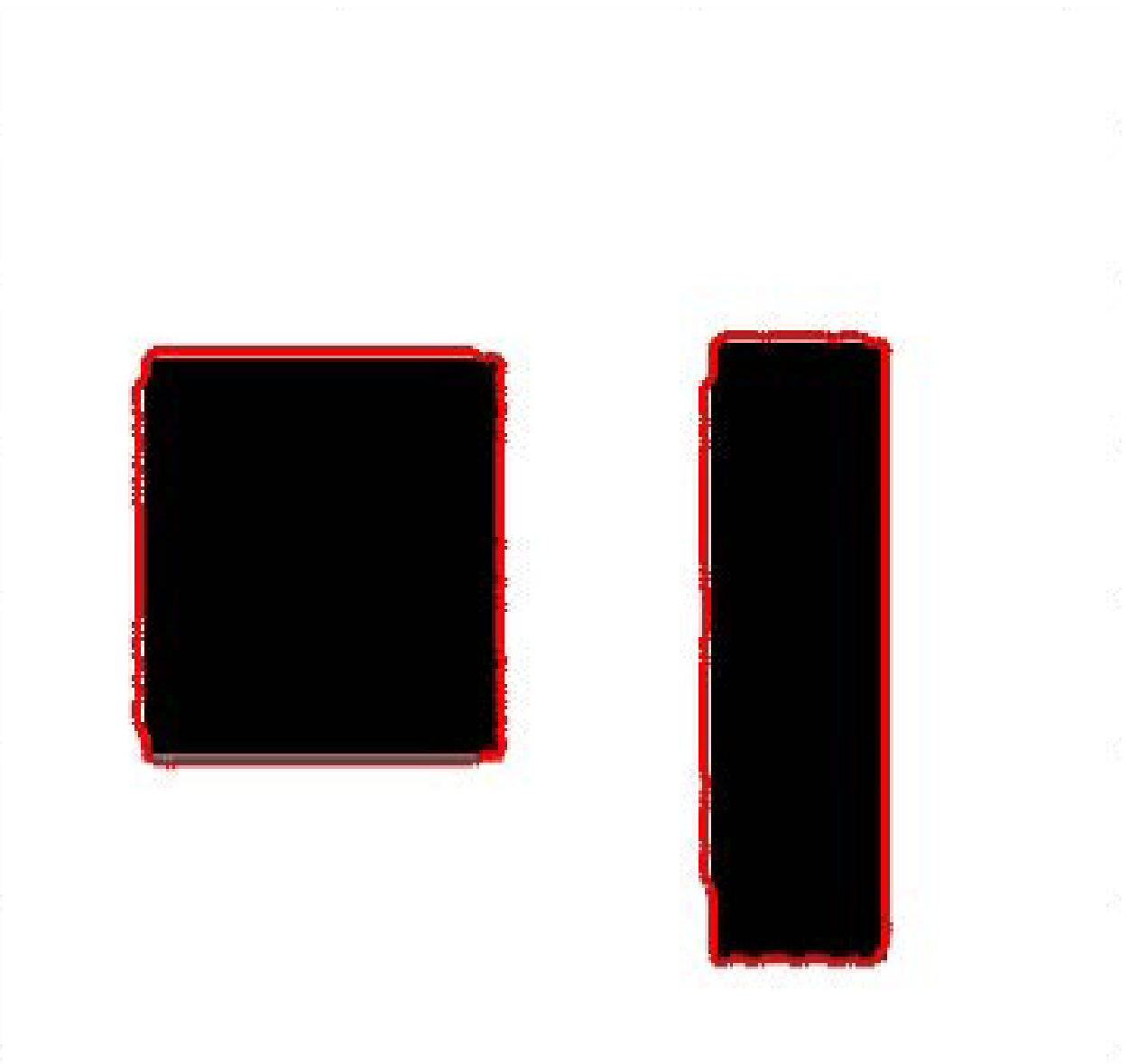}}&
{\includegraphics[width=0.15\textwidth,height = 1.1in]{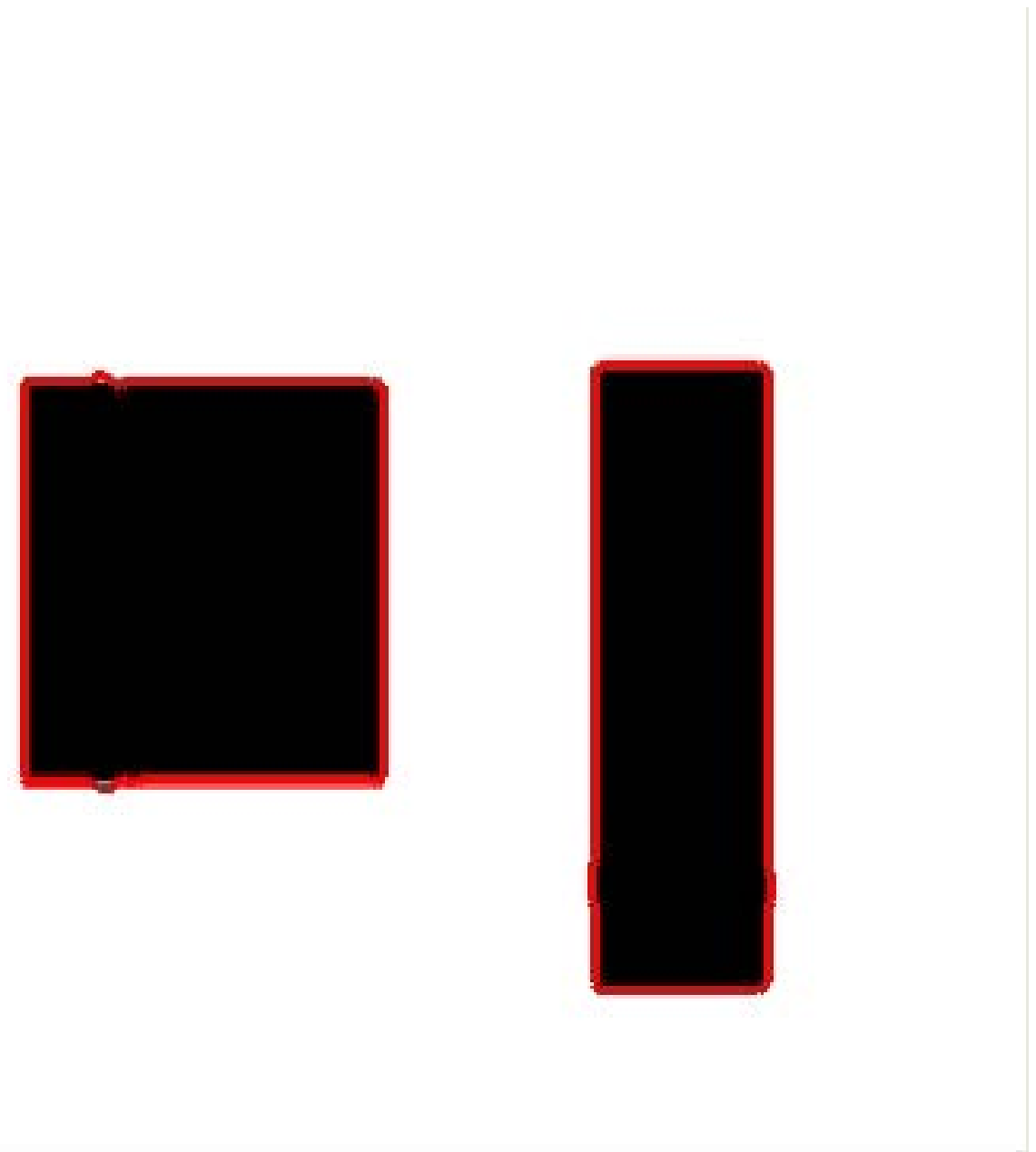}}&\\
\midrule
&{\includegraphics[width=0.15\textwidth,height = 1.1in]{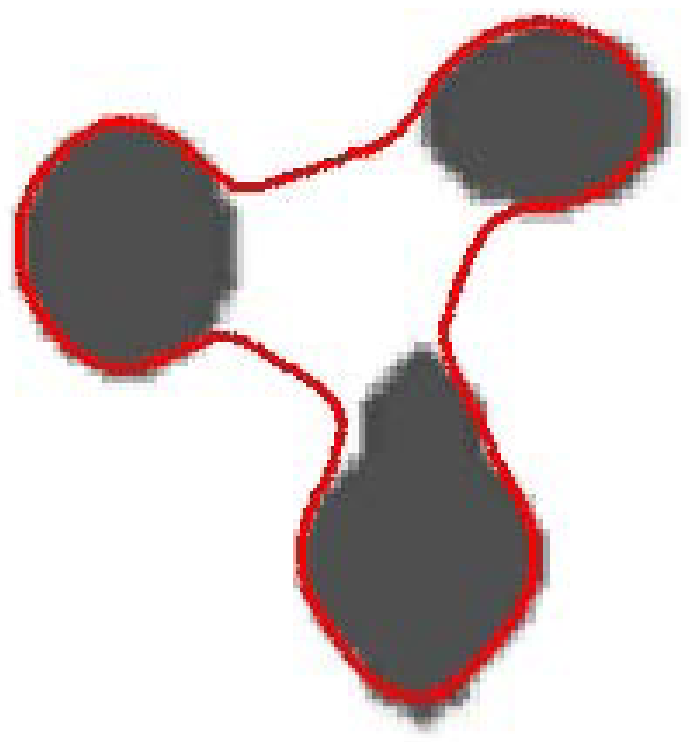}}&
{\includegraphics[width=0.15\textwidth,height = 1.1in]{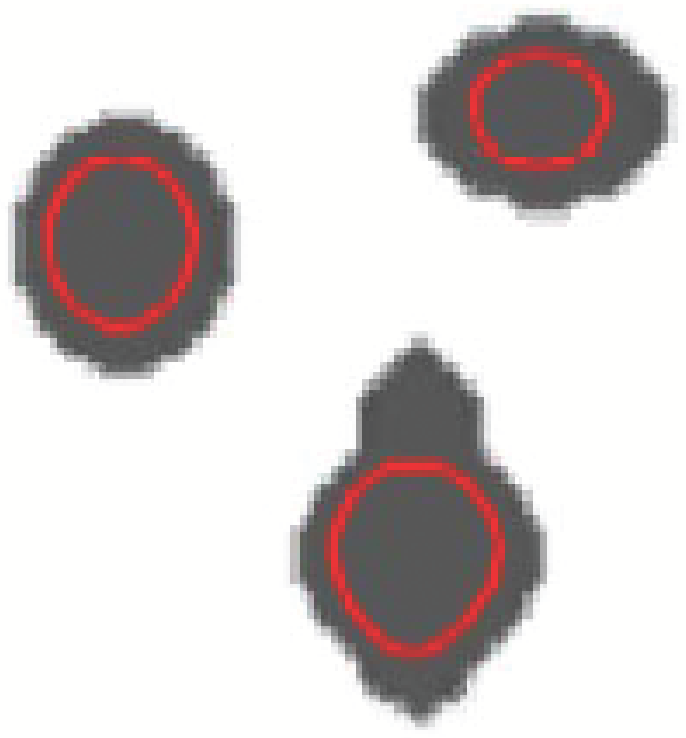}}&
{\includegraphics[width=0.15\textwidth,height = 1.1in]{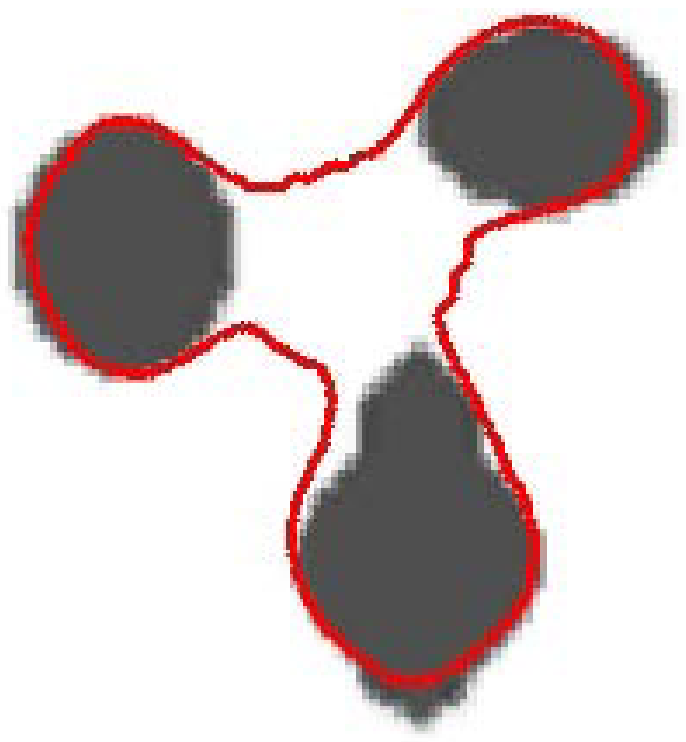}}&
{\includegraphics[width=0.15\textwidth,height = 1.1in]{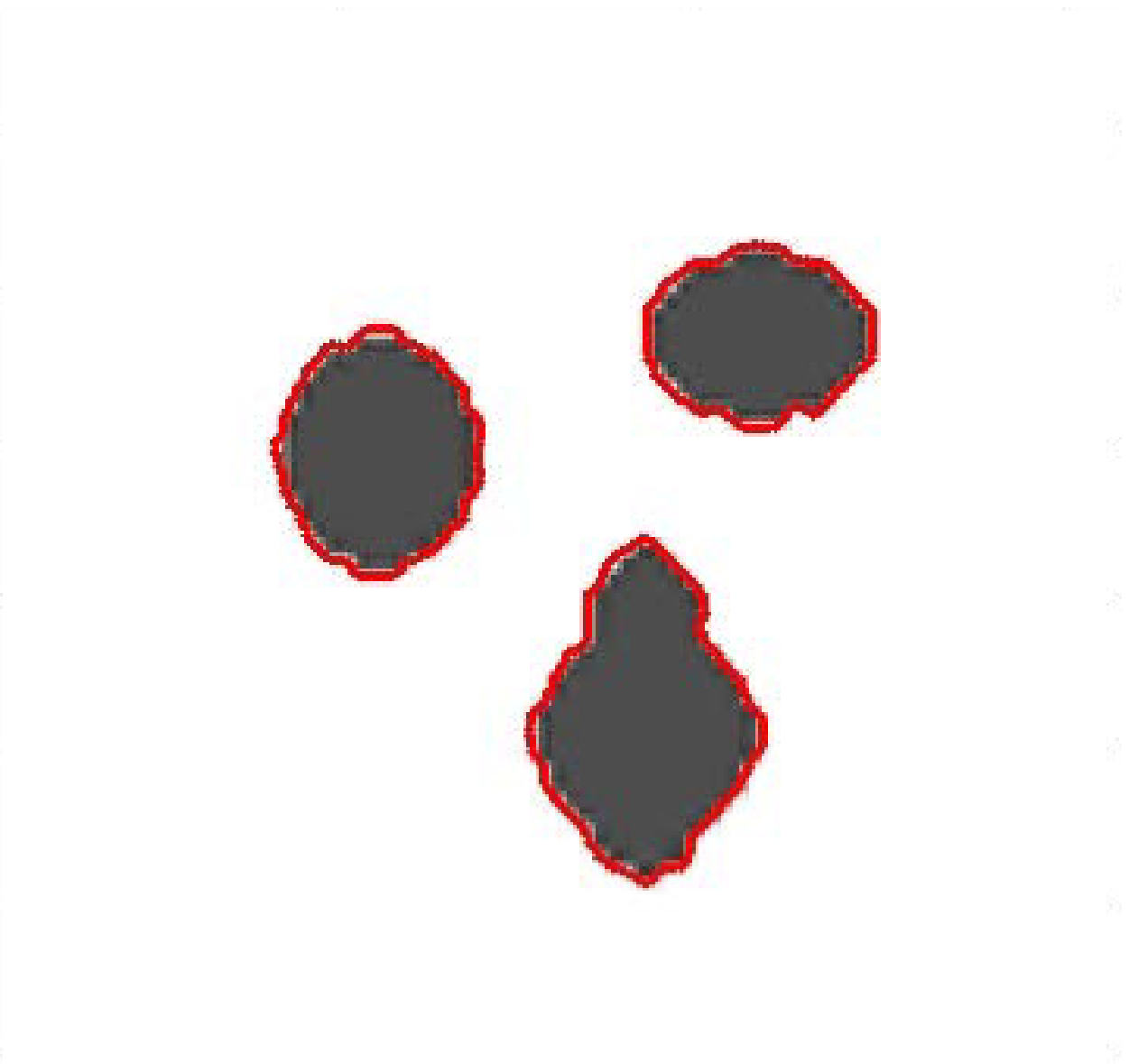}}&
{\includegraphics[width=0.15\textwidth,height = 1.1in]{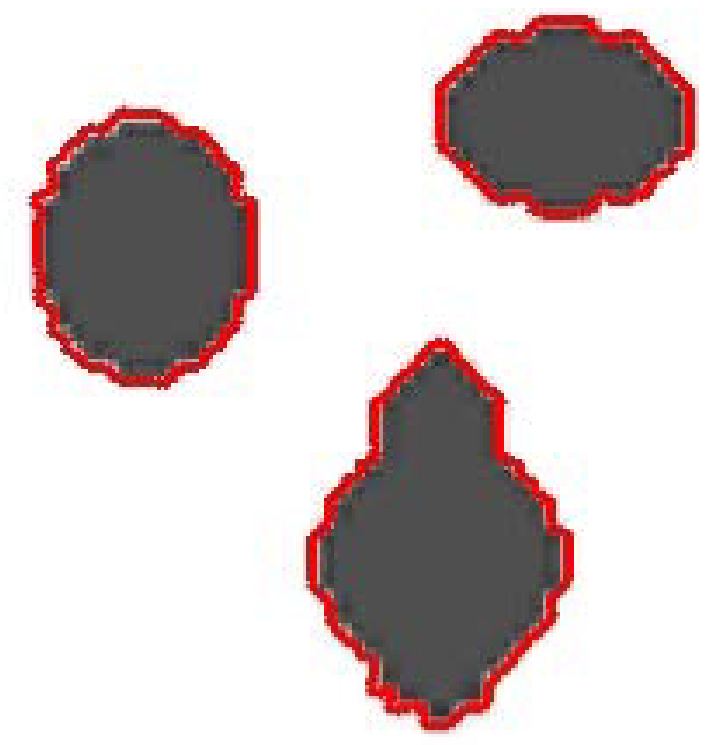}}&\\
\midrule
&{\includegraphics[width=0.15\textwidth,height = 1.1in]{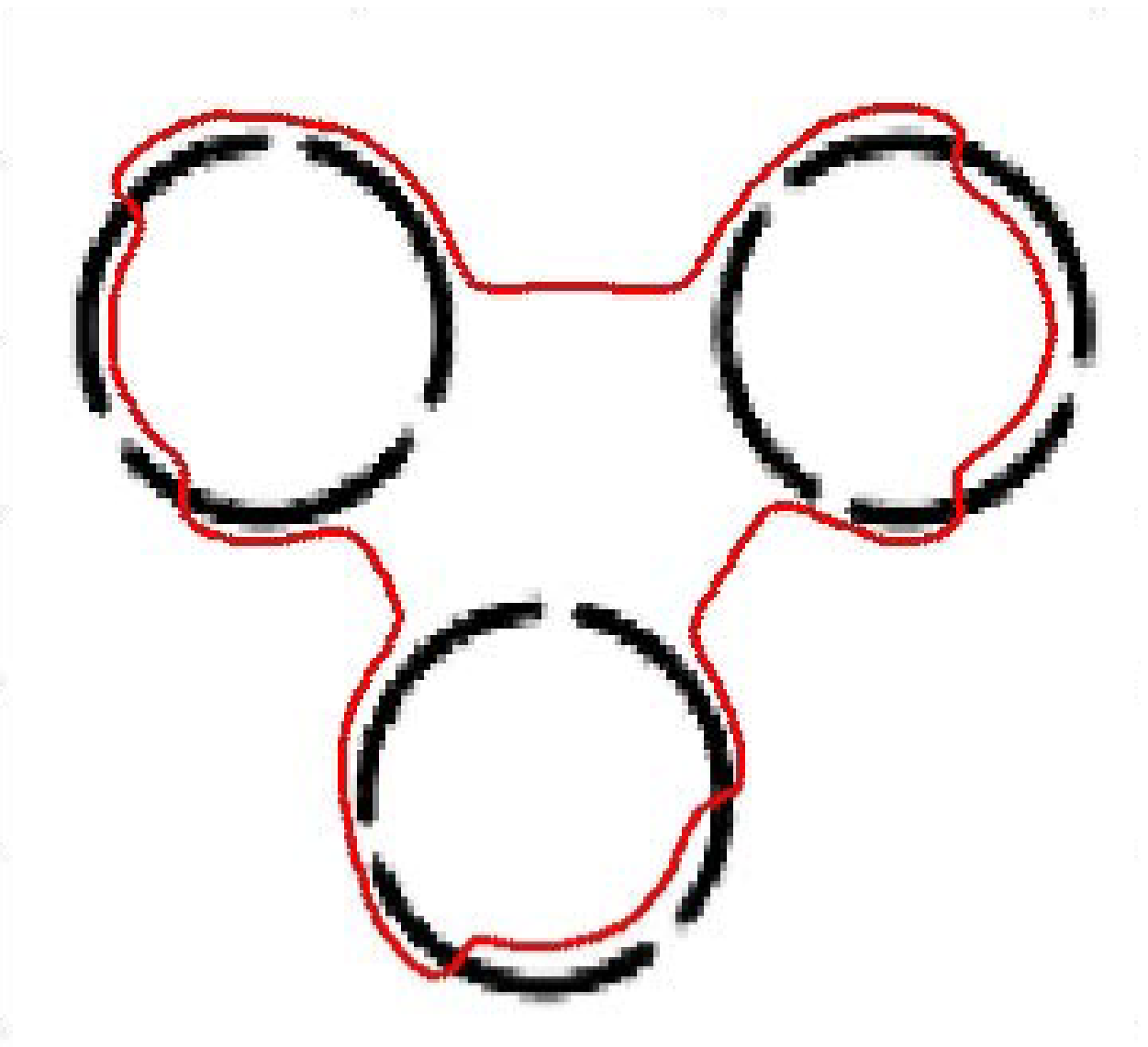}}&
{\includegraphics[width=0.15\textwidth,height = 1.1in]{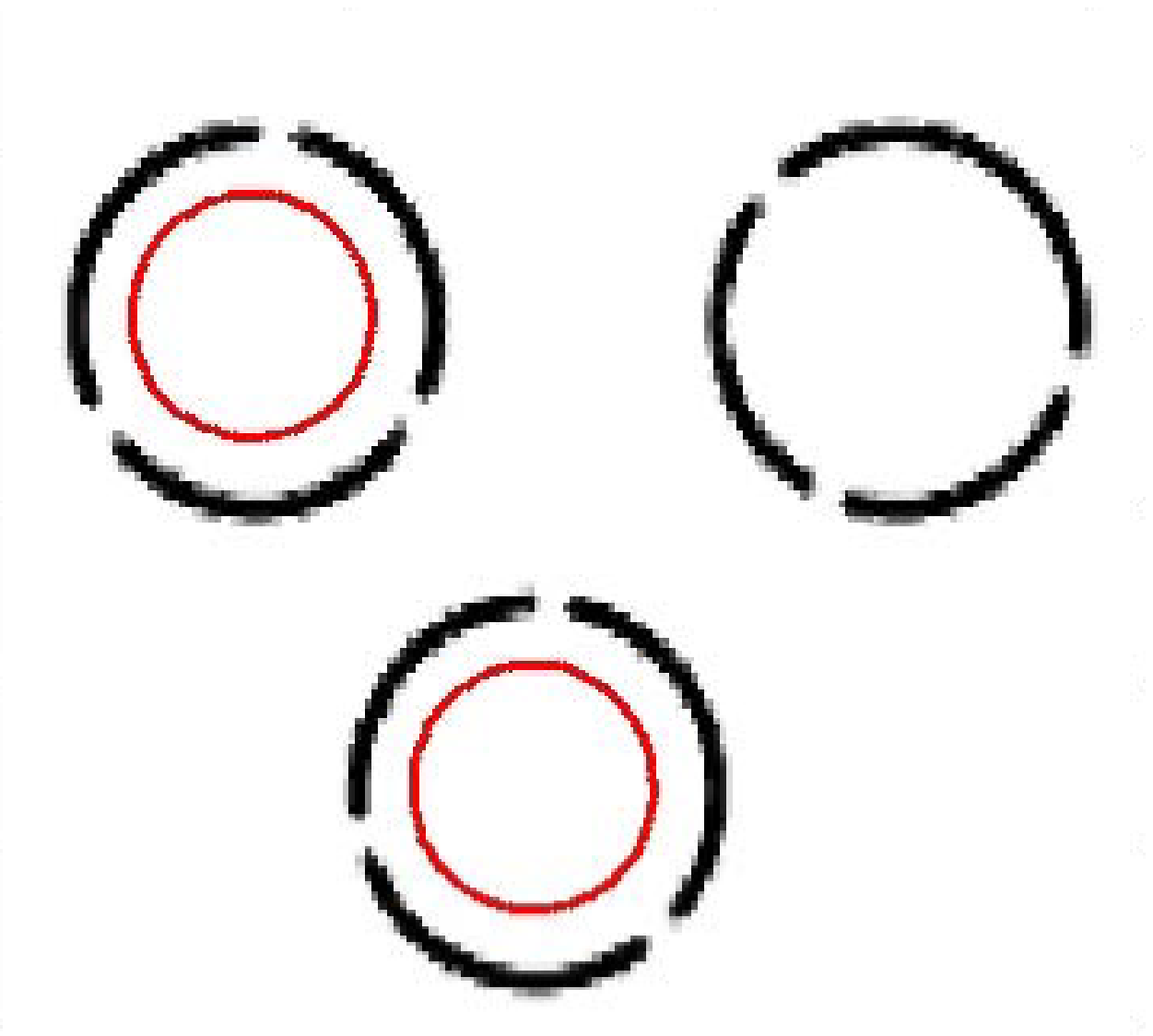}}&
{\includegraphics[width=0.15\textwidth,height = 1.1in]{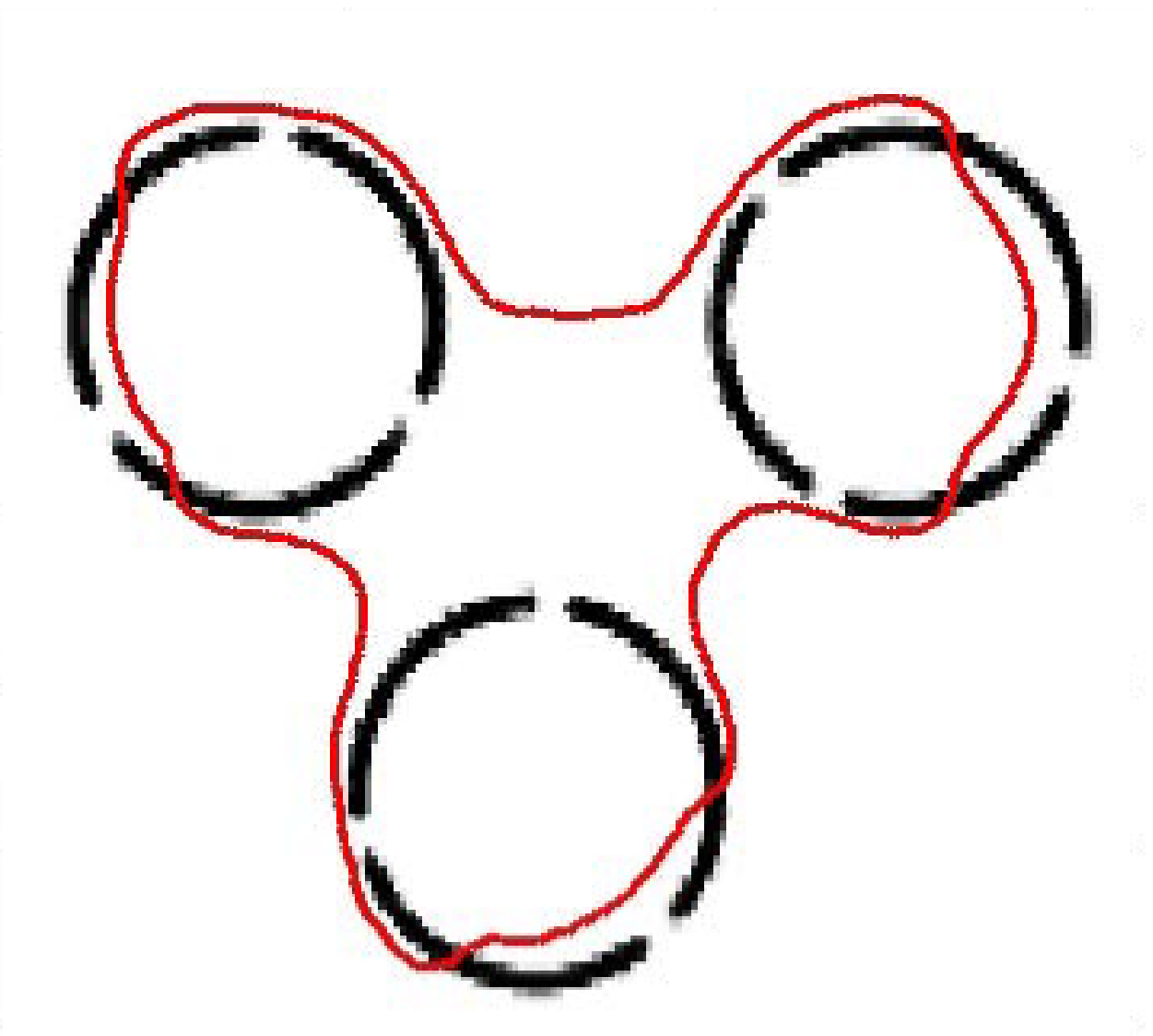}}&
{\includegraphics[width=0.15\textwidth,height = 1.1in]{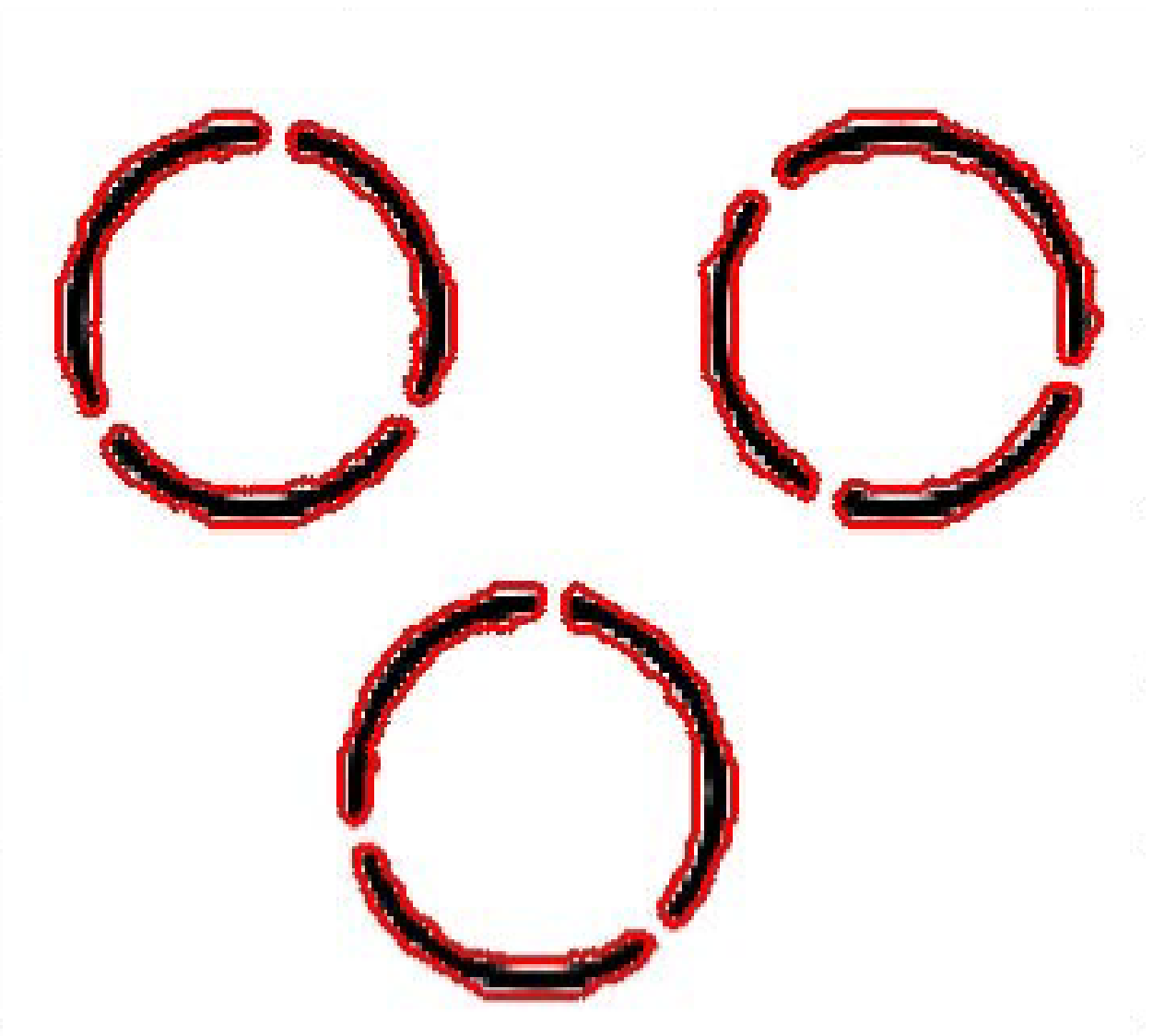}}&
{\includegraphics[width=0.15\textwidth,height = 1.1in]{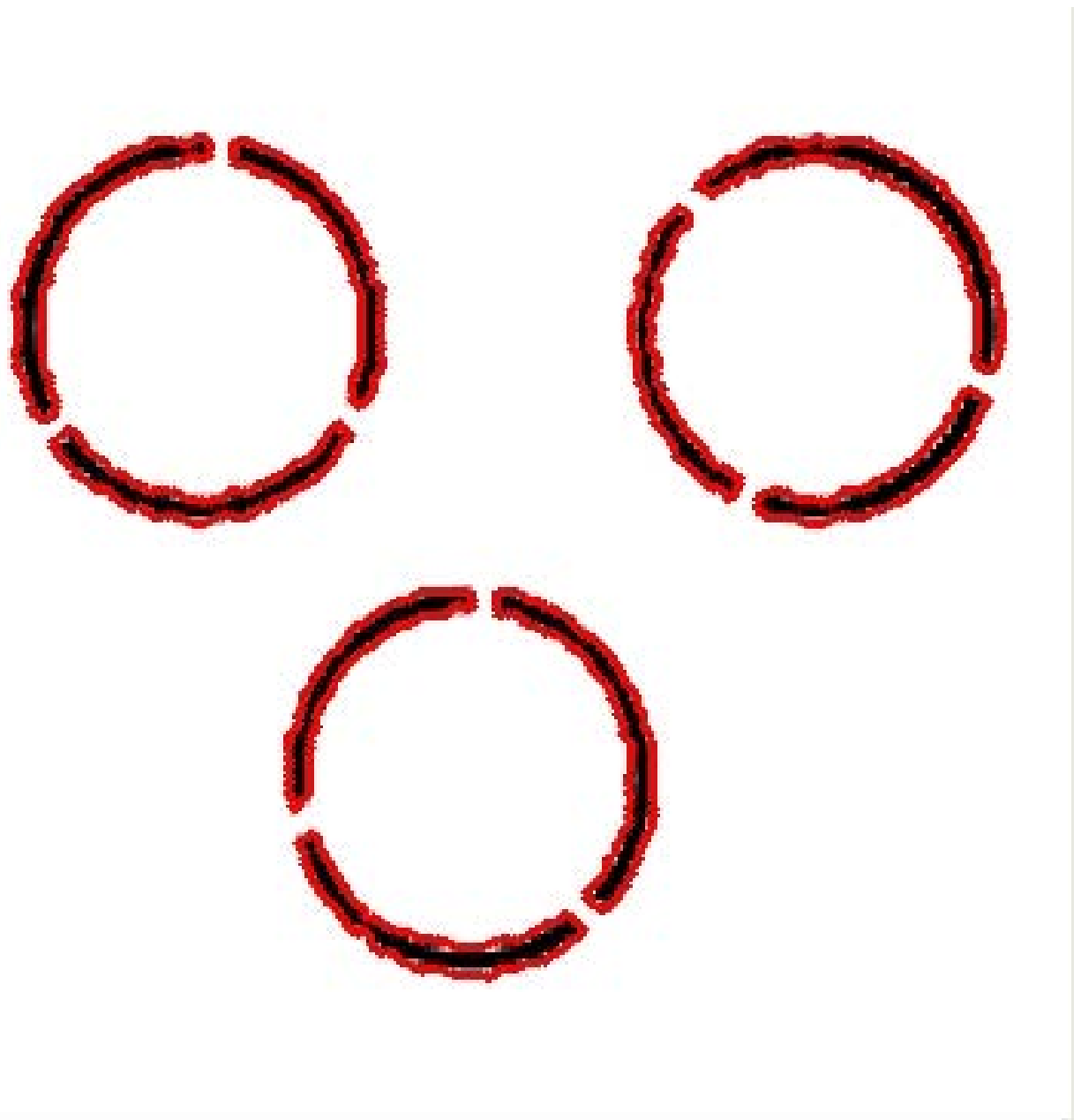}}&\\
\midrule
&{\includegraphics[width=0.15\textwidth,height = 1.1in]{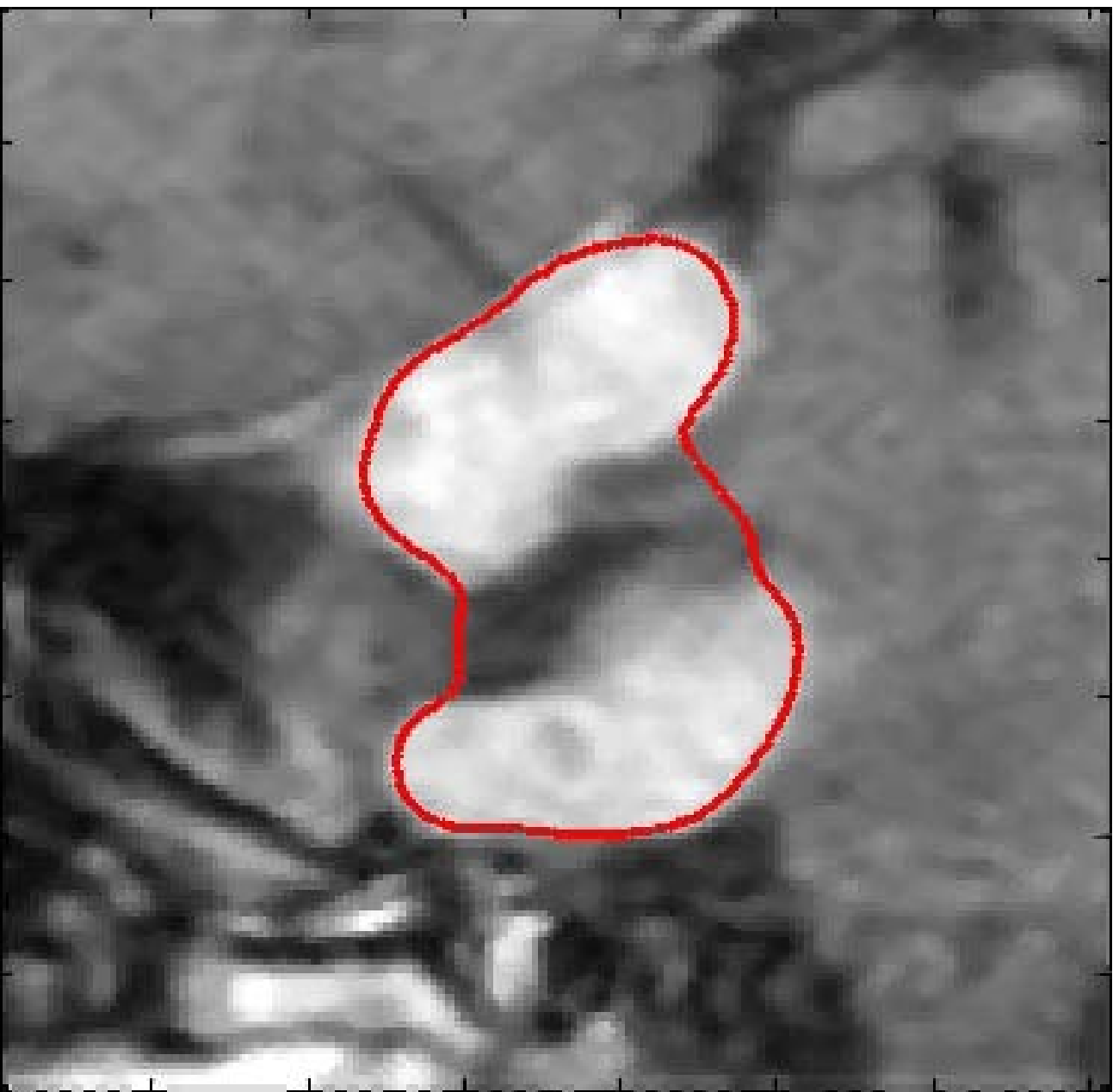}}&
{\includegraphics[width=0.15\textwidth,height = 1.1in]{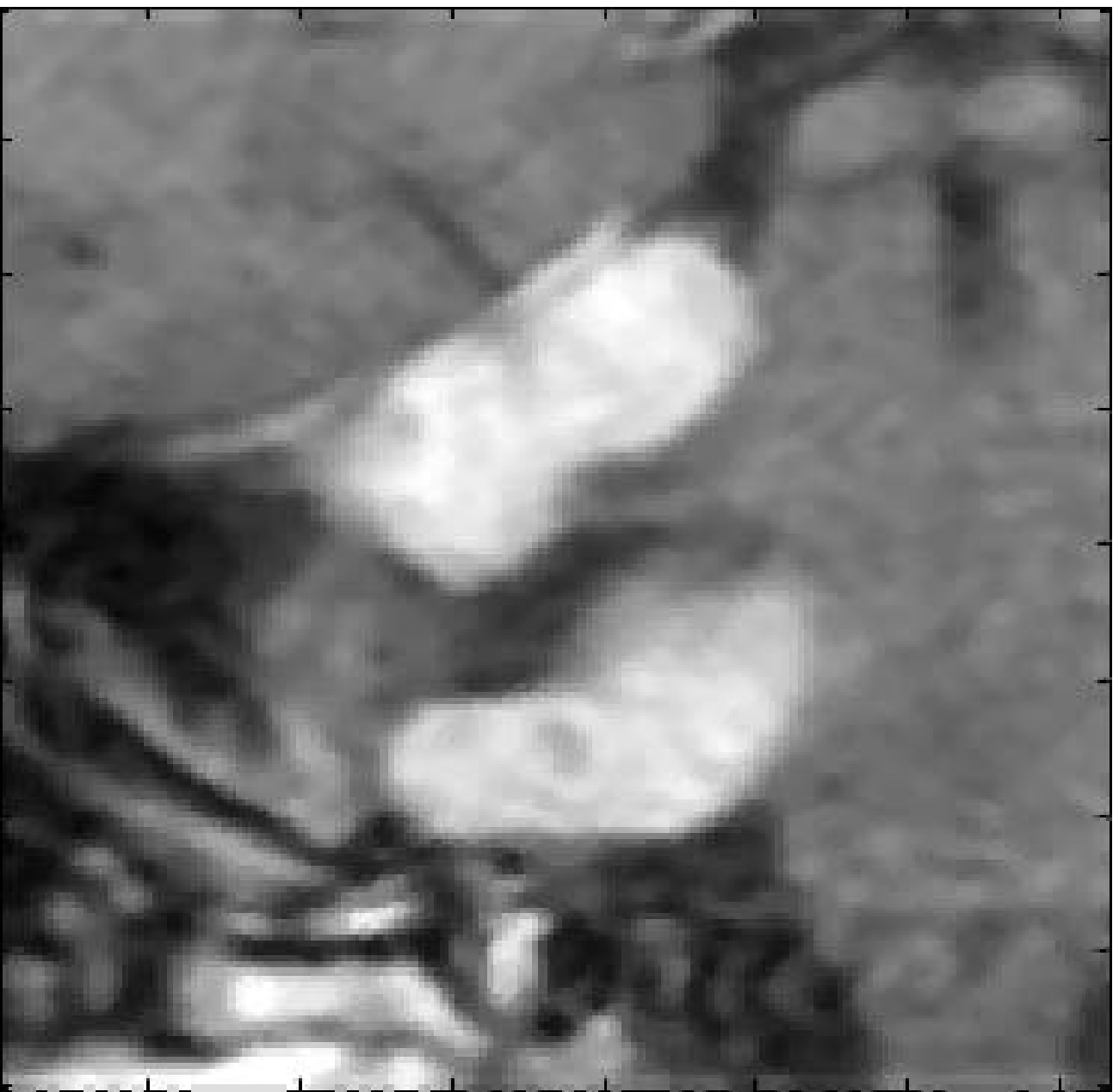}}&
{\includegraphics[width=0.15\textwidth,height = 1.1in]{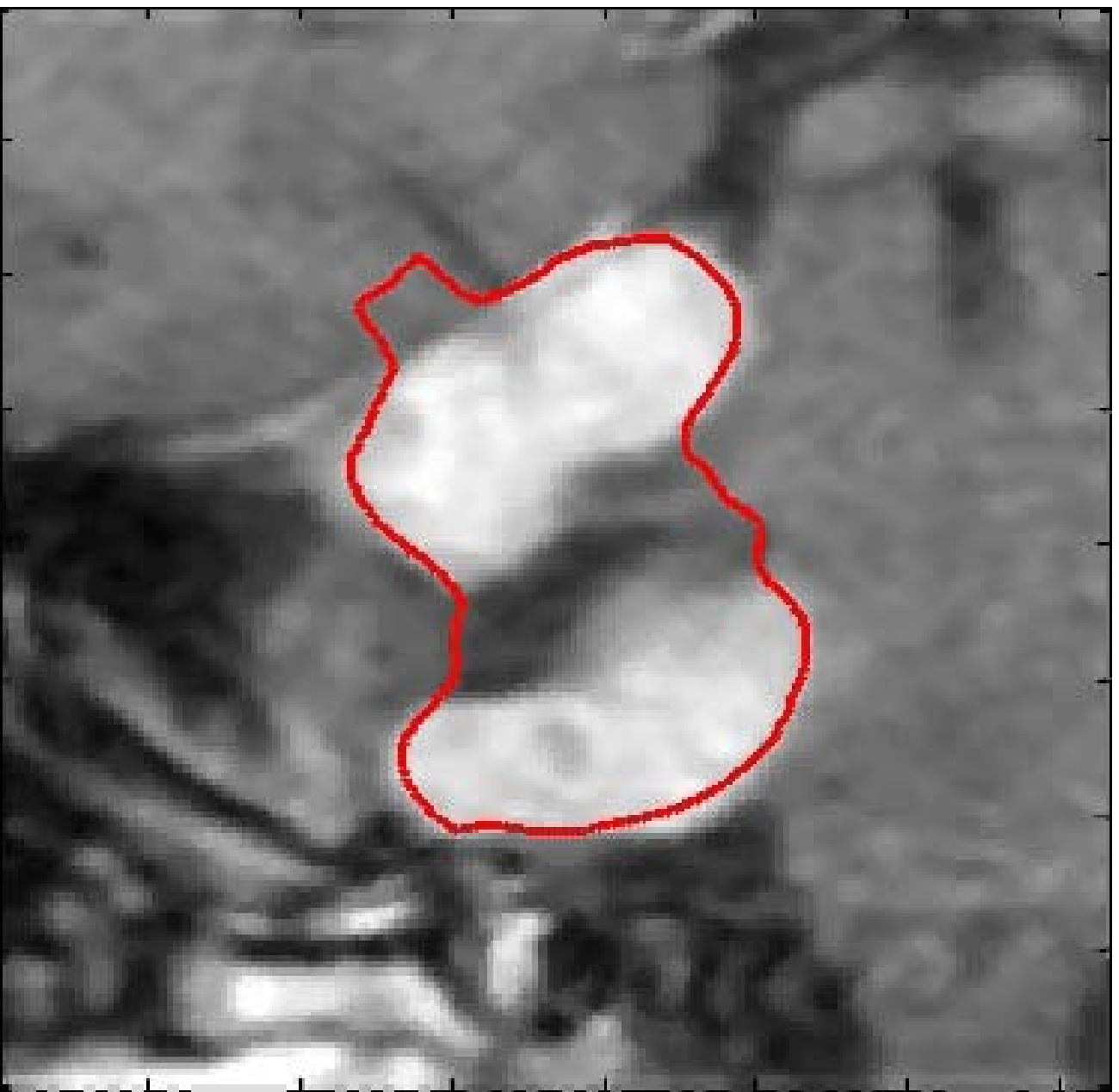}}&
{\includegraphics[width=0.15\textwidth,height = 1.1in]{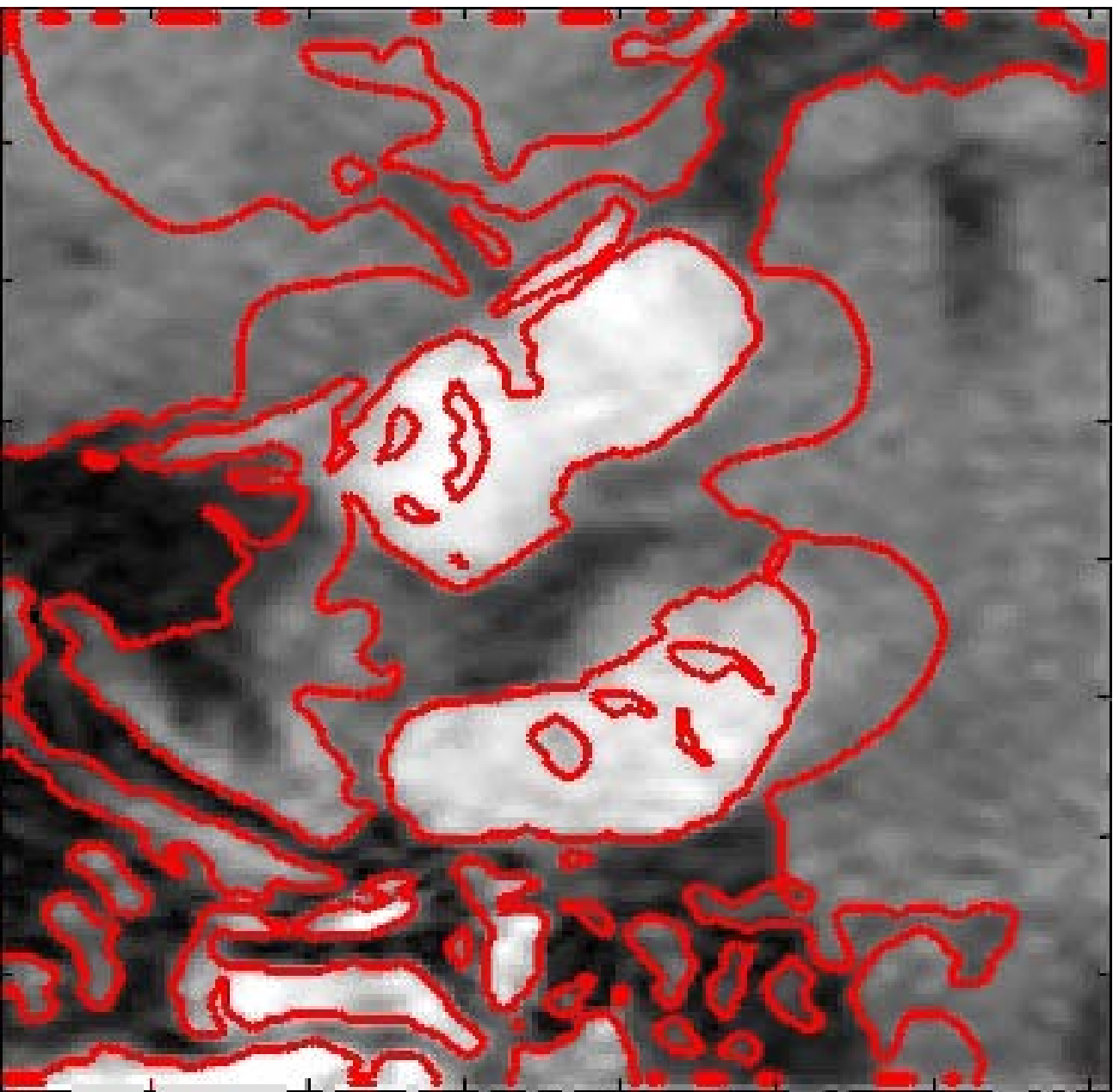}}&
{\includegraphics[width=0.15\textwidth,height = 1.1in]{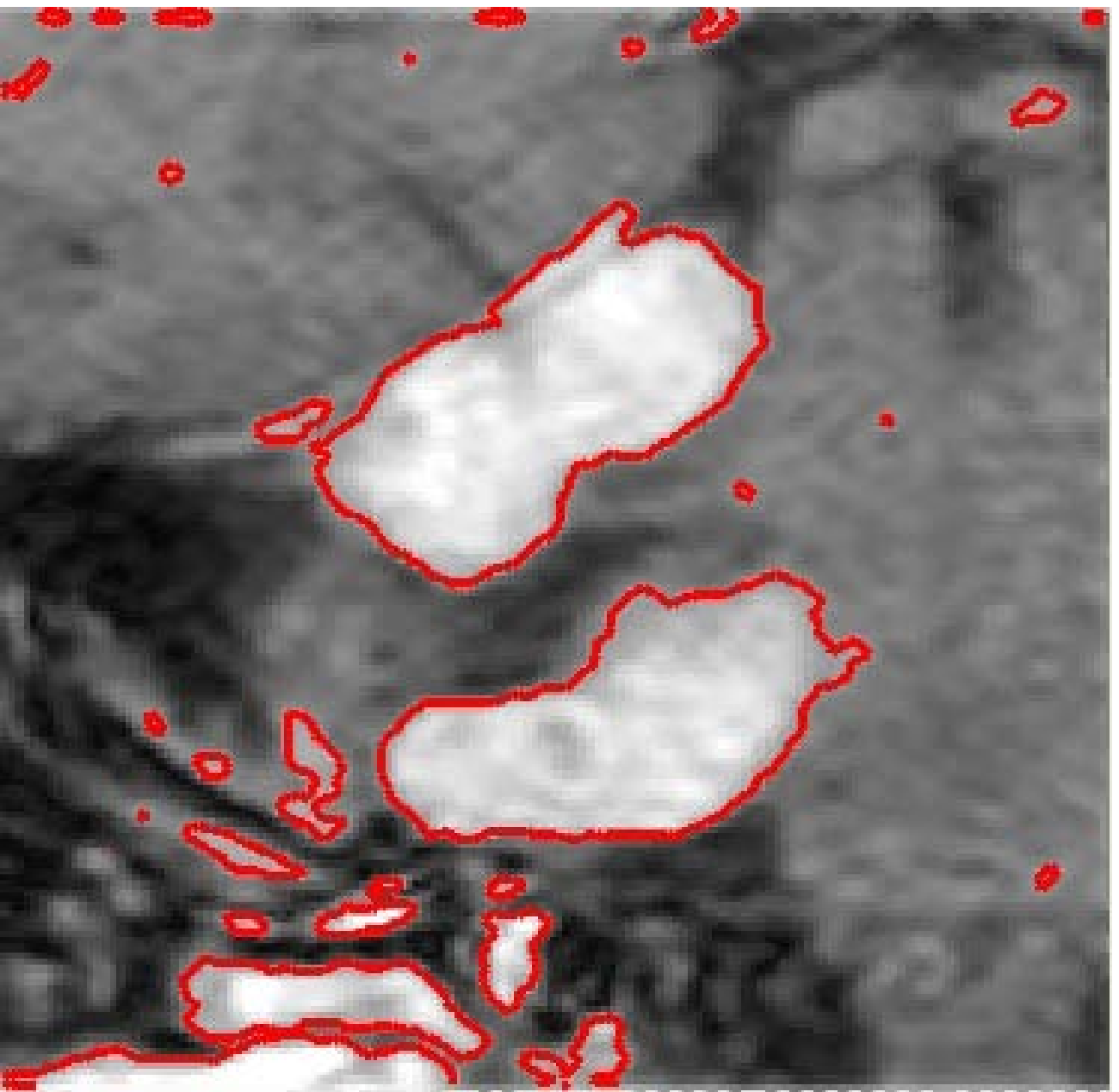}}&\\
\midrule
&{\includegraphics[width=0.15\textwidth,height = 1.1in]{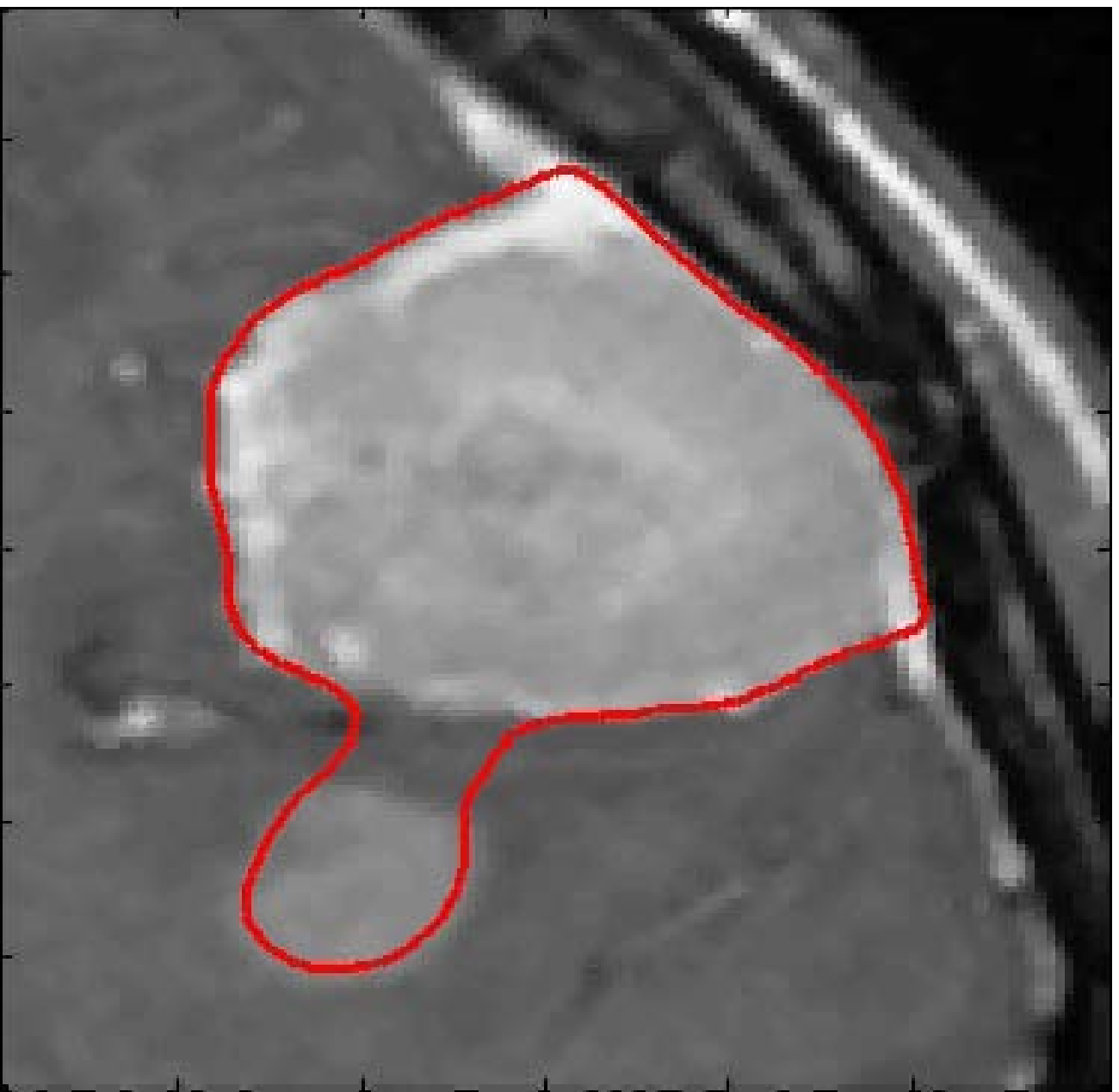}}&
{\includegraphics[width=0.15\textwidth,height = 1.1in]{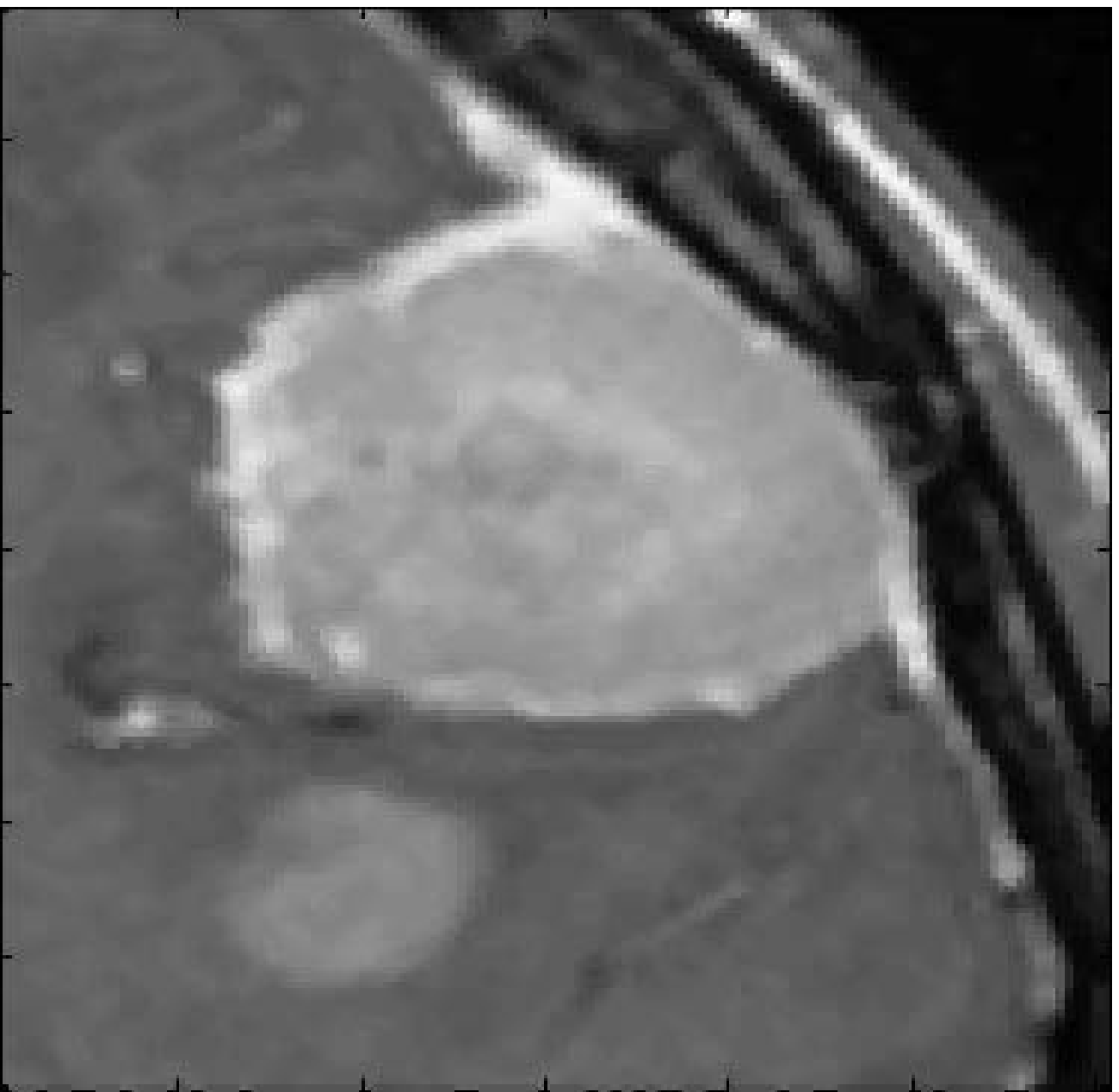}}&
{\includegraphics[width=0.15\textwidth,height = 1.1in]{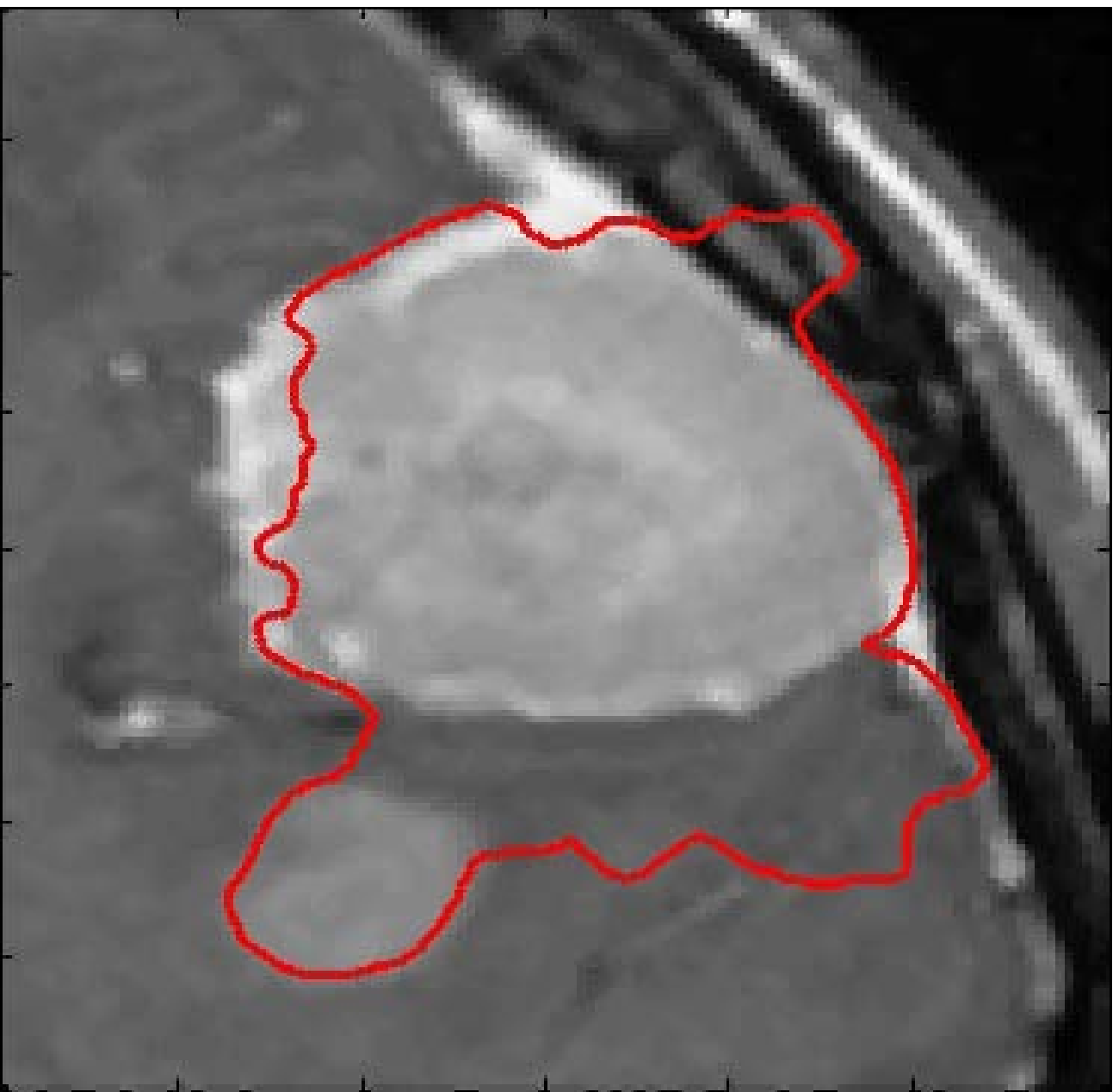}}&
{\includegraphics[width=0.15\textwidth,height = 1.1in]{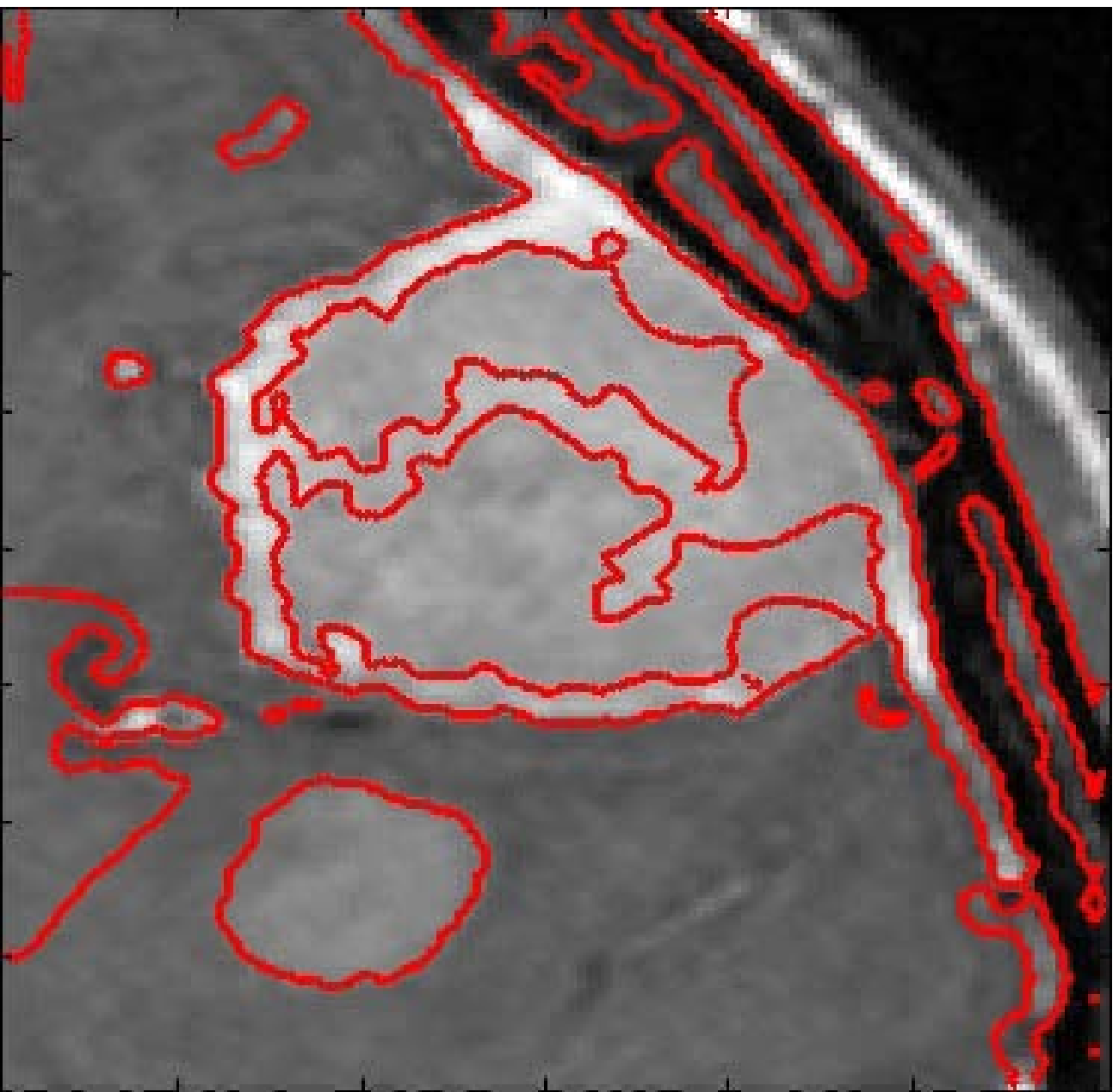}}&
{\includegraphics[width=0.15\textwidth,height = 1.1in]{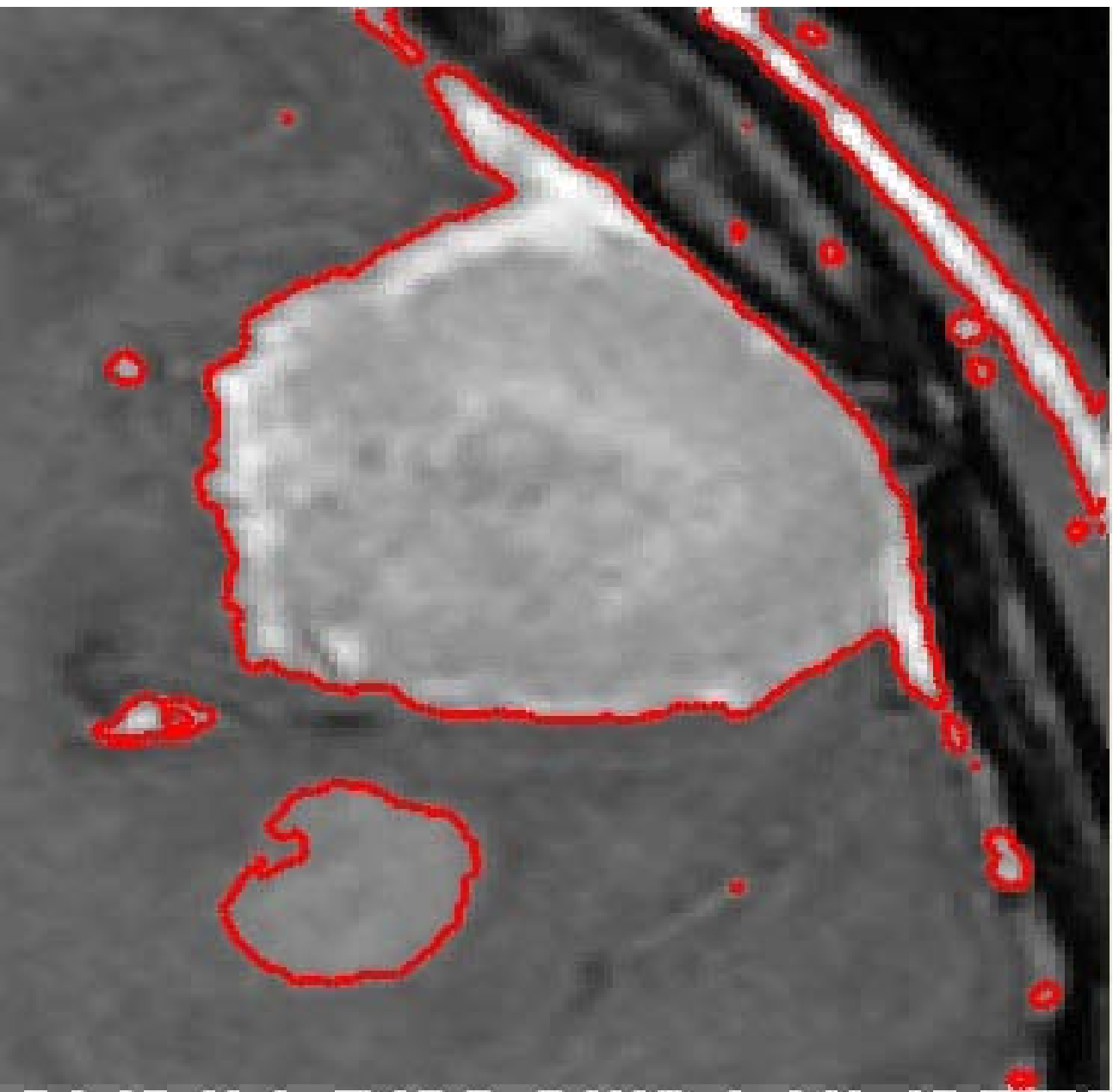}}&\\
\hline
\end{tabular}
\caption{Results by other methods. First column from the left is the object extraction by GAC; Second column is the object extraction by GAC with Balloon; Third column is the object extraction by GAC with adaptive Balloon; Fourth column is the object extraction by MAC; Last column is the object extraction by Chan-Vese active contour.}\label{Fig:Others}
\end{figure*}

The curve evolution process of the system of alternating curve evolutions following (\ref{EQS:ALT}), called the GAC+EF, are shown in Figure \ref{Fig:Result_GACEF}. The objects are correctly segmented. Figure \ref{Fig:Others} shows the segmentation results produced by other methods, including the GAC, GAC with Balloon, GAC with adaptive Balloon, MAC and the Chan-Vese model, with the same initialization shown in \ref{Fig:Init_inputs}. It can be observed that none of these compared methods can extract all the objects accurately like GAC+EF. GAC suffers from the early stop of the curve evolution. GAC with Balloon can surpass edges and converge to the interior boundaries or vanish; GAC with adaptive Balloon can be applied for extracting 2 objects of simple topology as shown in Figure \ref{Fig:Others}, yet it still suffers from the PSP in other images. The gradient field used for the former three methods are extended by Gradient Vector Flow. Interestingly, both the GAC and the GAC with the adaptive Balloon might stop quite close to the Pseudo Stationary Positions as shown in Figure \ref{Fig:PSPshow}, while GAC with adaptive Balloon can escape from some Pseudo Stationary Positions of relatively simple topology. It can also be observed that both MAC and Chan-Vese model can extract the narrow curved regions in the image of three disconnected circles shown in Figure \ref{Fig:Others}. It implies that MAC shares certain characteristics of the region based methods. The edge detection process in MAC does not produce a local gradient field but the indication of intensity homogeneity. The Chan-Vese active contour is not sensitive to initialization but it often captures other undesired regions. The MAC generally fails to converge to the object boundaries if there are too many edges and corners distributed everywhere in the images.

\begin{figure}[!h]
\centering
  \includegraphics[width=0.4\textwidth]{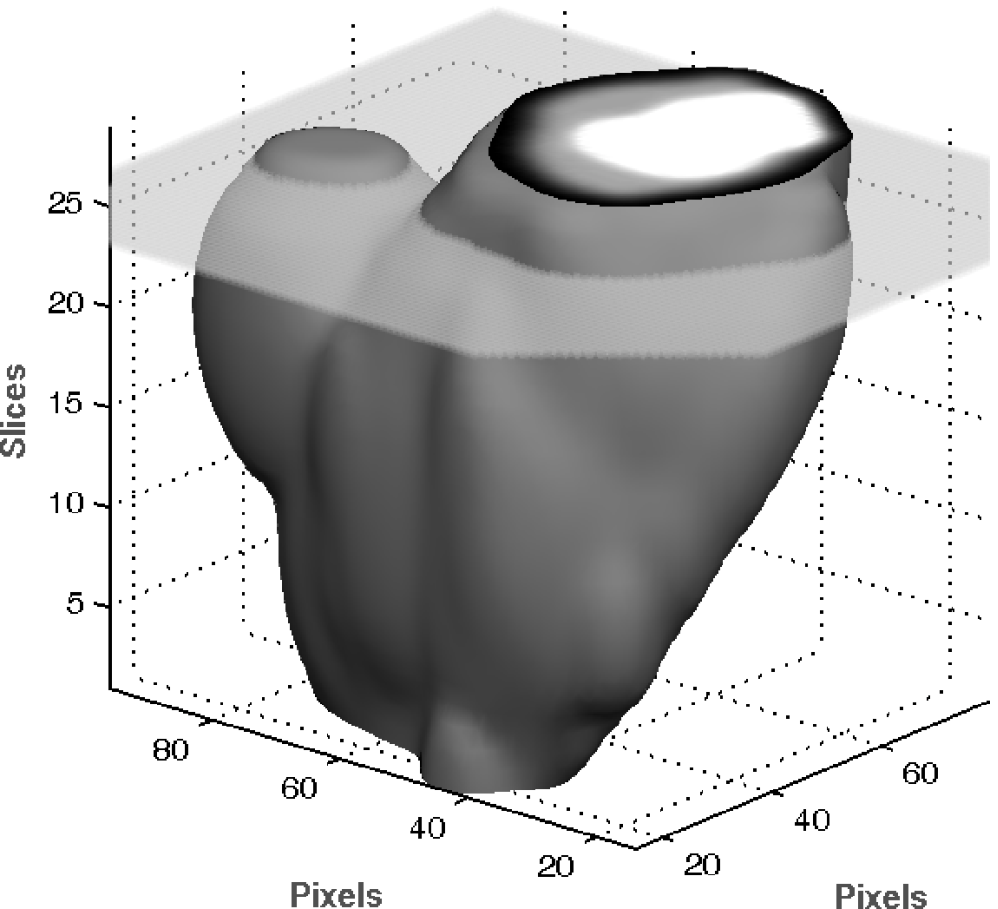}
  \caption{The reconstructed 3D tumor from the MRI sequence of Figure \ref{Fig:Init_inputs}(e). The cutting plate corresponds to the segmentation result shown in the right most figure in Figure \ref{Fig:Result_GACEF}(c) }\label{Fig:3D}
\end{figure}

Figure \ref{Fig:3D} visualizes the 3D tumor model that is obtained by the segmentation of the entire MRI sequence of Figure \ref{Fig:Init_inputs}(e) using GAC+EF. The cutting plate corresponds to the segmentation result shown in the bottom right corner in Figure \ref{Fig:Result_GACEF}.

\subsection{The tangential and normal velocities}

\begin{figure*}[!h]
\centering
\begin{tabular}{c|c|c|c|c}
{\includegraphics[width=0.5in,height = 0.5in]{sb2_GAC00_frame_0000.eps}}&
{\includegraphics[width=0.5in,height = 0.5in]{three_GAC00_frame_0000.eps}}&
{\includegraphics[width=0.5in,height = 0.5in]{new3s_GAC0_frame_0000.eps}}&
{\includegraphics[width=0.5in,height = 0.5in]{1_GAC0_frame_0000.eps}}&
{\includegraphics[width=0.5in,height = 0.5in]{two_GAC0_frame_0000.eps}}\\\hline
\vspace{-10pt}
{\includegraphics[height=0.135\textwidth,width=0.31\textheight,angle=270]{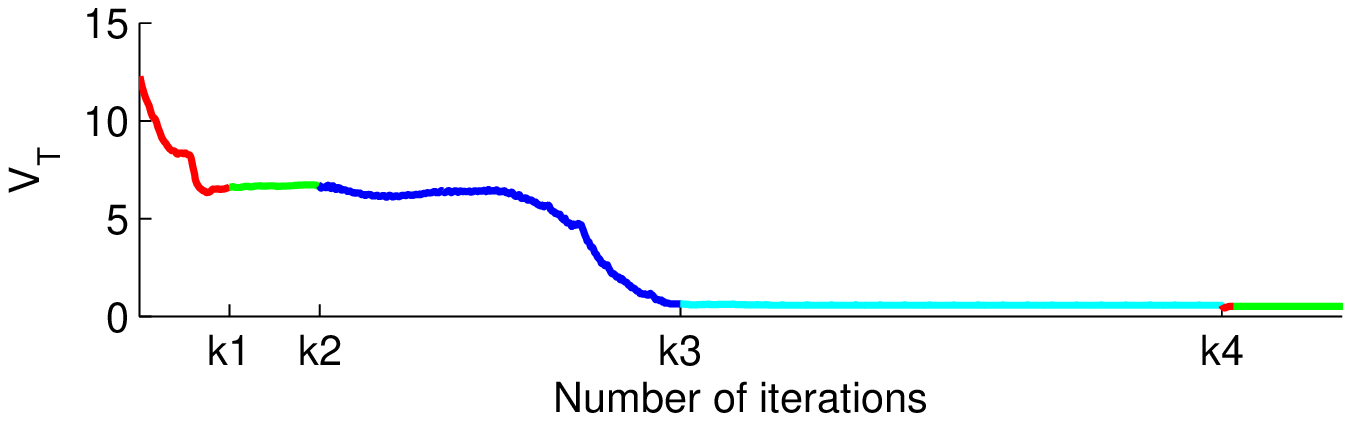}}&
{\includegraphics[height=0.135\textwidth,width=0.31\textheight,angle=270]{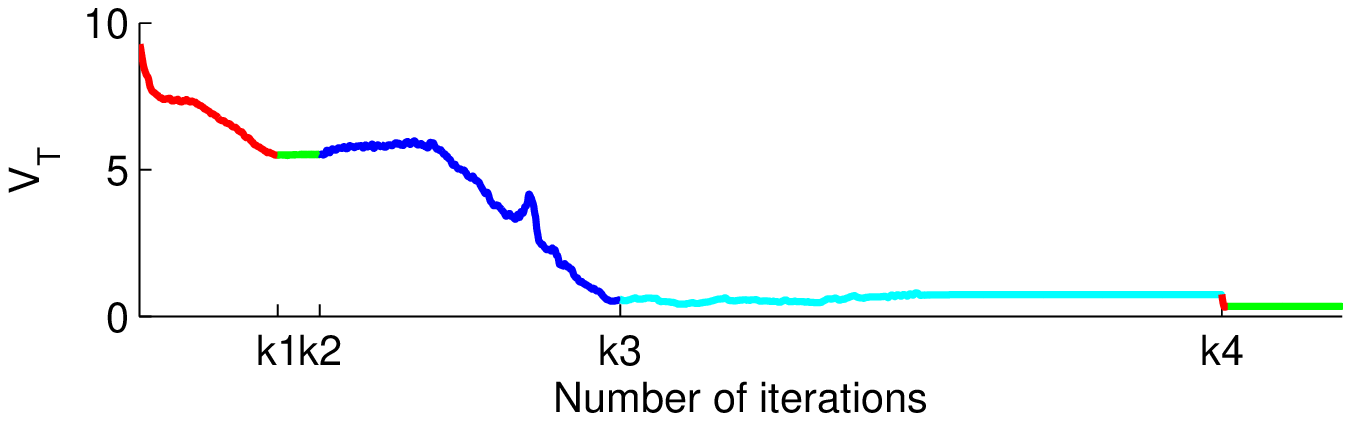}}&
{\includegraphics[height=0.135\textwidth,width=0.31\textheight,angle=270]{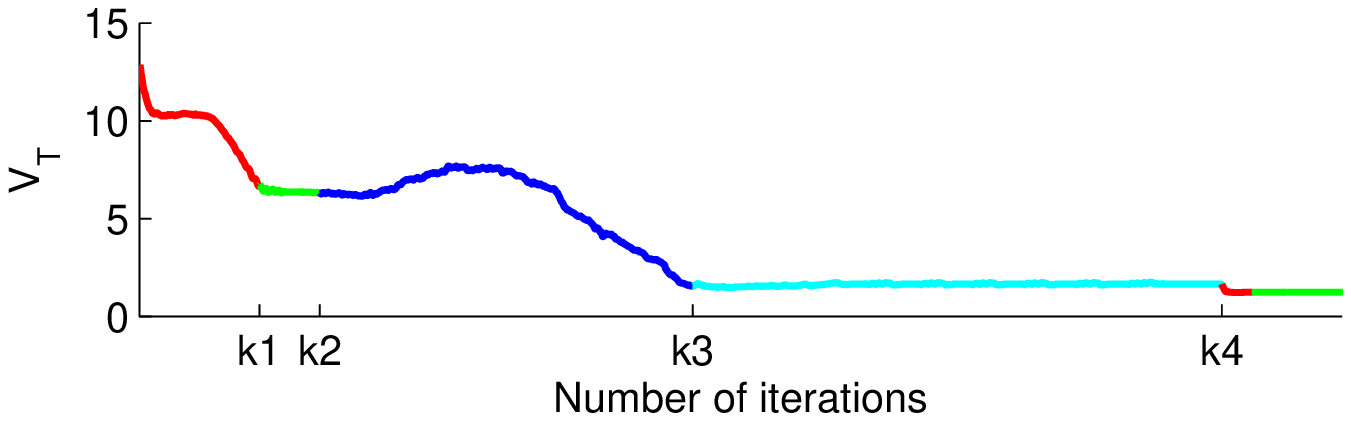}}&
{\includegraphics[height=0.135\textwidth,width=0.31\textheight,angle=270]{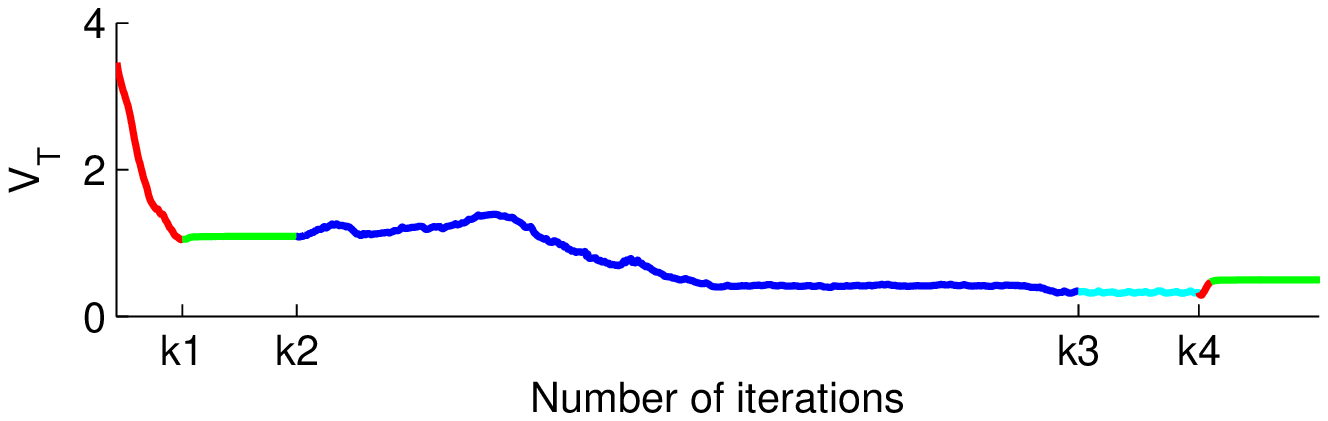}} &
{\includegraphics[height=0.135\textwidth,width=0.31\textheight,angle=270]{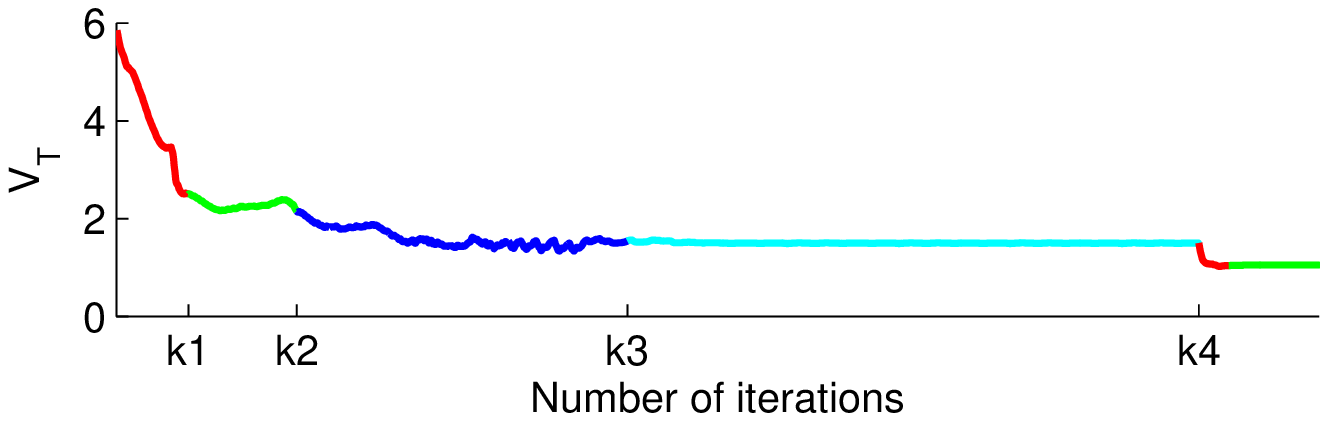}}\\\vspace{-10pt}
{\includegraphics[height=0.135\textwidth,width=0.31\textheight,angle=270]{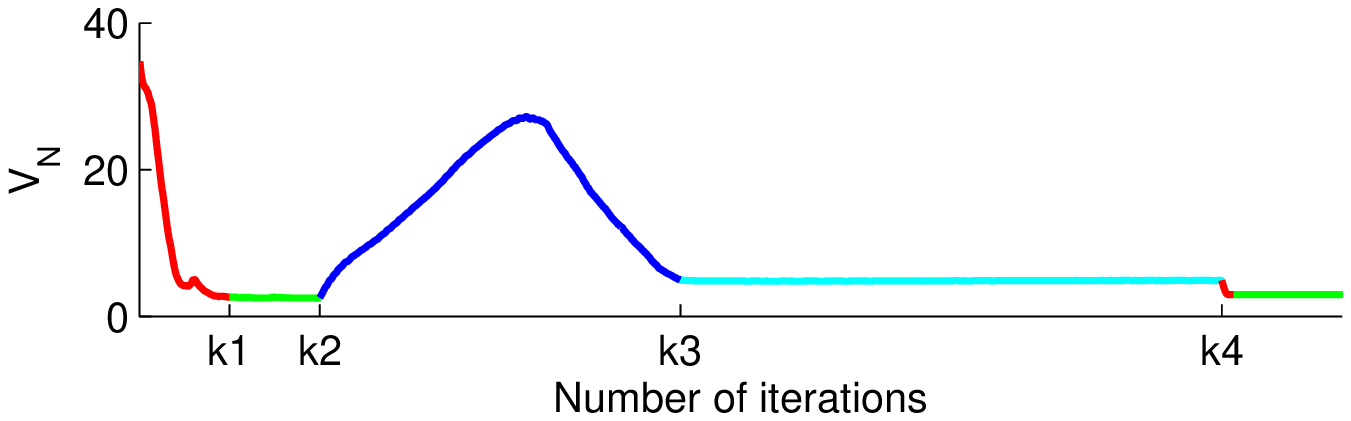}}&
{\includegraphics[height=0.135\textwidth,width=0.31\textheight,angle=270]{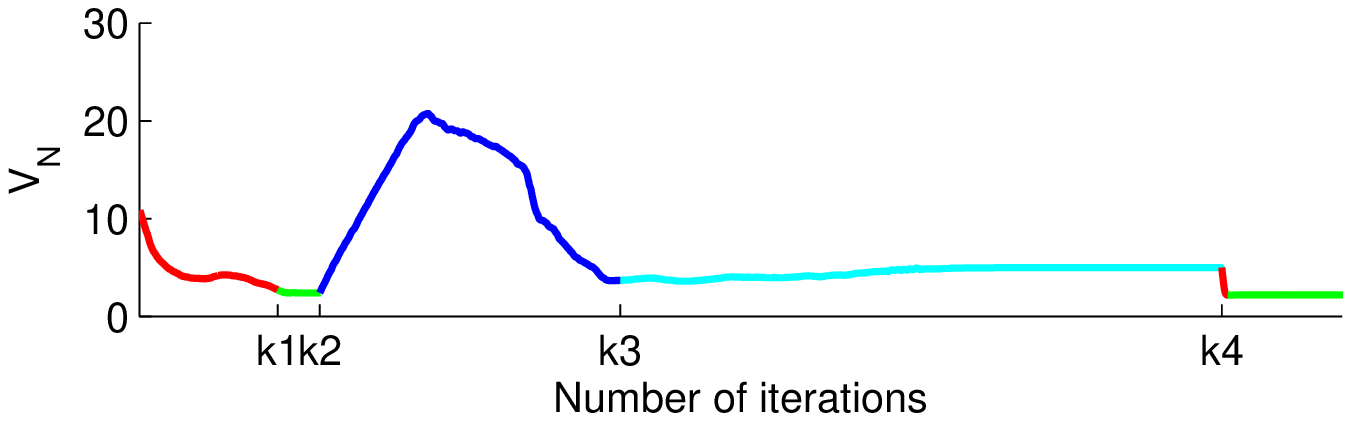}}&
{\includegraphics[height=0.135\textwidth,width=0.31\textheight,angle=270]{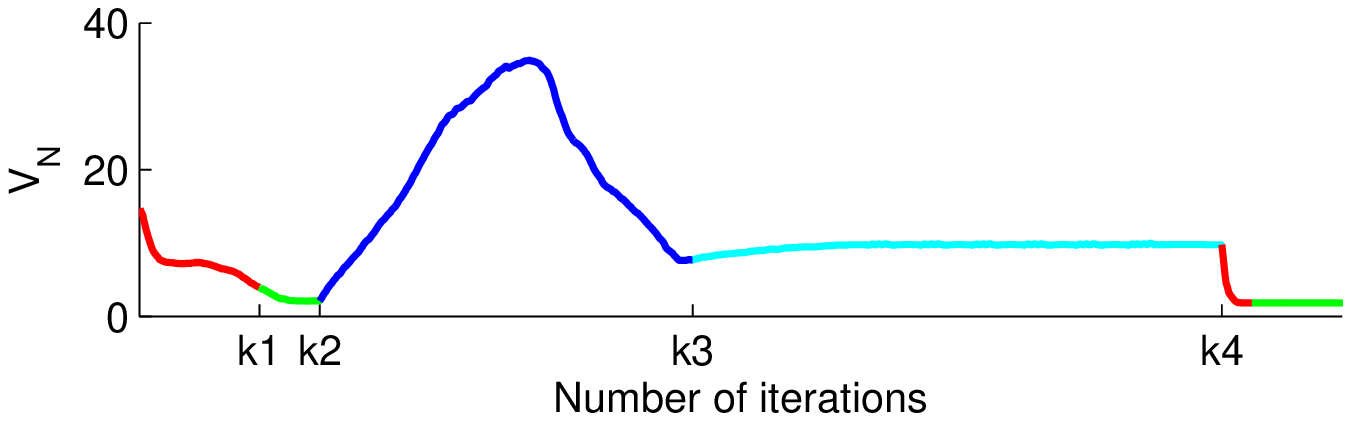}}&
{\includegraphics[height=0.135\textwidth,width=0.31\textheight,angle=270]{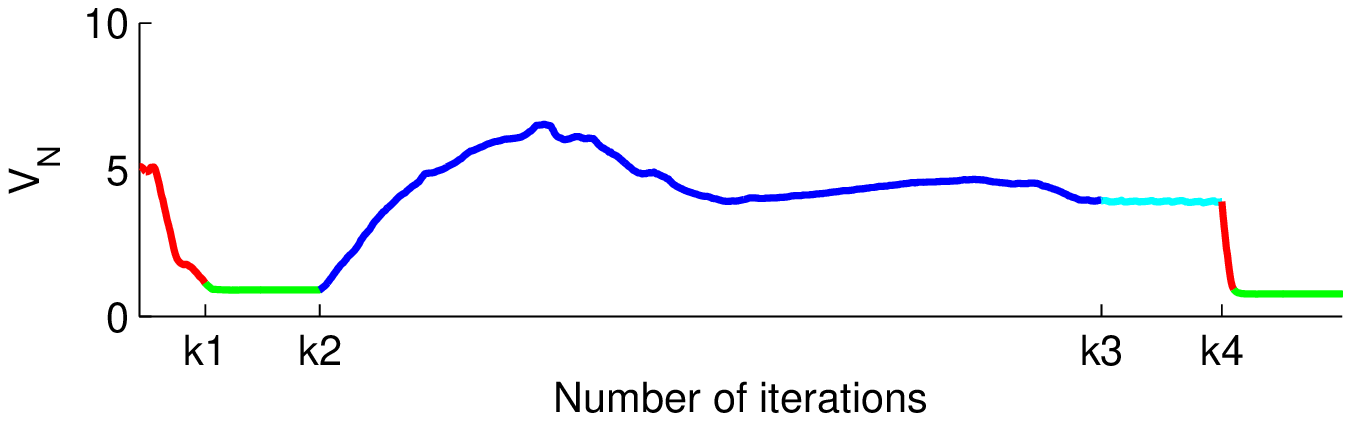}}&
{\includegraphics[height=0.135\textwidth,width=0.31\textheight,angle=270]{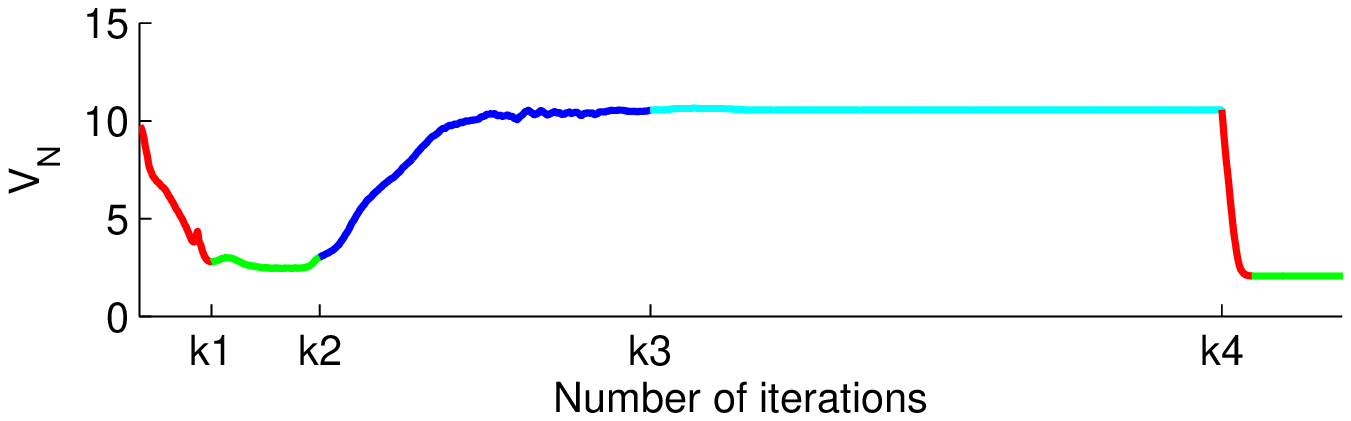}}\\
\end{tabular}
\caption{The normal and tangential velocities during curve evolutions. The top row shows the total magnitude of the tangential velocity along the curves; The second row shows the total magnitude of the normal velocity along the curves. k1,k2,k3,k4 area some key iterations in the curve evolutions. 0-k2 is the curve evolution by GAC; k2-k4 is by EF; from k4 onwards is by GAC}\label{Fig:Qtt}
\end{figure*}

In the following, a quantitative analysis of the proposed method is provided. From the quantitative results, the stable convergence of the curve evolution algorithms can also be observed. One major contribution in this paper is about the proposed solution, namely the EF, to the Pseudo Stationary Phenomenon that occurs if the tangential velocity on the converged curve does not vanish. The quantitative results demonstrate that the proposed method can reduce the tangential velocity effectively. The tangential velocity in the curve evolutions of the proposed method is shown in the second row from top in Figure \ref{Fig:Qtt}. The tangential velocity is quite large when GAC converges (0-k1), but due to the EF (k2-k4) the tangential velocity approaches zero, where k3 is the iteration of convergence. The velocities do not reach zero exactly because of the errors in discretization. The normal velocity in the third row from top in Figure \ref{Fig:Qtt} shows that the normal velocity drops significantly during the direct curve evolutions of the gradient descent flow. The iterations corresponding to k1, k2, k3, and k4 in Figure 3.12 are given in Table \ref{TB:KEYIter}. The computational times for the curve evolutions are also presented. The computational cost can be reduced by using fast implementations, e.g.  \cite{Adalsteinsson95NarrowBand} \cite{SethianBook} \cite{Shi08FastLS}.

\begin{table*}[!h]
\setlength{\tabcolsep}{6pt}
\renewcommand{\arraystretch}{2}
\caption{Image sizes, key iterations in Figure \ref{Fig:Qtt}, and time costs of the curve evolutions}\label{TB:KEYIter}
\centering
{\small\begin{tabular*}{\textwidth}{@{\extracolsep{\fill}}m{1cm}m{2cm}m{0.5cm}m{0.5cm}m{0.8cm}m{0.8cm}m{1cm}m{1cm}m{1cm}}
  \midrule
  \vspace{2pt}
Input image                   & Image size (px)   & k1  & k2 & k3 & k4 & Total time (s) & GAC time (s) & EF time (s) \\ \midrule
{\includegraphics[width=0.5in,height = 0.5in]{sb2_GAC00_frame_0000.eps}} &    80$\times$80    & 150  & {300} & 900 & {1800} &  62.44 & 4.68 & 18.73\\
{\includegraphics[width=0.5in,height = 0.5in]{three_GAC00_frame_0000.eps}} &    80$\times$80    & 230 & -- & 800 & -- &  62.33 & 7.17 & 15.58\\
{\includegraphics[width=0.5in,height = 0.5in]{new3s_GAC0_frame_0000.eps}} &    80$\times$80    & 200  & -- & 920 & -- &  63.34 & 6.33 & 19.64\\
{\includegraphics[width=0.5in,height = 0.5in]{1_GAC0_frame_0000.eps}} &  128$\times$128
& 110  & -- & 1600 & -- & 119.44 & 6.57 & 77.64 \\
{\includegraphics[width=0.5in,height = 0.5in]{two_GAC0_frame_0000.eps}} &  128$\times$128    & 120  & -- & 850 & -- &  120.23 & 7.21 & 33.06\\\hline
\end{tabular*}}
\end{table*}

\section{Discussions and conclusion}\label{SEC:CON}
The PSP problem may happen not only in object segmentation. It may also exist in other computer vision or computer graphics tasks involving optimizing closed curves. Hence, the proposed method may be extended to address the PSP in the different contexts. The proposed method is derived based on geometrical observations and analysis of the PSP problem. Experimentally, it shows that the EF repositions the curve to escape from the PSP. This suggests that one may also approach the PSP from the perspective of repositioning a pseudo-stationary curve. The assumption of small gradients along object boundary is not always valid. For example, the inhomogeneous region may leads to inhomogeneous gradients. This paper addresses the issue of early termination of curve evolution for images containing moderately complex structures and relatively homogeneous regions. The intensity inhomogeneity in the medical images can be corrected by preprocessing \cite{Vovk07CorInhomo}. This work provides new geometric insights of the problem of early termination of the curve evolution for general edge-based active contours, giving rise to new criteria and solution of edge-based active contours.